%% file: step3p5_turbo_2026.tex
\documentclass[11pt, a4paper, logo, copyright, nonumbering]{step3p5_turbo_2026}

\usepackage[utf8]{inputenc}
\usepackage[T1]{fontenc}
\usepackage{hyperref}
\hypersetup{
    pdfborderstyle={/S/S/W 1},
    linkbordercolor={0.4 0.7 1},
}
\usepackage{url}
\usepackage{booktabs}
\usepackage{amsfonts}
\usepackage{nicefrac}
\usepackage{microtype}
\usepackage{xcolor}
\usepackage{comment}
\usepackage{fancyvrb}
\usepackage{listings}
\usepackage{algorithm}
\usepackage{algorithmicx}
\usepackage{algpseudocode}
\usepackage{subcaption}
\usepackage{cancel}
\usepackage{xspace}
\usepackage{makecell}
\usepackage{geometry}
\usepackage{multirow}
\usepackage{cite}
\usepackage{blindtext}
\usepackage{multicol}
\usepackage{hyperref}
\usepackage{fontawesome5}

\usepackage{microtype}
\usepackage{booktabs}
\usepackage{amsmath}
\usepackage{graphicx}
\usepackage{CJKutf8}
\usepackage[colorinlistoftodos]{todonotes}

\newcommand{\eg}{\textit{e.g.}}
\newcommand{\ie}{\textit{i.e.}}

\newcommand{\ourmodel}{Step 3.5 Flash\xspace}

\usepackage[most]{tcolorbox}

\newtcolorbox{MyBox}[1]{
    colback=white,
    colframe=black,
    sharp corners,
    breakable,
    left=2mm,
    right=2mm,
    top=2mm,
    bottom=2mm,
    boxrule=0.1mm,
    fontupper=\ttfamily\small
}
\definecolor{glm_bg}{RGB}{235, 245, 255}
\definecolor{glm_text}{RGB}{0, 110, 220}
\definecolor{header_gray}{RGB}{242, 242, 242}
\definecolor{section_gray}{RGB}{230, 230, 230}
\newcommand{\best}[1]{\textbf{#1}}
\newcommand{\glmcell}[1]{\cellcolor{glm_bg}\textcolor{glm_text}{\textbf{#1}}}
\newcommand{\na}{-}

\definecolor{neg}{HTML}{CB4335}
\definecolor{pos}{HTML}{27AE60}
\definecolor{delta_gray}{HTML}{6C8EBF}
\definecolor{light_bg}{RGB}{235, 245, 255}

\title{\centering \ourmodel: Open Frontier-Level Intelligence \\ with 11B Active Parameters}

\newcommand{\HFicon}{\raisebox{-0.18em}{\includegraphics[height=1.1em]{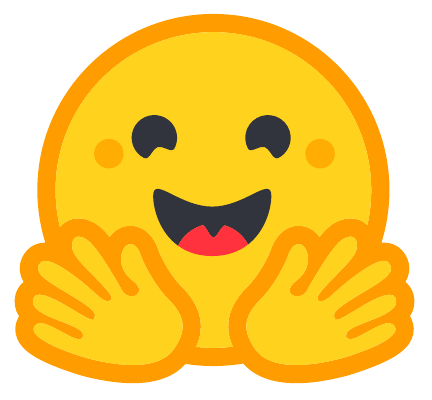}}}
\newcommand{\GHicon}{\raisebox{-0.18em}{\includegraphics[height=1.1em]{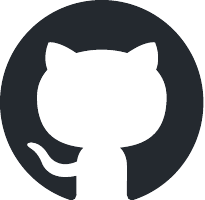}}}
\newcommand{\STicon}{\raisebox{-0.18em}{\includegraphics[height=1.1em]{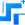}}}

\author[*]{
\large StepFun Team\\
\normalsize
\href{https://github.com/stepfun-ai/Step-3.5-Flash}{\GHicon\ GitHub}\quad
\href{https://huggingface.co/stepfun-ai/Step-3.5-Flash}{\HFicon\ HuggingFace}\quad
\href{https://static.stepfun.com/blog/step-3.5-flash/}{\STicon\ ModelBlog}
\vspace{-1cm}
}

\begin{document}

\begin{abstract}
We introduce \textbf{\ourmodel}, a sparse Mixture-of-Experts (MoE) model that bridges the gap between frontier-level agentic intelligence and computational efficiency. We focus on what matters most when building agents: reasoning that's sharp, and execution that's fast and reliable. Reflecting these priorities, \ourmodel pairs a \textbf{196B-parameter foundation} for high-fidelity modeling with \textbf{11B active parameters} for efficient inference, optimized by interleaved \textbf{3:1 Sliding Window/Full Attention} and \textbf{Multi-Token Prediction} (MTP-3) to minimize the latency and cost of multi-round agentic interactions.~Toward frontier-level intelligence, we design a scalable RL framework that integrates verifiable signals and preference feedback while maintaining stability during large-scale off-policy training to drive consistent self-improvement across mathematics, code, and tool use.
\ourmodel demonstrates strong intelligence across agent, coding, and math tasks, achieving 85.4\% on IMO-AnswerBench and 86.4\% on LiveCodeBench-v6 (2024.08–2025.05), 88.2\% on $\tau^2$-Bench, 69.0\% on BrowseComp (w. Context Manage), and 51.0\% on Terminal-Bench~2.0 \textemdash performance on par with frontier models such as GPT-5.2 xHigh and Gemini 3.0 Pro.
By redefining the efficiency frontier, \ourmodel provides a high-density foundation for deploying sophisticated agents in real-world industrial environments.
\end{abstract}

\maketitle
\input{src/figure_tex/01_teaser.tex}

\begingroup
\setlength{\parskip}{7.5pt}
\selectfont
\tableofcontents
\endgroup

\input{src/01_intro}
\input{src/011_arch}

\input{src/012_infra}

\input{src/02_pre_training}

\input{src/03_post_training}
\input{src/04_evaluation}
\input{src/05_discussion}

\input{src/tables/appendix-author-list}

\newpage
{\Large\textbf{Appendix}}
\addtocontents{toc}{\protect\setcounter{tocdepth}{2}}
\appendix
\input{src/A_append}

\bibliographystyle{unsrt}
{
\linespread{0.75}\selectfont
\bibliography{references.bib}
}

\end{document}

%% file: src/figure_tex/01_teaser.tex
\begin{figure}[H]
  \centering
    \includegraphics[width=0.95\linewidth]{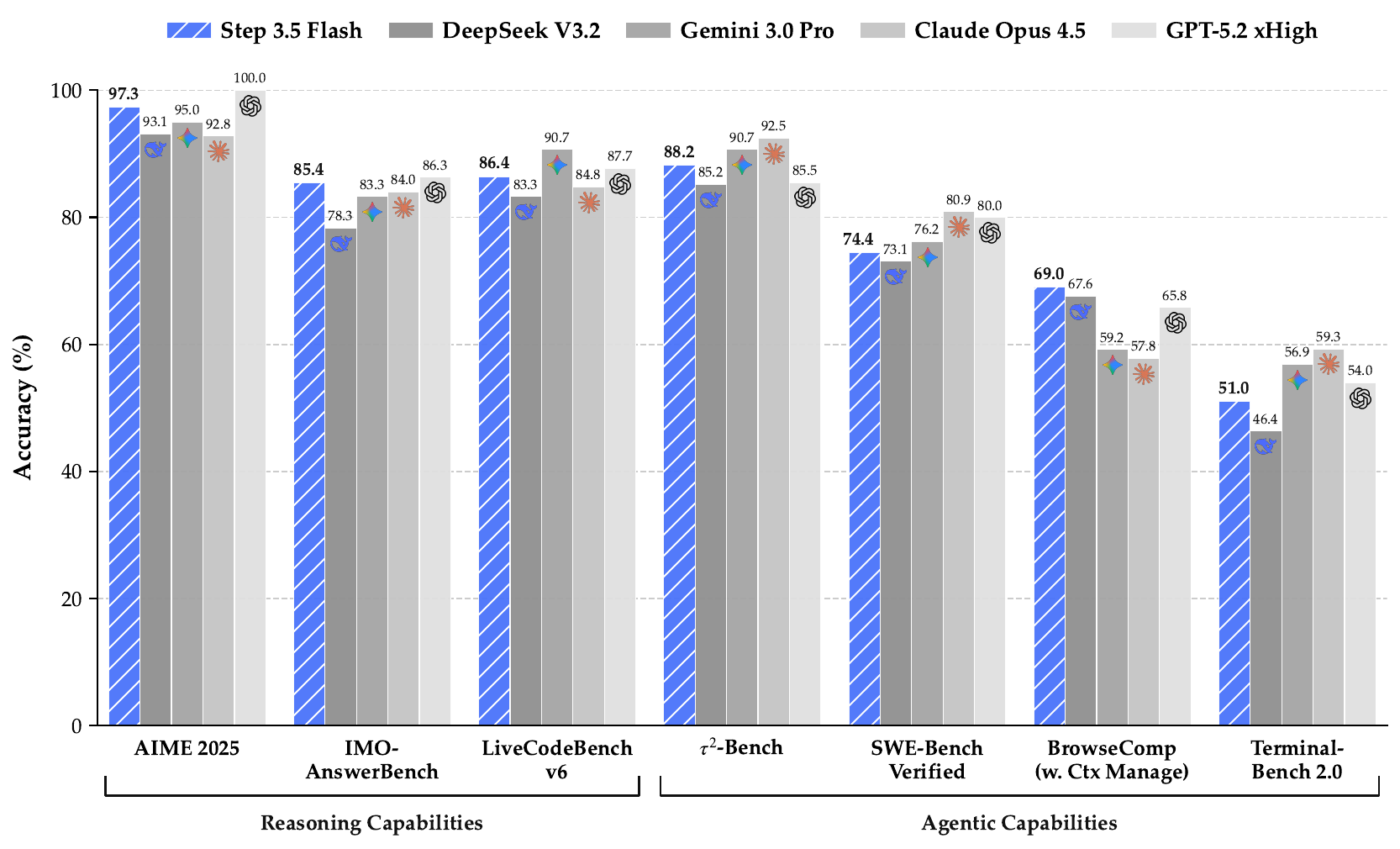}
  \caption{\ourmodel achieves frontier-level intelligence with only 11B active parameters (196B MoE), comparable to leading closed and open-source models.}
  \label{fig:teaser}
\end{figure}

%% file: src/01_intro.tex
\newpage
\section{Introduction}
\label{sec:intro}
While open-source large language models (LLMs)~\cite{deepseekai2024deepseekv32, zeng2025glm, xiao2026mimov2flashtechnicalreport, team2025longcat, kimi_k2, minimax_m2p1} have rapidly narrowed the performance gap with closed-source frontier systems~\cite{gpt_5p2,gemini_3,opus} across verifiable tasks~\cite{10.5555/3524938.3525416,hmmt25,jain2024livecodebench}, new challenges emerge as agentic systems gain prominence.
In particular, open-source models still trail closed-source frontiers in complex reasoning.
Furthermore, critical efficiency bottlenecks hinder their application in long-context agentic tasks~\cite{swe_verified,yang2025swesmith,tau2bench2025,terminalbench2025,wei2025browsecomp,zhou2025browsecompzhbenchmarkingwebbrowsing,mialon2023gaiabenchmarkgeneralai,chen2025xbench,sharma2025researchrubrics}, let alone deployment in edge or resource-constrained settings.

In designing the architecture of \ourmodel, we focus on two core aspects: efficiency and capacity.
We adopt a sparse Mixture-of-Experts (MoE) ~\cite{fedus2021switchtransformers,zoph2022stmoedesigningstabletransferable,pmlr-v162-du22c,lepikhin2020gshardscalinggiantmodels,dai2024deepseekmoeultimateexpertspecialization}  architecture with 196B total parameters and only 11B activated per token, together with a 3:1 ratio of sliding-window attention (SWA)~\cite{DBLP:journals/corr/abs-1904-10509} to full attention and multi-token prediction (MTP-3)~\cite{gloeckle2024better,deepseek2024deepseekv3,xiaomi2025mimo,xiao2026mimov2flashtechnicalreport} to reduce long-context latency.
To improve capacity under hybrid attention with minimal overhead, we increase the number of query heads in sliding-window attention (SWA) layers from 64 to 96 and use head-wise gated attention \cite{qiu2025gatedattentionlargelanguage}.
This design enables large-scale online deployment, sustaining $\sim$170 tokens/s on Hopper GPUs during the first week on OpenRouter \footnote{https://openrouter.ai}.

On the pretraining side, we treat stability as a first-class requirement and build a comprehensive observability and diagnostic stack via a lightweight asynchronous metrics server with micro-batch-level continuous logging.
This infrastructure enables systematic identification and mitigation of large-scale MoE failure modes (\textit{e.g.}, Muon-related precision sensitivity, expert collapse \cite{step3blog}, and activation blow-ups \cite{kimi_k2,openai2025gptoss120bgptoss20bmodel}). Combined with an improved Muon optimizer \cite{jordan2024muon} that offers more accurate and stable updates,
we achieve stable training over 17.2T high-quality and diverse tokens with only a single transient loss spike.
With this stable training regime, \ourmodel Base achieves competitive performance against larger counterparts, such as DeepSeek-V3.2-Exp Base \cite{deepseekai2024deepseekv32} and Kimi-K2-Base \cite{kimi_k2}, on math, coding and knowledge benchmarks. Notably, on SimpleQA~\cite{wei2024measuring}, it scores 31.6\%, surpassing DeepSeek-V3.2-Exp Base despite using only one-third of the parameters.

Toward frontier-level intelligence, current post-training systems face two tightly coupled challenges: inefficient iteration of domain-specific experts for self-distillation~\cite{deepseekai2024deepseekv32, zeng2025glm, xiao2026mimov2flashtechnicalreport, team2025longcat} and limited scalability of Reinforcement Learning~(RL) to long-horizon reasoning for MoE models.
Training a single generalist to directly cover diverse domains often sacrifices domain-specific expertise, whereas maintaining separate expert models leads to fragmentation and an unsustainable cost of continual multi-model iteration.
At the same time, as models are extended to deeper reasoning trajectories, even small token-level discrepancies in off-policy rollouts can accumulate into high-variance gradients.
This effect is particularly severe in MoE models, where expert-level routing induces larger distributional shifts and destabilizes optimization in the frontier performance regime~\cite{zheng2025group,yao2025offpolicy,ma2025stabilizing,deepseekai2024deepseekv32}.

To address these challenges, we propose a unified post-training recipe for large-scale RL built on a shared SFT foundation.
The framework alternates between domain-specific specialization and global synthesis, enabling efficient expert iteration while maintaining a single, high-performing generalist.
A dedicated mid-training phase scales the context window to 128k and strengthens core agentic and reasoning capabilities via synthetic data, providing a strong initialization for downstream post-training.
To support stable and scalable RL within this unified framework, we introduce Metropolis Independence Sampling-Filtered Policy Optimization (MIS-PO)~\cite{metropolis1953equation,hastings1970monte}, replacing continuous importance weighting with discrete, distributional filtering at both token and trajectory levels.
By restricting optimization to samples within a stable trust region, MIS-PO substantially reduces gradient variance while preserving effective learning signals, enabling RL to scale reliably to long-horizon reasoning and agentic behaviors.

\ourmodel achieves competitive performance with leading frontier models and systems across a broad range of reasoning and agentic benchmarks, despite 11B active parameters.
It delivers strong results under standard inference on reasoning tasks, including 85.4\% on IMO-AnswerBench~\cite{luong2025robustmathematicalreasoning} and 86.4\% on LiveCodeBench-v6~(2024.08–2025.05)~\cite{jain2024livecodebench}, while also demonstrating robust long-horizon, tool-augmented capabilities with 88.2\% on $\tau^2$-Bench~\cite{tau2bench2025}, 69.0\% on BrowseComp (with context management)~\cite{wei2025browsecomp}, and 51.0\% on Terminal-Bench 2.0~\cite{terminalbench2025}.
With PaCoRe~\cite{hu2026pacorelearningscaletesttime} deep think inference, \ourmodel further improves performance on reasoning-intensive benchmarks requiring extended deliberation and multi-round synthesis.
Taken together, these results indicate that \ourmodel substantially narrows the gap between advanced open models and frontier proprietary systems in both reasoning and agentic settings.

%% file: src/011_arch.tex
\section{Architecture}
\label{sec:arch}

\subsection{Design Philosophy}

The architecture of {\ourmodel} reflects a paradigm shift in model--system co-design. Beyond the traditional objectives of intelligence and cost, the era of autonomous agents elevates a third critical constraint: \textbf{\textit{inference latency}}. In interactive agentic workflows~\cite{anthropic_agent_workflow,codex_agent_workflow}, minimized latency translates directly to reduced wall-clock time for task completion, or conversely, allows for increased intelligence within a fixed time budget via test-time scaling~\cite{snell2025scaling,muennighoff2025s1,yang2025multiverse,hu2026pacorelearningscaletesttime}.

Agentic workloads typically exhibit a distinct profile: extensive context prefilling followed by prolonged, multi-turn interactive decoding.
Accordingly, we co-design {\ourmodel} for low wall-clock latency along three coupled axes: \emph{attention} (to accelerate long-context processing and have good affinity with MTP), \emph{sparse MoE} (to prevent stragglers in distributed deployments that reduce throughput), and \emph{multi-token prediction} (MTP; to facilitate fast generation through speculative decoding).

\paragraph{Attention.}
To accelerate prefilling, we employ a hybrid attention mechanism~\cite{Beltagy2020Longformer,gemmateam2025gemma3technicalreport,openai2025gptoss120bgptoss20bmodel} to mitigate the quadratic complexity of long-context processing.
For decoding, we prioritize architectural compatibility with speculative decoding~\cite{10.5555/3618408.3619203}, since verification efficiency is the dominant lever on bandwidth-bound hardware.
These considerations motivate two attention design decisions:
\begin{itemize}
    \item \textbf{\textit{Sliding-Window Attention (SWA).}}
We select SWA~\cite{DBLP:journals/corr/abs-1904-10509} over linear attention~\cite{10.5555/3524938.3525416,schlag2021linear} to maximize decoding efficiency.
Although both have linear complexity, the state-update mechanism of linear attention complicates efficient draft tree generation and parallel tree verification needed for speculative decoding~\cite{wang2025opt,xiong2025dyspec,10.5555/3692070.3694435}.
In contrast, SWA preserves standard attention semantics and remains inherently amenable to parallel verification via $KV$ masking.
Moreover, in the absence of robust empirical evidence that linear attention yields superior long-context modeling for agentic tasks, we find that SWA with window size $W{=}512$ strikes a favorable balance between kernel efficiency and capturing local dependencies.

    \item \textbf{\textit{Hardware-Aligned Grouped-Query Attention (GQA-8).}}
Targeting deployment on standard 8-GPU server nodes, we configure the model with eight $KV$ heads (GQA-8)~\cite{ainslie-etal-2023-gqa}.
This aligns $KV$-cache sharding with 8-way tensor parallelism and improves memory access patterns.
Crucially, while GQA-8 makes attention more memory-bandwidth bound, it also creates computational slack that can absorb speculative drafting and verification overhead, enabling aggressive multi-token speculation without a proportional latency penalty.
\end{itemize}

\paragraph{Sparse MoE.}
On the feed-forward side, we employ fine-grained MoE~\cite{fedus2021switchtransformers,zoph2022stmoedesigningstabletransferable,pmlr-v162-du22c,lepikhin2020gshardscalinggiantmodels,dai2024deepseekmoeultimateexpertspecialization} to reduce the average FFN compute while maintaining capacity.
Expert parallelism (EP)~\cite{lepikhin2020gshardscalinggiantmodels} is utilized to enable scalable deployment.
However, under EP, end-to-end latency can be dominated by \emph{stragglers} induced by routing imbalance: token assignment skew concentrates workload on a small subset of experts and their hosting GPUs, throttling throughput at synchronization points.
We therefore introduce an \textit{EP-Group Balanced MoE Routing} strategy.

\paragraph{Multi-Token Prediction (MTP).}
To further reduce autoregressive latency, we incorporate Multi-Token Prediction (MTP)~\cite{li2024eagle,deepseek2024deepseekv3} as a complementary lever to speculative decoding~\cite{10.5555/3618408.3619203}.
To keep speculation lightweight, we streamline the MTP heads by leveraging SWA and dense FFNs~\cite{xiao2026mimov2flashtechnicalreport}.

We further constrain the model size to under 200B parameters, enabling high-performance inference within the 128GB memory budget of high-end workstations.

\subsection{Sparse MoE Backbone with Hybrid Attention}

\begin{figure}[t]
    \centering
    \includegraphics[width=1.0\linewidth]{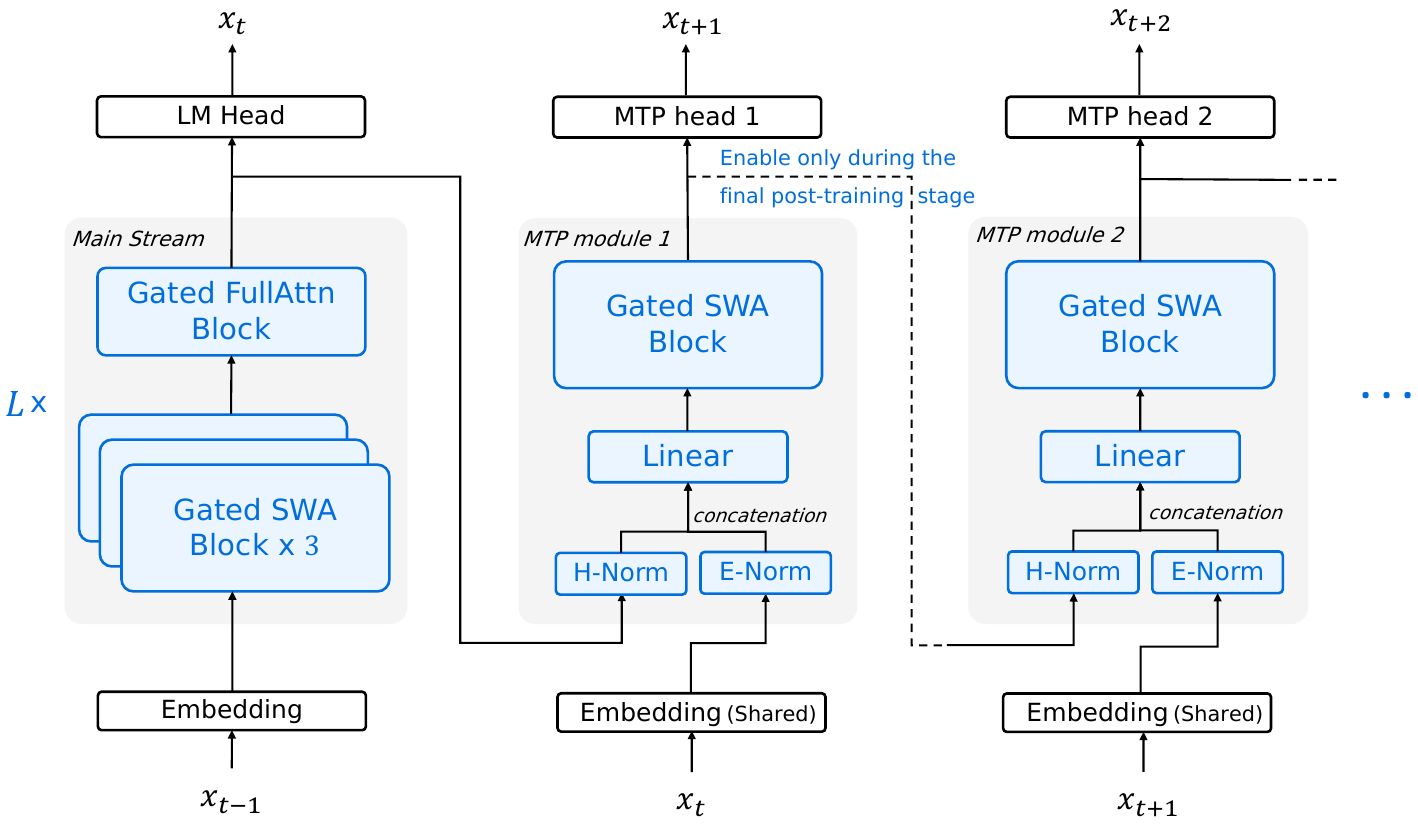}
    \caption{{Illustration of \ourmodel.}
The model uses head-wise gated attention \cite{qiu2025gatedattentionlargelanguage} with a leading Full Attention layer followed by $L=11$ Hybrid Blocks, each interleaving 3 Sliding Window Attention (SWA) layers with one Full Attention layer (\emph{for visual clarity, the first layer is omitted in the figure}).
We apply zero-centered RMSNorm \cite{gemmateam2024gemmaopenmodelsbased} throughout.
The first three blocks use dense FFNs; later blocks employ sparse MoE FFNs.
MTP modules use SWA and dense FFNs. To limit overhead, only MTP module 1 is trained during main training; MTP modules 2–3 are cloned from it and jointly fine-tuned in a lightweight final phase.
\label{fig:arch}}
\end{figure}

As illustrated in Figure~\ref{fig:arch}, {\ourmodel} adopts a 45-layer sparse-MoE Transformer backbone (3 dense layers and 42 MoE layers) paired with a specialized hybrid attention {layer layout}.
Each MoE layer contains 288 routed experts plus one shared expert, with a top-$k$ router activating $k{=}8$ experts per token. This configuration maintains an extensive knowledge capacity (196B total parameters) while restricting per-token activation to just 11B, ensuring inference latency remains low enough for highly responsive agent interaction.
Table~\ref{tab:arch_hparams} summarizes key architecture hyperparameters of {\ourmodel}.

\paragraph{Hybrid Attention Layer Layout.}\label{sec:hybrid_swa}
To balance long-context efficiency with robust long-range connectivity, {\ourmodel} leverages an interleaved attention layout at a $3:1$ ratio (SWA : Full) inspired by \cite{openai2025gptoss120bgptoss20bmodel,gemmateam2025gemma3technicalreport,cohere2025commandaenterprisereadylarge}, denoted as $S3F1$.
This configuration repeats a four-layer motif consisting of three SWA layers ($W{=}512$) followed by a single full GQA-8 layer.
However, in our initial experiments, a naive interleaving strategy consistently underperforms a dense attention baseline across various benchmarks (Table~\ref{tab:swa_pretrain_results}).
To bridge this performance gap without adding practical overheads,
we leverage two complementary enhancements:
(i) an increased SWA query-head count, and
(ii) adopting head-wise gated attention~\cite{qiu2025gatedattentionlargelanguage}.

\paragraph{Augmented Query Heads in SWA.}

Using a higher query-head number (from $64$ to $96$) effectively mitigates performance drop typically observed when transitioning from a uniform full-attention architecture to the $S3F1$ layout (Table~\ref{tab:swa_pretrain_results}). We consider this to be nearly a ``free lunch''. Because in long-text scenarios, the overhead of naive SWA is very small, even though our solution scales up significantly.

\paragraph{Head-wise Gated Attention.}
A limitation of naive SWA is its inability to effectively absorb unused attention weights when there is no useful information in the input window~\cite{xiao2024efficient,sun2024massive,gu2025when, qiu2025gatedattentionlargelanguage}. Previous work~\cite{openai2025gptoss120bgptoss20bmodel,xiao2026mimov2flashtechnicalreport}
introduce learnable, data-independent sink tokens into the window to address this issue. Instead, we opt for a different approach by integrating a parameter-efficient head-wise gating mechanism~\cite{jumper_highly_2021,qiu2025gatedattentionlargelanguage,lin2025forgetting}, which can be viewed as integrating data-dependent sink tokens.
Please refer to Appendix~\ref{appx:gated_attn_details} for implementation details and further discussion. Head-wise gating is also negligible to both theoretical FLOPs and practical latency. We report more performance analysis and benchmarks for gating and augmenting the number of SWA heads in Appendix~\ref{appx:attn_enhancement_benchmark}.

\paragraph{MoE Expert-parallel Load Balancing.}\label{sec:moe_design}

We use loss-free load-balancing~\cite{deepseek2024deepseekv3,wang2024lossfreebalancing} to encourage
\emph{global} token balance across experts.
However, this approach does not guarantee balanced loads across EP ranks at the micro-batch level, potentially leading to stragglers and reduced throughput.
We therefore introduce an EP-level balancing loss that explicitly promotes uniform rank-level utilization~\cite{dai2024deepseekmoeultimateexpertspecialization}.

\noindent
EP partitions experts $\mathcal{E}$ into $G$ disjoint groups $\{\mathcal{E}_g\}_{g=1}^{G}$ across ranks. For token $t$, let $S_t$ denote the top-$K$ experts (mask $s_{t,e}=\mathbf{1}[e\in S_t]$) and $p_{t,\cdot}$ the routing probabilities. Then, the EP load balancing loss $\mathcal{L}_{EP}$ is:
\begin{equation}
\begin{aligned}
p_e = \frac{1}{T}\sum_{t=1}^{T} p_{t,e}, & &
f_e = \frac{1}{TK}\sum_{t=1}^{T} s_{t,e}, & &
p_g = \sum_{e\in\mathcal{E}_g} p_e, & &
f_g = \sum_{e\in\mathcal{E}_g} f_e, & &
\mathcal{L}_{\mathrm{EP}} &= G\sum_{g=1}^{G} f_g\,p_g.
\end{aligned}
\end{equation}

\paragraph{Multi-token Prediction (MTP).}
To speedup speculative decoding on long-context agentic workloads, we attach three lightweight multi-token prediction (MTP) heads.
Each MTP head consists of a SWA and a dense FFN, adding only 0.81B parameters ($\sim$0.41\%).
We index these heads by their \emph{additional} prediction offset beyond the standard LM head: for $h\in\{1,2,3\}$, {MTP-$h$} predicts the token $x_{t+1+h}$ conditioned on the backbone hidden states at position $t$.
To control training overhead, we activate and optimize only MTP-$1$ in most training stages.
Once the backbone is well-trained, we initialize MTP-$2$ and MTP-$3$ from MTP-$1$ and jointly train all MTP heads in a lightweight post-training phase.
Inspired by Fast-MTP~\cite{cai2025fastmtpacceleratingllminference}, we adopt position-dependent loss reweighting across prediction offsets in MTP heads to prevent over-optimizing for distant-token predictions.

\subsection{Architecture Ablations and Results}

We conduct extensive experiments to validate key design choices in \ourmodel, focusing on (i) attention layouts, including SWA and head scaling, and (ii) head-wise gated attention versus sink tokens. To ensure our efficiency optimizations do not degrade model performance, we adopt two complementary ablation protocols: one evaluates full end-to-end pipelines covering pre-training, 32k long-context extension, and 64k context-length supervised fine-tuning (SFT), and the other scales the analysis up to 100B parameters to study how these design choices behave with scale. Detailed architecture and evaluation setups for all tables are provided in Appendix~\ref{sec:arch_ablation}.  Key findings from these large-scale experiments are summarized below.

\paragraph{SWA w.r.t. Long Context.}
\label{sec:swa_ratio_head_30b}
We train a 30B-A3B model through the full pipeline (1.4T-token pretraining followed by SFT) to evaluate the end-to-end impact of hybrid attention on reasoning and long-context performance.
We ablate four attention layouts: all-full attention ($FFFF$), alternating SWA/full ($S1F1$), a 3:1 SWA-to-full layout ($S3F1$), and an $S3F1$ variant with increased SWA query heads (\texttt{$S3F1$+Head}).
To isolate attention-structure effects, we fix the SWA window size to $W{=}512$ and disable MTP (see Appendix, Table~\ref{tab:train_setting_1p4t} and Table~\ref{tab:swa_pretrain_results}).

Table~\ref{tab:swa_sft_results} shows a clear cost--quality trade-off across layouts.
$S3F1$ achieves the lowest normalized attention-side FLOPs (normalized to 1.00 for prefill and 1.00 for decode separately), whereas $FFFF$ is $\sim$2.68$\times$/2.90$\times$ as expensive as $S3F1$; however, $S3F1$ exhibits a consistent quality degradation (\eg, LongCtx drops from 28.8 to 27.5).

Increasing the number of SWA query heads largely compensates for this loss.
Notably, \texttt{$S3F1$+Head} already surpasses $FFFF$ during pretraining (55.7 vs.\ 54.1), and remains competitive after post-training: LongCtx improves from 27.5 to 28.2 and Sci from 42.4 to 44.0, closing most of the gap to the $FFFF$ baseline with negligible additional attention cost.
The remaining downside is limited and localized (\eg, a modest drop on Code to 18.3), while overall quality trends favor \texttt{$S3F1$+Head}.

Interestingly, the alternating $S1F1$ layout delivers the best overall SFT quality and the strongest LongCtx score (29.6), but requires substantially higher attention-side prefill/decode FLOPs ($\sim$1.58/1.65), about a 60\% cost increase relative to \texttt{$S3F1$+Head}.
We therefore adopt \texttt{$S3F1$+Head} as the default configuration for long-context agentic workloads, prioritizing its much lower prefill/decode cost with strong and stable long-context performance.

\begin{table}[t]
\centering
\small
\renewcommand{\arraystretch}{1.12}
\resizebox{\textwidth}{!}{
\begin{tabular}{l c c c ccccccc}
\toprule
\multirow{2}{*}{\textbf{Layout}} &
\multirow{2}{*}{\makecell{\textbf{SWA} \\ \textbf{Heads}}} &
\multicolumn{1}{c}{\textbf{Rel. FLOPs}} &
\multirow{2}{*}{\shortstack{\textbf{Pre-train}\\\textbf{Avg.}}} &
\multicolumn{7}{c}{\textbf{Downstream Performance}} \\
\cmidrule(lr){3-3} \cmidrule(lr){5-11}
& & \textbf{\small{Decode / Prefill}} & &
\textbf{Reasoning} & \textbf{Math} & \textbf{Code} & \textbf{Sci} & \textbf{General} & \textbf{LongCtx} & \textbf{Avg.} \\
\midrule

{$FFFF$} & 32 &
$\sim$2.68 / 2.90 &
54.1 &
40.8 & 40.9 & \textbf{19.6} & 42.7 & 26.5 & 28.8 & 33.2 \\

{$S1F1$} & 32 &
$\sim$1.58 / 1.65 &
54.6 &
\textbf{42.1} & \textbf{42.3} & 19.3 & \textbf{44.5} & \textbf{26.8} & \textbf{29.6} & \textbf{34.1} \\

{$S3F1$} & 32 &
\phantom{$\sim$}\textbf{1.00 / 1.00} &
53.6 &
40.2 & 40.4 & 18.9 & 42.4 & 25.4 & 27.5 & 32.5 \\

\texttt{$S3F1$+Head} & 48 &
$\sim$1.01 / 1.02 &
\textbf{55.7} &
40.6 & 40.3 & 18.3 & 44.0 & 26.0 & 28.2 & 32.9 \\

\bottomrule
\end{tabular}
}
\caption{Downstream results on 30B-A3B. $F$ denotes full attention and $S$ denotes SWA.
$S3F1$ indicates three $S$ layers followed by one $F$ layer in the hybrid layout.
Rel. FLOPs are normalized to the $S3F1$ configuration and averaged over 64k/256k contexts (Table~\ref{tab:rel}).
Pre-train Avg. aggregates results across general, math, and code benchmarks (Table~\ref{tab:ablation_study}).}
\label{tab:swa_sft_results}
\end{table}

\begin{table}[t]
    \centering
    \footnotesize
    \renewcommand{\arraystretch}{1.15}
    \begin{tabular}{@{}l c c c c c c c@{}}
        \toprule
        \textbf{Method} &
        \textbf{BBH} & \textbf{MMLU} & \textbf{GPQA} & \textbf{MBPP} & \textbf{C-EVAL} & \textbf{CMMLU} & \textbf{Avg.} \\
        \midrule
        {Sink Token} &
        70.6 & 65.1 & 27.2 & 61.2 & 76.2 & 74.6 & 62.5 \\
        {Head-wise Gate} &
        \best{73.7} & \best{67.0} & \best{28.1} & \best{62.6} & \best{77.9} & \best{77.1} & \best{64.4} \\
        \bottomrule
    \end{tabular}
    \caption{Pretraining-only evaluation on a 100B-A10B model under the $S3F1$ layout.
    Head-wise gating consistently outperforms a fixed sink token across benchmarks, including the overall average.}
    \label{tab:gate_vs_sink_100b}
\end{table}

\paragraph{Head-wise Gated Attention vs.\ Sink Tokens.\label{sec:head_vs_sink}}
We conduct {scaled, controlled pretraining experiments} on a {100B-A10B MoE} to study attention-side mechanisms under realistic scaling conditions.
Specifically, we compare \textit{sink tokens} and \textit{head-wise gated attention} while holding the attention layout fixed to the same $S3F1$ configuration with window size $W{=}512$.
As shown in Table~\ref{tab:gate_vs_sink_100b}, head-wise gating consistently improves quality, raising the average performance from 62.46 to 64.43 (+1.97).
We therefore adopt head-wise gated attention as the default mechanism in subsequent studies.

%% file: src/012_infra.tex
\section{Infrastructure}
\label{sec:infra}

\subsection{Compute Cluster}
{\ourmodel} is trained on a large-scale cluster with 4,096 NVIDIA H800 GPUs. Each node contains 8 GPUs interconnected through NVLink and NVSwitch for high-bandwidth intra-node communication. For inter-node connectivity, the cluster relies on 8×200 Gbps RoCE
links to maintain efficient synchronization and data exchange at scale.

\subsection{Training Framework}
The training of {\ourmodel} is powered by our internal \emph{Steptron} framework, a lightweight high-performance system built on top of PyTorch~\cite{Ansel_PyTorch_2_Faster_2024} and Megatron-LM~\cite{megatron-lm}.
Steptron unifies the full model development pipeline, supporting large-scale pre-training, post-training, and reinforcement learning (RL) workloads under a single engineering stack.

{\ourmodel} employs a hybrid parallelization strategy, including 8-way pipeline parallelism (PP)~\cite{narayanan2021efficientlargescalelanguagemodel} with virtual pipeline stages (VPP), and 8-way expert parallelism (EP)~\cite{lepikhin2020gshardscalinggiantmodels}, and ZeRO-1 Data Parallelism (DP)~\cite{rajbhandari2020zero}.
In order to facilitate efficient training of \ourmodel, we employ the following engineering techniques.

\paragraph{Decoupled Parallelism.}
Following Megatron-Core~\cite{liu2025moeparallelfoldingheterogeneous}, we implement a decoupled parallelization scheme that allows the attention and MoE modules to use different parallelization strategies. We assign them independent parallel groups and perform gradient reduction and scaling within each module's corresponding data-parallel group.

\paragraph{Communication Optimization.}
Concurrent DP communication streams for decoupled attention and MoE can saturate RoCE links, incurring considerable increases in DP overheads due to congestion. To address this, we propose two complementary communication optimizations that jointly reduce iteration time by up to 5\%. First, \emph{fabric-aware communication scheduling} partitions DP traffic into intra-node NVLink and inter-node RoCE phases, and pipelines them to fully utilize both fabrics. Second, \emph{communication-aware rank placement} uses job-level communication profiles to place ranks across switches, reducing hop counts and steering heavy traffic away from inter-switch hotspots.

\paragraph{Muon ZeRO-1 Resharding.}
Muon~\cite{jordan2024muon} requires full (unsharded) per-parameter gradients for Newton--Schulz orthogonalization, which conflicts with ZeRO-1~\cite{rajbhandari2020zero} reduce-scatter that shards a parameter's gradient across DP ranks. The current implementation in Megatron-LM resolves this mismatch by naively all-reducing FP32 gradients to reconstruct full gradients prior to the Muon update but nearly doubles communication. We instead assign whole parameters to DP ranks and repack the gradients buffer into a rank-major buffer so a single reduce-scatter delivers each parameter's complete gradient to its owner. Since padding to the fattest rank incurs overhead that grows with the data-parallel size, we apply this only to expert parameters and use DP all-reduce for non-expert parameters. This hybrid strategy reduces end-to-end iteration time by approximately 5\% with less than 4~GB additional memory compared to the naive all-reduce baseline.

\paragraph{GPU Kernels Optimization.}
We also apply kernel-level optimizations to improve training efficiency.
In attention, we fuse QK normalization with RoPE.
In MoE, we fuse multiple small operators to reduce kernel-launch overhead and memory traffic, and implement a fused MoE gather/scatter with grouped GEMM, similar to SonicMoE~\cite{guo2025sonicmoeacceleratingmoeio}.

\paragraph{Fine-grained Selective Checkpointing.}
Our training framework supports fine-grained activation recomputation with per-layer, submodule-level toggles (\eg, attention, FFN, normalization, SiLU, and MoE permutation), enabling selective recomputation of only the most memory-intensive components to reduce peak memory with minimal overhead.

\subsection{High-Throughput Lightweight Monitoring}\label{sec:log}

We collect a comprehensive suite of metrics (e.g., expert distribution within each micro-batch and gradient norms) for fine-grained monitoring of the training.
However, the telemetry scale is immense: a 4{,}096-GPU workload generates nearly 6 million messages per iteration. Conducting a synchronous global reduction within the main loop would introduce a significant overhead of several seconds, effectively doubling the iteration time, which is clearly intolerable for high-performance training. To mitigate this, we develop a Lightweight Metrics Server to decouple telemetry processing from the training path. Each rank utilizes \textit{StepRPC}, an in-house asynchronous communication framework, to asynchronously offload local metrics to the remote server. This approach reduces telemetry overhead to approximately 100 ms per iteration.

The Metrics Server buffers incoming metrics and triggers reduction and database persistence only after receiving \textit{end-of-iteration} signals from all participating ranks, eliminating synchronization in the main loop. To ingest and process millions of messages with low latency, the server is implemented as a high-concurrency multi-process system with two decoupled modules: (i) a Message Receiver optimized for high-throughput ingestion, and (ii) a Reduction Processor responsible for aggregation and persistence. By exploiting multi-core parallelism within and across these modules, the server keeps pace with the telemetry stream and ensures that metrics management never lags behind training.

%% file: src/02_pre_training.tex
\section{Pre-Training and Mid-Training}
\label{sec:training_pre_mid}

\paragraph{Overview.}
This section summarizes our {pre-training and mid-training} process, with an emphasis on the practical stability constraints of large-scale sparse MoE training.
We first describe training stability diagnostics and mitigations (Section~\ref{sec:stability}), then detail the curriculum used for pre-training and mid-training, including the data mixture, schedule, and key hyper-parameters (Section~\ref{sec:training_curriculum}).

\subsection{Training Stability}
\label{sec:stability}

\begin{figure}[t]
    \centering
    \includegraphics[width=1.\linewidth]{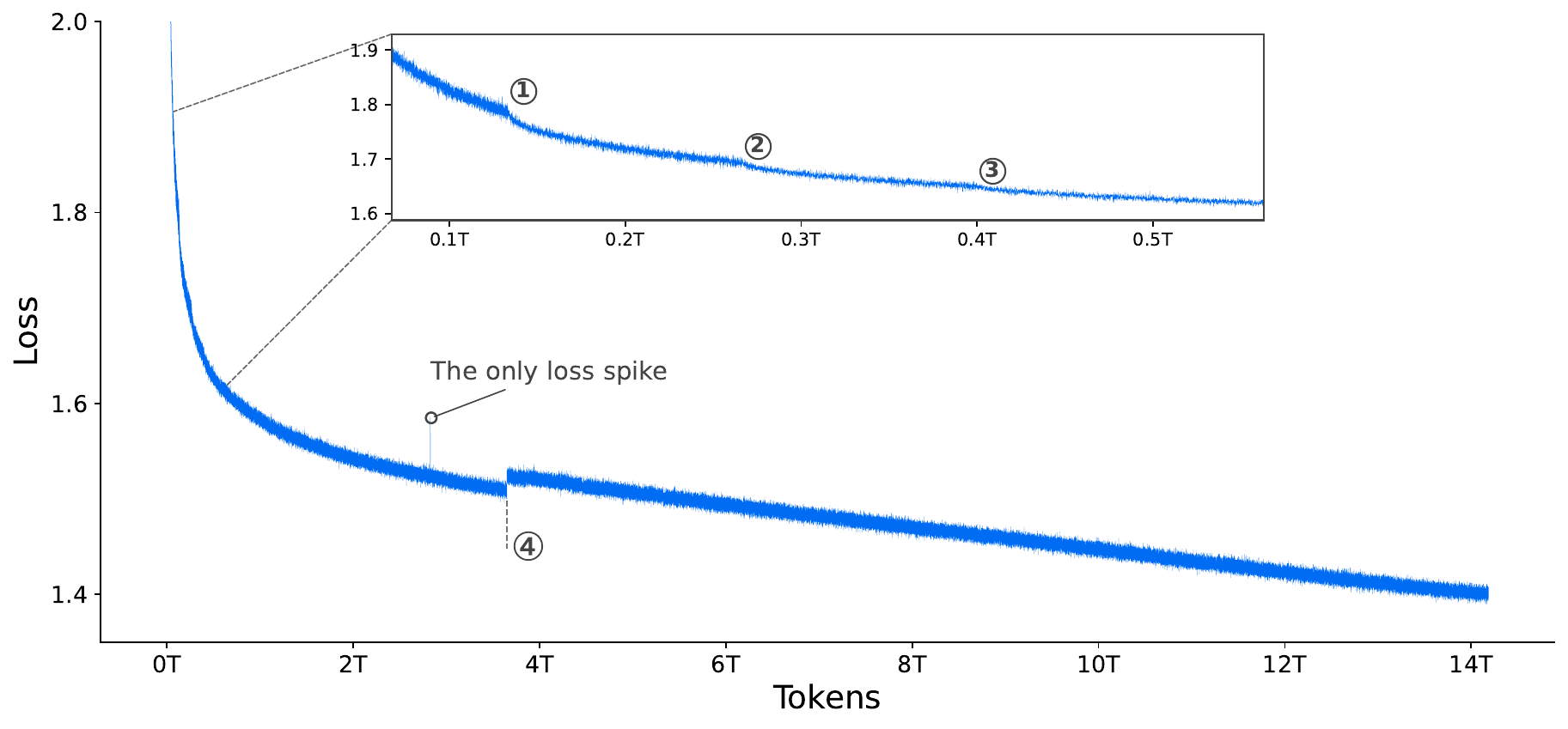}
    \caption{Per-step training loss of \ourmodel, plotted \textbf{without smoothing or sub-sampling}. We observe merely \textbf{one} isolated loss spike across the full training duration. The initial training steps are omitted for clarity. Markers \textbf{\large \ding{172}}--\textbf{\large \ding{174}} indicate batch size increases to 8,192, 12,288, and 16,384, respectively. Marker \textbf{\large \ding{175}} denotes the activation of the loss mask on meta tokens (see Appendix \ref{app:meta} for details).}
    \label{fig:loss-fig}
\end{figure}

Training stability is a \textbf{first-class} requirement for large-scale sparse MoE pre-training.
To make stability actionable, we build a comprehensive observability and diagnostic stack based on a lightweight asynchronous metrics server with micro-batch-level continuous logging (described in Section~\ref{sec:log}).
This infrastructure provides fine-grained visibility into both optimizer-level and expert-level signals, enabling systematic mitigation of recurring failure modes in large-scale MoE training.

In practice, we find three dominant instabilities that the metrics stack helps surface early and localize precisely:
(i) transient loss spikes and occasional stochastic numerical blow-ups caused by Muon’s~\cite{jordan2024muon} numerically sensitive polar-factor iteration under reduced precision,
(ii) expert-side collapse ("dead experts") that can occur even when router dispatch statistics remain apparently healthy, and
(iii) localized activation blow-ups confined to a small subset of experts.

With the mitigations guided by these diagnostics, the pre-training loss remains smooth throughout the run, exhibiting only a single loss spike. Figure~\ref{fig:loss-fig} shows the full curve prior to learning-rate cooldown.

\subsubsection{Numerical Sensitivity of Muon}

Muon approximates a semi-orthogonal update direction via a Newton--Schulz (NS) iteration~\cite{bernstein2024oldoptimizernewnorm}. In early experiments, we find modest, consistent loss reduction when using a faster-converging orthogonalization approximation. We therefore adopt the {Polar Express}~\cite{amsel2025polarexpress} iteration and run a fixed $T{=}6$ steps to balance optimization quality and throughput.

However, we occasionally observe sharp, unrecoverable loss spikes despite using the recommended safety scaling~\cite{amsel2025polarexpress}. The spikes are non-deterministic (often avoided by resuming from a nearby checkpoint),
suggesting a numerical pathology. Simulations indicate that \texttt{bfloat16} Polar Express can rarely yield extreme intermediate outliers under certain update statistics due to cumulative error in addition. We therefore cast \textbf{only} the Polar Express iteration (state and intermediates) to \texttt{float16} while keeping the rest of the training mixed-precision. After this change, the spikes do not recur.

\subsubsection{Expert Collapse Beyond Routing Collapse}

Step-3~\cite{step3blog}, our prior work, reports that MoE training may exhibit "dead experts", often described as experts receiving
negligible token dispatch for extended periods and therefore obtaining little effective gradient signal. In our prior investigation, we find that expert collapse can also manifest
as an \emph{expert-side} pathology even when router dispatch remains stable, \ie, vanishing expert activations and
stagnant or decaying expert parameter norms.

We observe that two factors are particularly influential: (i) Routed-expert aggregation requires explicit scaling. When incorporating a shared expert, it is
	    important to introduce an explicit scaling factor to calibrate the relative contribution of the shared expert and
	    the routed experts. While smaller models may implicitly learn such a balance, larger models are less reliable at
    self-calibration. A mismatch can suppress the effective contribution of routed experts even if routing frequencies
    appear healthy.
(ii) {Micro-batch balancing can be overly restrictive under fine-grained sparsity.} For sparse,
    fine-grained MoE designs, micro-batch-level load-balancing constraints (as commonly implemented in
    Switch-style routing \cite{fedus2021switchtransformers}) can become overly stringent. As analyzed in
    \cite{qiu2025demonsdetail}, micro-batch LBL may induce excessive cross-expert competition and hinder effective specialization.

We therefore prefer broader-scope balancing (\eg, global-batch statistics)~\cite{qiu2025demonsdetail,yang2025qwen3} or loss-free bias adjustment based on observed load~\cite{wang2024lossfreebalancing,deepseek2024deepseekv3}.
In practice, router dispatch statistics are typically stable and are not sensitive indicators of expert collapse.
We recommend monitoring {expert-side} signals, including per-expert
	activation norms (\eg, RMS/mean norm at the MoE FFN intermediate) and parameter norms (\eg, Frobenius
norms of expert projection matrices).
When a subset of experts drifts toward near-zero activations/updates while the median remains stable (\eg, decreasing min-to-median ratios), it provides an early warning of expert “death”.

\definecolor{weightblue}{HTML}{aec9f7}
\definecolor{basegray}{HTML}{777776}
\definecolor{actblue}{HTML}{006CF4}

\subsubsection{Localized Activation Blow-up in MoE Layers}\label{sec:clip}
As expert specialization matures during the main training phase, we observe a localized stability pathology in the deeper MoE layers. Specifically, the activation \textbf{norm} of a small subset of experts (often just one or two per layer) grows rapidly, while the majority of experts in the same layer remain well-behaved. This disparity results in a heavy-tailed activation distribution: the median expert activation norm remain stable, but the maximum activation norm explodes, significantly increasing the risk of numerical overflow and downstream instability.

\begin{figure}[ht]
    \centering
    \includegraphics[width=\linewidth]{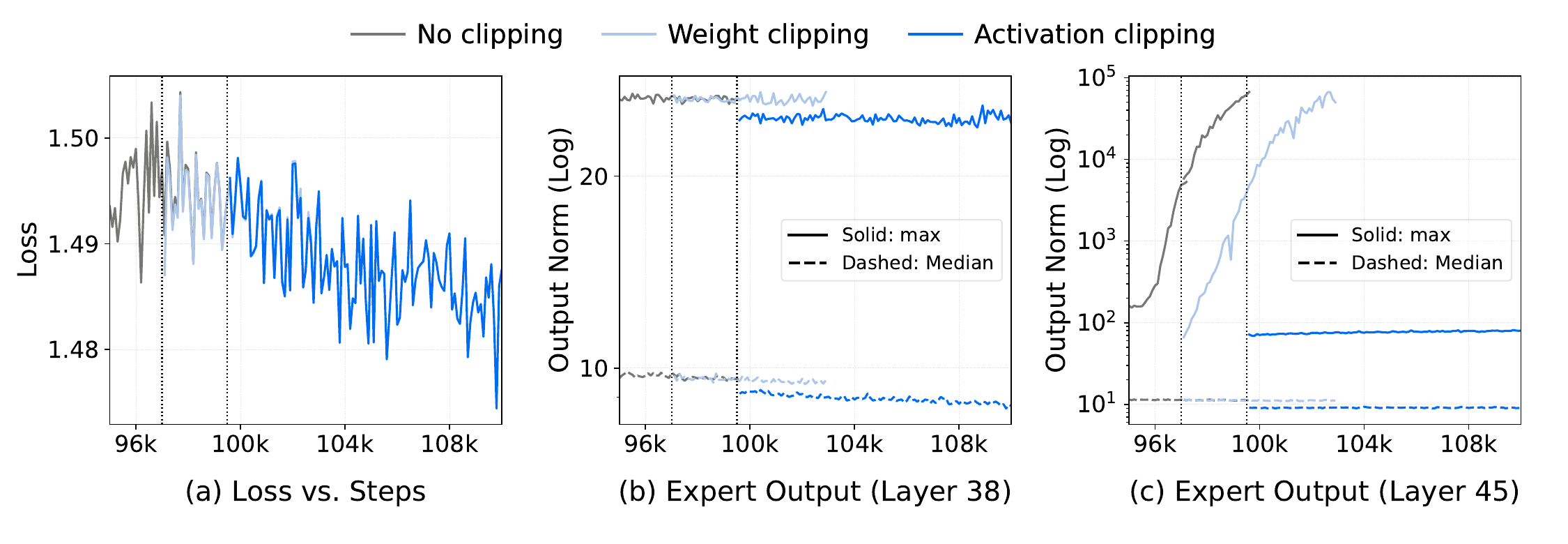}
\caption{{Analysis of expert activation stability and mitigation strategies.}
In Panels (b)--(c), solid lines represent the maximum expert output norm, while dashed lines represent the median.
{(1) Depth-Dependent Instability:} While training loss appears identical across methods (Panel a) and middle layers remain stable (\eg, Layer 38 in Panel b), the final layers (\ie, Layer 45 in Panel c) suffer from catastrophic norm explosion in the \textit{No clipping} baseline.
{(2) Mitigation:} \textit{Weight clipping} merely delays this explosion. In contrast, \textit{Activation clipping} effectively bounds maximum norms, ensuring stability across all layers.
\label{fig:moe_clip_triptych}}
\end{figure}

\noindent Figure~\ref{fig:moe_clip_triptych} illustrates this failure mode. \textbf{Remarkably, this internal instability is entirely masked by the training loss}, which shows negligible variation despite the underlying explosion in norms shown in Panel (a). We track this phenomenon by monitoring the dispersion of per-expert FFN output norms. As observed in Panels (b) and (c), while the middle layers (\eg, Layer 38) retain stable distributions, the final layers (\ie, Layer 45) exhibit a rapidly widening gap between the maximum (solid lines) and the median (dashed lines). This indicates that activation energy is concentrating dangerously in a few "rogue" experts in the deeper network. To mitigate this, we evaluate two distinct interventions:

\begin{itemize}
    \item {Weight clipping on expert projections:} We constrain the norm of the MoE FFN expert projection matrices. For each expert projection matrix $W$, if its maximum activation norm $\max_x\lVert Wx \rVert$ exceeds a threshold $\tau$, we rescale it via $W \leftarrow W \cdot \frac{\tau}{\max_x\lVert Wx \rVert}$. This is similar to MuonClip in attention~\cite{kimi_k2}, but we perform clipping offline on the checkpoint rather than on-the-fly.
    \item {Activation clipping inside experts:} We apply element-wise clipping directly to the MoE FFN intermediate activations prior to the output projection, as in \cite{openai2025gptoss120bgptoss20bmodel}.
\end{itemize}

\noindent Although the training loss appears indistinguishable across different mitigation strategies in Figure~\ref{fig:moe_clip_triptych} (a), the max-to-median ratio reliably unmasks underlying instability. As evidenced in Panels (b) and (c), activation clipping ensures a stable trajectory for internal norms, whereas weight clipping alone fails to prevent the recurrence of outlier experts. Consequently, we establish the max-to-median ratio of per-expert activation norms as a robust and necessary metric for monitoring training stability.

The activation blow-up is driven by several factors. We observe that high-frequency bi-grams can trigger expert specialization. When using pre-norm~\cite{radford2019language,xiong2020layer}, a single expert can amplify its output boundlessly and dominate the final output norm, leading to near-deterministic prediction behavior. This risk is exacerbated by SwiGLU~\cite{shazeer2020gluvariantsimprovetransformer}, where strong alignment between the gate and up-projection branches produces sparse activations with extreme magnitudes. Muon further accelerates this collapse by amplifying persistent low-rank updates. A detailed analysis is provided in Appendix~\ref{app:clip}.

\subsection{Training Curriculum}
\label{sec:training_curriculum}

The training proceeds from broad open-domain coverage to increasingly agentic and long-context specialization. We first {pre-train} at {4k} context on a broad open-domain mixture to establish general-purpose capabilities, then {anneal} the mixture toward higher-quality knowledge and more  software-development data (code, PRs, issues, and commits) while extending the context window to {32k} .
Next, a dedicated mid-training stage expands the context window from 32k to 128k to strengthen long-horizon reasoning and improve initialization for downstream post-training and agentic workloads.
Overall, we train on approximately 17.6T tokens for pre-training and 750B tokens for mid-training.

\subsubsection{Data Mixture}
\label{sec:training_data}

\noindent Our corpus combines general open-domain data with agentic-oriented data.
We summarize the key sources below, more details can be refered in Appendix~\ref{sec:step-data}.

\paragraph{General Knowledge Data.}
To support broad world knowledge, we build \textbf{StepCrawl} (Appendix~\ref{subsubsec:stepcrawl}), an in-house crawling and curation infrastructure beyond standard Common Crawl~\cite{commoncrawl}, to harvest trillions of high-quality tokens at scale from web pages (HTML) and book-/document-like sources (ePub/PDF).
All content is processed with multi-stage quality filtering, site/category tagging, deduplication, and sanitization.

\paragraph{Code Data.}
Strong code capacity is foundational for agentic models.
Our code corpus is curated and refined using a modified OpenCoder~\cite{huang2025opencoderopencookbooktoptier} pipeline.
We relax filtering from a zero-tolerance policy to allowing 0--6 heuristic violations (Appendix \ref{sec:pure_code}) per document, balancing quality and diversity, and upsample code-centric data during annealing and mid-training to strengthen agent-related programming.

\paragraph{PR/Issue/Commit Data.}
To better match real software-engineering workflows, we curate a comprehensive {PR/Issue/Commit} dataset(Appendix~\ref{sec:pr_issue_data}) from GitHub repositories with 10+ stars.
This includes (1) {Base Data} validated against \texttt{git diff} (deduplicated against benchmarks~\cite{jimenez2024swebench,yang2025swesmith});
(2) {PR-Dialogue Data} derived from PR threads and commits using Agentless-style templates~\cite{xia2024agentless} for file localization and code repair; and
(3) derivative software-engineering corpora used in mid-training and post-training.

\paragraph{Tool-Use and Reasoning Data.}
To improve tool-use robustness and multi-step reasoning, we add synthetic and semi-synthetic data spanning math/code/science/general knowledge, and domain-specific samples targeting search agent, SWE agent, and tool execution.
During mid-training, we further introduce long-context samples (natural long documents and long-form synthetic tasks) to reinforce planning and reasoning over extended contexts.

\subsubsection{Schedule}
\label{sec:schedule}

\paragraph{Pre-training schedule.}
Pre-training consists of two stages:
\begin{enumerate}[leftmargin=*,nosep]
    \item \textbf{Pre-training Stage 1: Open-domain pre-training} ({14.6T} tokens, {4k} context).
    Broad open-domain training to maximize coverage and foundational capability.

    \item \textbf{Pre-training Stage 2: Annealing + long-context initialization} ({3T} tokens, {4k to 32k} context).
    We anneal the data mixture toward code and PR/Issue/Commit-centric sources, while increasing the share of higher-quality knowledge and reasoning-dense samples.
    This stage starts with {2T} tokens at 4k context, then transitions to {1T} tokens at 32k context under the same annealed mixture to initialize long-context training.

\end{enumerate}

\paragraph{Mid-training schedule.}
Mid-training also consists of two stages:
\begin{enumerate}[leftmargin=*,nosep]
    \item \textbf{Mid-training Stage 1: Specialization at 32k} ({386B} tokens, {32k} context).
    We replay {81B} tokens (21\%) from pre-training to mitigate distribution shift and stabilize specialization, while emphasizing software-engineer and tool-use-centric mixtures.

    \item \textbf{Mid-training Stage 2: Long-context specialization} ({364B} tokens, {128k} context).
    We retain {10.5B} replay tokens, and further specialize long-context capability with a mixture of synthetic long-horizon reasoning and natural long documents (selected from pre-training data with length $>32$k), plus domain-specific data for code agent, search agent, and tool-use.
\end{enumerate}

\subsubsection{Hyper-Parameters}
\label{sec:hyperparams}

\paragraph{Pre-training hyper-parameters.}
We use the Muon optimizer~\cite{jordan2024muon} throughout pre-training, set weight decay to 0.1 and gradeint clip to 1.0.
The learning rate is linearly warmed up from 0 to $2.5\times10^{-4}$ over the first 2{,}000 steps and then cosine-decayed to $5\times10^{-5}$ over {Pre-training Stage~1}.
In {Pre-training Stage~2}, we apply a secondary cosine decay from $5\times10^{-5}$ to $2\times10^{-5}$ over the 4k portion (2T tokens) and keep the learning rate fixed at $2\times10^{-5}$ for the 32k portion (1T tokens).
The global batch size gradually increases from 4096 to 16384 over the first 400B tokens, and keeps 16384 in the remaining training, and is set to 2k for the 32k portion of annealing.
The MTP loss weight is set to 0.3 in {Pre-training Stage~1} and 0.1 in {Pre-training Stage~2}, following~\cite{deepseek2024deepseekv3}.
For loss-free load balancing, the bias update rate is 0.001 for the first 14.6T tokens and decays to 0.0 during annealing, and an EP-group balance loss with coefficient 0.001 is applied throughout pre-training.
For RoPE~\cite{su2024roformer}, we use $\theta=10{,}000$ for both full attention and sliding window attention (SWA) during 4k training, and set $\theta_{\mathrm{Full}}=1{,}000{,}000$ only for full attention and maintain $\theta_{\mathrm{SWA}}=10{,}000$ for the 32k portion of annealing.

\paragraph{Mid-training hyper-parameters.}
We continue to use Muon~\cite{jordan2024muon} during mid-training.
We freeze the MoE router weights and disable the EP-group balance loss and fix the MTP loss weight to 0.1 for both mid-training stages.
The learning rate is warmed up from 0 to $2\times10^{-5}$ over the first 3\% of iterations, kept constant in {Mid-training Stage~1}, and decayed to $7.3\times10^{-6}$ in {Mid-training Stage~2}.
For RoPE selective scaling, we set $\theta_{\mathrm{Full}}=1{,}000{,}000$ at 32k ({Mid-training Stage~1}) and increase to $\theta_{\mathrm{Full}}=5{,}000{,}000$ at 128k ({Mid-training Stage~2}), while keeping $\theta_{\mathrm{SWA}}=10{,}000$ throughout mid-training~\cite{xiong2024effective}.

%% file: src/03_post_training.tex
\section{Post-Training}
\label{sec:post-train}

In this section, we introduce a unified post-training recipe for large-scale Reinforcement Learning (RL), which begins with a unified Supervised Fine-Tuning (SFT) model. This framework enables consistent self-improvement by combining verifiable reward signals with human preference feedback, while maintaining stability even during large-scale off-policy training for Mixture-of-Experts (MoE) models.
The process follows a two-phase approach similar to prior works~\cite{liu2025deepseek, zeng2025glm}. First, we construct \textbf{Expert Models} by enhancing the unified SFT baseline with domain-specific RL across Math, Code, STEM, Tool-use, Long Context Understanding, Human Preference, and Agentic Reasoning. These specialized experts are then distilled into a generalist model using \textbf{Self-Distillation} and \textbf{Scalable RL}, ensuring the final model remains competitive with specialized baselines across diverse tasks. By systematically alternating between targeted specialization and broad synthesis, we achieve robust generalization without compromising expert-level performance.

\subsection{Expert Model Construction and Self-Distillation}
\label{sec:sft}

We employ a two-stage SFT pipeline to build a robust foundation for subsequent RL.
The first stage executes large-scale multi-domain SFT spanning Math, Code, STEM, Logic, General QA, Code Agent, Tool-use, Search Agent, and Long Context Understanding. Difficulty-aware filtering and strategic balancing are applied to foster broad agentic behaviors.
The second stage explicitly maximizes reasoning density by injecting out-of-distribution (OOD) signals~\cite{ye2025limoreasoning, muennighoff2025s1}, comprising $\sim$30k expert-level chemistry trajectories and synthetic arithmetic tasks. This targeted exposure to distinct reasoning patterns unlocks latent capabilities within just three epochs, equipping the model with the sophisticated structural complexity necessary to initialize the subsequent domain-specific RL phase.

Following domain-specific RL, we consolidate the divergent expert capabilities into a unified student model, initialized from the mid-train checkpoint. In this phase, the expert models generate high-quality trajectories using a prompt distribution shared with the first-stage SFT corpus, offering a more stable and efficient alternative to direct RL integration. This approach employs rejection sampling to eliminate undesirable patterns such as language mixing or overthinking, centralizing expert knowledge into a single student model. By establishing this high-quality foundation, self-distillation significantly reduces the optimization burden on subsequent RL stages.

\paragraph{Hyper-Parameters.}
The Muon optimizer~\cite{jordan2024muon} is employed with a $3\%$ warmup and a cosine decay from $1.0 \times 10^{-5}$ to $5.0 \times 10^{-6}$.
We freeze the MoE router weights and disable the EP-group balance loss similar to mid-training.
The SFT training is executed with an MTP loss weight of 0.1, a global batch size of 32, and a global sequence length of $128\text{k}$. Regarding Rotary Position Embeddings (RoPE)~\cite{su2024roformer}, we maintain $\theta_{SWA}=10,000$ and adjust $\theta_{Full}=5,000,000$ to accommodate the 128k context length~\cite{xiong2024effective}.

\subsection{Scalable RL}
\label{sec:rl}

\begin{figure}[t!]
    \centering
    \includegraphics[width=1.0\linewidth]{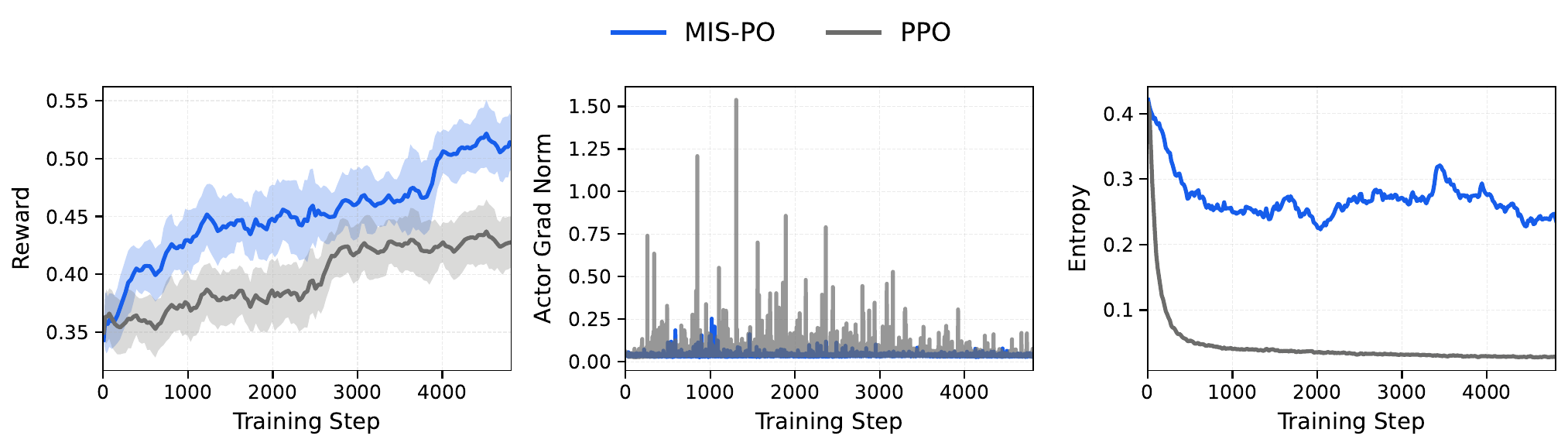}
    \caption{\textbf{Scalability comparison between MIS-PO and PPO on our internal model}.
    \textbf{(1) Efficiency}: MIS-PO demonstrates superior sample efficiency, achieving higher reward plateaus with an accelerated convergence trend.
    \textbf{(2) Stability}: MIS-PO significantly stabilizes training dynamics by suppressing gradient noise and eliminating the large spikes in the policy gradient norm.
    \textbf{(3) Exploration Persistence}: MIS-PO exhibits slower entropy decay, enabling a better exploration–exploitation balance.}
    \label{fig:mispo_vs_ppo}
\end{figure}

In RL for LLMs, we optimize a policy $\pi_\theta$ to maximize terminal rewards over trajectories $\tau = (s_0, a_0, \dots, s_T)$, where $a_t$ denotes the token generated at state $s_t$.
For reasoning tasks, however, this process faces severe instability arising from high gradient variance, further amplified by extremely long horizons and model scale (Figure~\ref{fig:mispo_vs_ppo} (2)).
This variance primarily from \textbf{infrastructure divergence} between high-throughput inference engines and training frameworks, as well as the \textbf{off-policy misalignment} inherent to iterative updates.
In such settings, importance sampling is inherently unstable, as minor token-level probability shifts compound into noisy gradients that impede convergence.

\subsubsection{MIS-Filtered Policy Optimization (MIS-PO)}
To address these stability challenges, we propose MIS-PO, a method inspired by Metropolis Independence Sampling (MIS)~\cite{metropolis1953equation,hastings1970monte}.
We treat the inference policy as a proposal distribution and the training policy as the target, restricting updates to samples that remain sufficiently close to the target distribution.
Unlike importance sampling, which scales gradients by bounded ratios and often suffers from high variance, MIS-PO applies binary masking to filter off-distribution samples and treats retained trajectories as effectively on-policy, resulting in significantly reduced gradient variance and stable optimization.

Formally, we define a binary indicator function $\mathbb{I}(x) = \mathbb{1}[\rho_{\min} \le x \le \rho_{\max}]$ and apply it at two distinct granularities. At the \textbf{token level}, the function filters the probability ratio $x_t = \pi_{\theta_{\text{old}}}(a_t|s_t) / \pi_{\theta_\text{vllm}}(a_t|s_t)$ to suppress localized mismatches between the training and inference policies~\cite{yao2025offpolicy}. At the \textbf{trajectory level}, we apply the same indicator to the geometric mean ratio $\bar{\rho}(\tau) = (\prod_t x_t)^{\frac 1T}$, effectively discarding entire trajectories that have drifted significantly from the target distribution.
The reformulated actor loss replaces continuous importance weights with these dual-level discrete masks:
\begin{equation}
    \mathcal{L}_{actor} = - \mathbb{E}_{\tau \sim \pi_{\theta_\text{vllm}}} \left[ \mathbb{I}(x_t) \cdot \mathbb{I}(\bar{\rho}(\tau)) \cdot \log \pi_\theta(a_t|s_t) \cdot \hat{A}_t \right].
    \label{eq:mispo_loss}
\end{equation}

By treating valid samples as effectively on-policy, this objective substantially reduces gradient variance for long-horizon reasoning tasks under a trust-region constraint.
Figure~\ref{fig:mispo_vs_ppo} presents an ablation study over approximately 5,000 training steps, where MIS-PO exhibits significantly lower noise in the actor gradient norm than PPO, indicating improved scalability. More ablations are shown in Appendix~\ref{sec:rl_ab}.

To further stabilize training dynamics, we employ several techniques: \textbf{Truncation-Aware Value Bootstrapping}~\cite{pardo2022timelimitsreinforcementlearning} to correct the ambitious reward bias introduced by context-length truncation and \textbf{Routing Confidence} monitoring to predict instability specific to MoE architectures.

\paragraph{Truncation-Aware Value Bootstrapping.}
Assigning zero rewards to context-truncated trajectories conflates truncation with task failure. This ambiguity penalizes long-chain reasoning by failing to distinguish between incomplete and incorrect outcomes.
To address this, we replace the zero reward with a bootstrapped value estimate of the final state, effectively treating truncation as a horizon interruption rather than a terminal failure. The modified reward for trajectory $\tau_i$ is defined as:
\begin{align}
    \hat{R_i} &= \begin{cases} V_\phi(s_T) & \text{if the response is truncated,} \\ R_i & \text{otherwise}. \end{cases}
\end{align}

Empirically, this truncation-aware value bootstrapping stabilizes training even at truncation rates as high as 20\%, preventing the reward degradation typically triggered by incomplete trajectories~\cite{deepscaler2025,yu2025dapo}. Ablation studies confirm that this technique is particularly beneficial for competition-level benchmarks, where long-horizon reasoning makes truncation effects most prevalent.

\paragraph{Routing Confidence as a Stability Proxy.}

Recent studies~\cite{zheng2025group,ma2025stabilizing} bridge RL stability with MoE routing consistency.
Building on this, we propose the \textbf{Routing Confidence} ($\Sigma_k$) as a proxy for stability, which is the average probability mass of activated experts.
Low $\Sigma_k$ implies high routing uncertainty, which amplifies the training-inference mismatch.
Through preliminary experiments, we identify a distinct stability phase transition: models with low routing confidence are brittle and require extreme stabilization (\textit{e.g.}, Router Replay~\cite{zheng2025group,ma2025stabilizing,deepseekai2024deepseekv32}, strict on-policy updates~\cite{hu2025openreasonerzero}).
In contrast, models with high routing confidence maintain robustness, enabling off-policy training without complex interventions.

\begin{figure}[t!]
    \centering
    \includegraphics[width=1.0\linewidth]{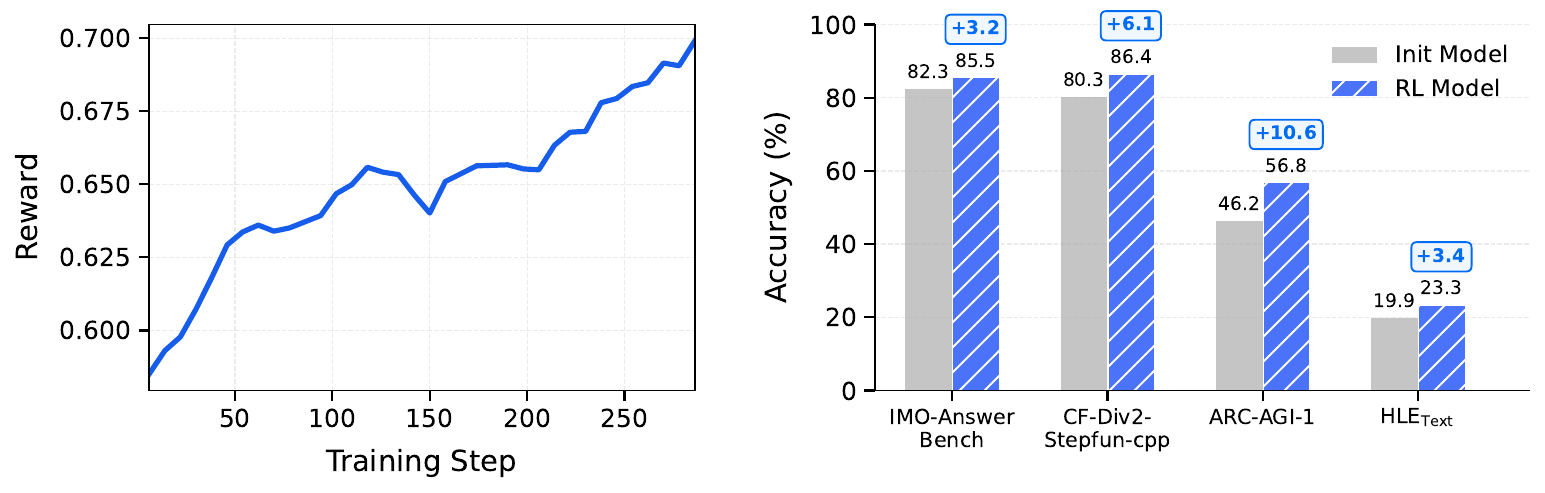}
    \caption{\textbf{RL training dynamics and cross-domain improvements of \ourmodel.} RL drives steady reward growth (left) and delivers consistent accuracy boosts across multiple benchmarks (right).}
    \label{fig:rlvr_curve}
\end{figure}

\paragraph{RL Training Dynamics.}
To provide a holistic view of our method, we illustrate the RL with verifiable rewards (RLVR) training dynamics and downstream evaluation improvements of \ourmodel in Figure~\ref{fig:rlvr_curve}. The steady rise in training rewards suggests a stable and effective learning process. Furthermore, \ourmodel achieves consistent performance gains across diverse evaluation benchmarks. Specifically, we observe substantial improvements of +3.2\% on IMO-AnswerBench~\cite{luong2025towards}, +6.1\% on CF-Div2-Stepfun-cpp~(Appendix \ref{cf-div2-step}: our custom CodeForces\footnote{https://codeforces.com/} Div.2 Benchmark), +10.6\% on ARC-AGI-1~\cite{chollet2019measure}, and +3.4\% on $\text{HLE}_\text{text}$~\cite{phan2025humanitysexam}.

\subsubsection{Reward System}
We decouple the RL framework into RL with verifiable rewards (RLVR~\cite{grporlvr}) and RL with non-verifiable rewards (\textit{\textit{e.g.}}, RLHF~\cite{openairlhf} ), each supported by a distinct reward tailored to its supervision characteristics.

\paragraph{Verifiable Rewards.}
For RLVR, each prompt is paired with a task-specific verifier that outputs a reward. The rule-based checkers are used for logic, instruction following, and code, while model-based verifiers are employed for STEM tasks.
In ablation studies over 450 RL training steps on our internal model, using model-based verifiers for STEM tasks outperforms direct vanilla math-verify by an average of 2.0\%; additional details are provided in Appendix~\ref{sec:rl_reward}.

\paragraph{Non-Verifiable Reward.}
We address non-verifiable tasks using a pairwise generative reward model (GenRM~\cite{genrm}) that benchmarks responses against a fixed reference.
GenRM is a reasoning model that outputs a confidence score indicating the likelihood of a response winning.
This score is subsequently converted into a Bradley–Terry win rate~\cite{bradley1952rank} to serve as the reward signal.
Length control is modeled within GenRM as a confidence score penalty and propagated to the win-rate reward, effectively suppressing excessive length growth during RL training.
We further ensure robustness by assigning zero reward to responses with fabricated citations, overconfident claims, or language inconsistencies.

\paragraph{Agent Reward.}
Search tasks are evaluated using an LLM based on entity-matching scores.
For report generation, a rubric-based LLM judge evaluates the research query, rubric specifications, and candidate reports, producing ternary judgments (\emph{satisfied}, \emph{partially satisfied}, \emph{unsatisfied})~\cite{hu2025stepdeepresearchtechnicalreport}. As the intermediate category often misaligns with expert preferences, we map the outputs to asymmetric binary rewards, yielding clearer learning signals and faster convergence toward expert-aligned behaviors.

\paragraph{GenRM Training and MetaRM.}
We initialize the GenRM by fine-tuning our SFT model with RM-specific prompts.
For RL training, we use curated pairwise preference data with a logsigmoid loss similar to the scalar reward model formulation.
To improve the robustness of GenRM, we penalize responses exhibiting spurious reasoning (\textit{i.e.}, correct preference derived from flawed logic) by integrating MetaRM, an additional verifier that reduces the training reward when such patterns are detected.
In ablation studies spanning 200 RL training steps on our internal model, MetaRM-augmented GenRM outperforms vanilla GenRM by 0.5\% - 3\% on every benchmark.

\subsubsection{Hyper-Parameters}
\label{sec:rl_hyperparameter}
For rollout, we set both the sampling temperature and top-$p$ to 1.0 with a maximum sequence length of 128k tokens.
Per generation, we sample 256 unique prompts with 16 responses each for reasoning tasks, 512 unique prompts with 8 responses each for human preference tasks, and 128 unique prompts with 8 responses each for tool-use tasks.
After rollout, completed samples are partitioned into mini-batches and used for training over a single epoch, with 4 mini-batches for the actor and 12 mini-batches for the critic. Optimization is performed using the Muon optimizer with a weight decay of 0.1. The actor is trained with a learning rate of $2 \times 10^{-6}$ and 20 warmup steps, while the critic uses a learning rate of $5 \times 10^{-6}$ with 50 warmup steps. Following ORZ~\cite{hu2025openreasonerzero}, we set both $\gamma$ and $\lambda$ to 1. We further adopt an unbiased KL loss~\cite{liu2025deepseek} with a coefficient of 0.001 in the final stage. For Equation~(\ref{eq:mispo_loss}), the token-level and trajectory-level masking bounds are set to $[0.5, 2]$ and $[0.996, 1.001]$, respectively.

\subsection{Data Synthesis \& Curation}
We construct a diverse and difficulty-balanced prompt pool by aggregating open-source data, synthetic generations, and user trajectories. A unified synthesis and curation pipeline is applied, combining strict global filtering with domain-specific refinement to maximize reasoning density. Data quality is ensured through a hybrid of rule-based heuristics and model-based fidelity checks. The resulting dataset contains 871k samples (7.23B tokens), with detailed statistics summarized in Table \ref{tab:sft_data_statistics}.

\begin{table}[t!]
    \centering
    \setlength{\tabcolsep}{4.0pt}
    \begin{tabular}{lrrr}
    \toprule
    \rowcolor{header_gray} \textbf{Domain} & \textbf{Num Samples} & \textbf{Tokens} & \textbf{Corpus Contribution} \\
    \midrule
    Math    & 68055 & 0.98B & 11.19\% \\
    Code    & 86421 & 1.23B & 21.10\% \\
    STEM    & 120399  & 0.55B & 6.31\% \\
    Logic   & 93323  & 0.81B & 13.87\% \\
    General & 314495 & 0.80B & 9.16\% \\
    Code Agent     & 37240 & 0.90B & 17.70\% \\
    Tool-use & 114507 & 0.76B & 8.72\% \\
    Search Agent
    & 20256 & 0.50B & 8.75\% \\
    Long Context & 15565 & 0.70B & 4.00\% \\
    \midrule
    Total   & 870687 & 7.23B & 100.00\% \\
    \bottomrule
\end{tabular}
\caption{Data Statistics of first-stage SFT.}
\label{tab:sft_data_statistics}
\end{table}
\subsubsection{General and Reasoning}
Our training corpus aggregates community prompts, expert responses, and synthetic data from diverse open-source, including \textbf{Mathematics}~\cite{li2024numinamath,albalak2025bigmath,mitra2024orcamath,aslawliet2024olympiads,aslawliet2024cnk12,openr12025math220k,deepmath2025deepmath103k,yu2025dapoopensourcellmreinforcement,hu2025openreasonerzero,guha2025openthoughts,muennighoff2025s1simpletesttimescaling,ji2025amthinkingv1advancingfrontierreasoning,limo2025lessismore}, \textbf{Coding}~\cite{li2023tacotopicsalgorithmiccode,deepcoder2025,wang2025codecontestshighqualitytestcase}, and \textbf{Science and Open-ended QA}~\cite{li2023camel,bercovich2025llamanemotronefficientreasoningmodels,fan2025megascience,zhao2024wildchat}.
To maximize reasoning density, we employ a unified pipeline that couples strict global filtering with domain-specific refinement, enforcing quality via a hybrid of rule-based heuristics and model-based fidelity checks. Specifically, in mathematics, we ensure numerical stability through specialist-guided rejection sampling and synthetic large-number arithmetic. For programming, we prioritize offline executability by selecting rigorous algorithmic challenges while strictly purging RAG-related hallucinations.
In particular, we mitigate the model's tendency to falsely claim access to external search engines or pretend to retrieve online solutions. Furthermore, we restrict scientific data to unambiguous questions with unique, determinable solutions.

To enable generalization across practical scenarios, we expand open-source checkers\footnote{https://github.com/allenai/open-instruct/tree/main/open\_instruct/IFEvalG} and augment samples with several real-world constraints.
In parallel, we collect general prompts from open-source, synthetic, and user trajectories to form a diverse, difficulty-balanced pool.
This process yields a high-fidelity dataset comprising millions of samples at the billion-token scale.

\subsubsection{Generalized Tool Learning}
We propose an execution-driven data generation framework for learning reliable tool-use behaviors in intelligent agents, addressing key limitations of existing synthetic pipelines such as data inconsistency, lack of verifiability, and model hallucinations. Instead of relying on random exploration~\cite{liu2025webexplorer,li2025websailor} or model-based simulation~\cite{kimi_k2,li2025simulating}, our approach decomposes tool-use behavior into atomic intents and models them using a finite state machine (FSM), explicitly separating abstract tool-call logic from parameterized execution constraints. Data is generated through a sample–execute–verify loop with rejection sampling, where all candidate trajectories are executed in real environments and validated by deterministic feedback, ensuring fidelity and eliminating hallucinated behaviors. By compositionally combining atomic intents, the framework supports scalable generation of complex, controllable tool-use scenarios.
Using this paradigm, we construct over 100K high-quality trajectories totaling billions of tokens, providing precise supervision for tool-based planning, reasoning, and execution.

\subsubsection{Code Agents}
Code agents can self-improve through a closed-loop intervention between verifiable \textbf{environment construction} and \textbf{solution generation}, where executable feedback continuously refines both capabilities.
We treat environment construction as a first-class capability alongside bug fixing and feature implementation, synthesizing it under verifiable reward signals.
To this end, we develop a specialized agentic pipeline evolved from the SWE-factory~\cite{guo2026swefactoryautomatedfactoryissue} framework, incorporating a cross-task memory pool that retrieves historical build successes as few-shot demonstrations and a loop-detection mechanism to prevent redundant exploration.
This pipeline achieves a 40\% environment-building success rate, forming a positive feedback loop for model self-evolution through dense supervision from construction trajectories, including shell commands and error recovery.
To further improve signal quality, we normalize environment construction trajectories by abstracting and masking transient failures and redundant execution patterns that do not contribute to the final resolution.
The bootstrapped environments function as dynamic testbeds, leveraging execution feedback and unit tests to generate high-quality synthetic data and reward signals for continuous alignment. Empirically, we observe a bidirectional transfer: construction expertise accelerates coding performance, while coding within these environments further improves construction accuracy, as shown in DockSmith~\cite{zhang2026docksmithscalingreliablecoding}.
Leveraging this evolution pipeline, we curate 50k verified environments spanning over 15k GitHub repositories and more than 20 programming languages. This diverse collection captures a broad spectrum of real-world scenarios, providing a robust foundation for training generalist code agents. Furthermore, we incorporate several prominent open-source environments, including SWE-smith~\cite{yang2025swe}, SWE-Gym~\cite{pan2024training}, R2E-Gym~\cite{jain2025r2e}, SWE-rebench~\cite{badertdinov2025swe}, and SETA~\cite{seta}.

\subsubsection{Search and Research Agents}

To facilitate advanced information-seeking, our pipeline integrates graph-based and multi-document synthesis to enforce multi-hop reasoning. By performing topological expansions on knowledge graphs (\textit{e.g.}, Wikidata5m~\cite{wang2021kepler}) and simulating cross-website browsing trajectories, we generate data that reflects real-world research complexity. Crucially, to guarantee the necessity of external retrieval, we validate generated queries against DeepSeek-R1~\cite{Guo_2025}, systematically excluding instances solvable by this strong reasoning model without tool interaction. The resulting trajectories are refined through a structured report generation pipeline~\cite{hu2025stepdeepresearchtechnicalreport} that enforces rigorous instruction compliance and structural integrity.
Specifically, we enforce strict adherence to preset research plans, discarding any trajectories that deviate from the structure. Subsequently, valid outputs undergo iterative cleaning via model-based judgers and heuristic rules to resolve fine-grained issues such as informal writing, temporal hallucinations, and mixed-language artifacts.
This end-to-end approach achieves industry-leading performance on the \textsc{ResearchRubrics}~\cite{sharma2025researchrubrics} benchmark.

\subsection{Agent Infrastructure}
\paragraph{Reasoning with Tool-Use Template Design.}
To effectively integrate reasoning and agentic capabilities into a single foundation model, it is crucial to determine the appropriate templates for the thinking process and tool usage. Regarding the reasoning template, we evaluate three management strategies.
The approach of discarding reasoning history at every turn~\cite{Guo_2025}, while incentivizing independent generation, leads to task failure in long-horizon tasks(\textit{e.g.}, coding sessions exceeding 100 turns). Conversely, retaining the full reasoning history incurs prohibitive context consumption, which rapidly saturates the model's capacity and blocks subsequent tool invocations. To resolve this, we adopt a selective retention strategy: preserving reasoning traces exclusively for the tool-use trajectory triggered by the most recent user instruction.
This design achieves an optimal trade-off between reasoning coherence and context efficiency, a practice aligned with recent frontier models~\cite{yang2025qwen3technicalreport,liu2025deepseek}.
Regarding the tool-use template, we compared the prevalent JSON and XML formats.
The rigid syntax of JSON, including escape sequences and delimiters, frequently induces parsing errors in small, under-trained models.
In contrast, the XML format allows for flat string output with significantly lower grammatical overhead. Therefore, we select the XML format to ensure robustness in complex, real-world agentic coding scenarios.

\paragraph{Scalable Code Agent Infrastructure.}
Our integrated architecture focuses on scalable session management and cross-framework generalization to facilitate high-throughput agentic coding. Central to this is a proprietary Session-Router that orchestrates container lifecycles via Kubernetes and ensures interaction consistency through Tmux. This architecture supports thousands of concurrent environments with seamless state persistence, eliminating the need for manual, scaffold-specific Docker configurations.
To ensure high generalization across diverse agentic workflows, we trained the model to adapt to a wide spectrum of interaction frameworks, ranging from academic standards (\textit{e.g.}, OpenHands~\cite{wang2024openhands}, SWE-agent~\cite{yang2024swe}, and Terminus-2~\cite{terminalbench2025}) to enterprise grade protocols (\textit{e.g.}, Kilocode~\cite{kilocode}, Roocode~\cite{roocode}, and ClaudeCode~\cite{claudecode}). By exposing the model to these varied interaction paradigms during training, we effectively prevent it from overfitting to specific pipeline patterns, ensuring it remains robust regardless of the underlying execution environment.

%% file: src/04_evaluation.tex
\section{Evaluations}

\definecolor{glm_bg}{RGB}{235, 245, 255}
\definecolor{glm_text}{RGB}{0, 110, 220}
\definecolor{header_gray}{RGB}{242, 242, 242}
\definecolor{section_gray}{RGB}{230, 230, 230}

\subsection{Pre-training Evaluations}\label{sec:evals}

\begingroup
\renewcommand{\arraystretch}{1.1}

\input{src/tables/pre-training-main-table}

\endgroup

\paragraph{Evaluation Setup.} We evaluate {\ourmodel} on a series of benchmarks, encompassing various capabilities:
(1) General language understanding and reasoning, including BBH~\cite{suzgun2022challengingbbh}, MMLU~\cite{hendrycks2020mmlu}, MMLU-Redux~\cite{gema2024donewithmmlu}, MMLU-Pro~\cite{wang2024mmlupro}, HellaSwag~\cite{zellers2019hellaswag}, WinoGrande~\cite{sakaguchi2019winogrande}, GPQA~\cite{rein2023gpqa}, SuperGPQA~\cite{du2025supergpqa}, and SimpleQA~\cite{openai2024simpleqa}.
(2) Mathematics reasoning, including GSM8K~\cite{cobbe2021gsm8k} and MATH~\cite{hendrycks2021math}.
(3) Coding, including HumanEval \cite{chen2021evaluatinglargelanguagemodels}, MBPP \cite{austin2021programsynthesislargelanguage}, HumanEval+, MBPP+~\cite{liu2023evalplus} and MultiPL-E~\cite{cassano2022multipl}.
(4) Chinese understanding, including C-Eval~\cite{huang2023ceval}, CMMLU~\cite{li2023cmmlu}, and C-SimpleQA~\cite{he2024chinesesimpleqa}.

\paragraph{Evaluation Results.}
Table~\ref{tab:pretrain_eval_official} summarizes the pre-training evaluation of \ourmodel\ across general reasoning, mathematics, code, and Chinese benchmarks.
Despite activating only 11B parameters (196B total), \ourmodel\ remains broadly competitive with substantially larger sparse baselines (15--37B activated; 309--1043B total), demonstrating a strong accuracy--efficiency trade-off.
On core general benchmarks, \ourmodel\ achieves 88.2 on BBH (within 0.5 of the best) and 85.8 on MMLU.
Notably, \ourmodel\ reaches 31.6 on SimpleQA, outperforming DeepSeek-V3.2-Exp Base (27.0) while using only 196B total parameters versus 671B (i.e., $\sim$3.4$\times$ total parameters), highlighting stronger capability density per parameter budget.
\ourmodel\ further demonstrates strong coding capabilities, including 81.1 on HumanEval, 67.7 on MultiPL-E HumanEval and 58.0 on MultiPL-E MBPP. Overall, these results show that \ourmodel\ delivers high strong performance per activated compute, providing a solid foundation for downstream reasoning and agentic post-training.

\subsection{Post-training Evaluations}
We evaluate Step 3.5 Flash on representative benchmarks, including the reasoning oritend HLE (text subset)~\cite{phan2025hle}, MMLU-Pro~\cite{wang2024mmlupro}, GPQA-Diamond~\cite{rein2023gpqa}, AIME2025~\cite{10.5555/3524938.3525416}, HMMT~\cite{hmmt25}, IMO-AnswerBench~\cite{luong2025towards};
the coding related LiveCodeBench-v6 (2024.08-2025.05)~\cite{jain2024livecodebench}, CF-Div2-Stepfun\footnote{https://huggingface.co/datasets/stepfun-ai/CF-Div2-Stepfun}, SWE-Bench Verified~\cite{swe_verified} and SWE-Bench Multilingual~\cite{yang2025swesmith};
the agent series $\tau^2$-Bench~\cite{tau2bench2025}, Terminal-Bench 2.0~\cite{terminalbench2025},  GAIA~\cite{mialon2023gaiabenchmarkgeneralai}, BrowseComp~\cite{wei2025browsecomp}, xbench-DeepSearch~\cite{chen2025xbench}, BrowseComp-zh~\cite{zhou2025browsecompzhbenchmarkingwebbrowsing}, and \textsc{ResearchRubrics}~\cite{sharma2025researchrubrics};
the general related ArenaHard v2~\cite{arenahard2024}, IFBench~\cite{pyatkin2025generalizing} and MultiChallenge~\cite{sirdeshmukh2025multichallengerealisticmultiturnconversation}; and the long-context related LongBench v2~\cite{bai2024longbenchv2}, MRCR~\cite{vodrahalli2024michelangelo}~\footnote{https://huggingface.co/datasets/openai/mrcr}, FRAMES~\cite{krishna2025fact} and RepoQA~\cite{liu2024repoqaevaluatinglongcontext}.

We further investigate the test-time scaling properties of Step 3.5 Flash on reasoning, general, and long-context benchmarks by adopting the \textbf{Parallel Coordinated Reasoning (PaCoRe)} paradigm~\cite{hu2026pacorelearningscaletesttime}.
Leveraging Step 3.5 Flash's extreme inference efficiency,
this approach decouples reasoning capacity from context limitations by launching parallel reasoning trajectories and synthesizing their insights into higher-fidelity solutions via multi round coordination.
Specifically, we employ a multi-round PaCoRe trajectory configuration as $\vec{K} = [4,4,4,4]$, yielding significant gains across benchmarks.

We maintain a maximum sequence length of 256k, using the default decoding configuration with decoding temperature and top-p of 1.0.
And we apply YaRN~\cite{peng2023yarnefficientcontextwindow} with a scaling factor of 2.0 on top of the original 128k positional embeddings, restricting it to full-attention layers only.
We report pass@1 accuracy for all approaches based on average performance of multiple independent generations per problem: 64 for AIME 2025, HMMT 2025 Feb., and HMMT 2025 Nov.; 8 for IMO-AnswerBench, LiveCodeBench, GPQA-Diamond, and MultiChallenge; 1 for HLE and 4 runs for all other benchmarks.
More details are provided in Appendix~\ref{sec:post_training_eval_appendix}.

\paragraph{Evaluation Results.} Table~\ref{fig:post-training-main-table} presents a comprehensive comparison of Step~3.5~Flash against a broad set of strong baselines across reasoning, code agents, general agents, long-context understanding, and general capability benchmarks.
Despite activating only 11B parameters (196B total), Step~3.5~Flash demonstrates strong performance across a wide range of tasks, particularly excelling on reasoning-intensive benchmarks such as AIME~2025, HMMT~2025~Feb., HMMT~2025~Nov., IMO-AnswerBench, and LiveCodeBench-v6.
It consistently outperforms open-source models with larger parameter counts and achieves performance on par with frontier models such as GPT-5.2~xHigh and Gemini~3.0~Pro.
Notably, Step~3.5~Flash achieves strong results on agentic evaluations, including SWE-Bench Verified, Terminal-Bench~2.0, BrowseComp (with Context Manager), GAIA, and $\tau^2$-Bench, highlighting robust tool-use and long-horizon decision-making capabilities.

\begingroup
\renewcommand{\arraystretch}{1.1}
\input{src/tables/post-traininig-main-table}

\endgroup

%% file: src/tables/pre-training-main-table.tex
\begin{table*}[h]
    \centering
    \footnotesize
    \setlength{\tabcolsep}{2pt}
    \begin{tabular}{l c | c c c c c c}
        \toprule
        \rowcolor{header_gray}
        \textbf{Benchmark} & \textbf{\# Shots} &
        \makecell[c]{\textbf{\ourmodel}\\\textbf{Base}} &
        \makecell[c]{\textbf{MiMo-V2 Flash}\\\textbf{Base}} &
        \makecell[c]{\textbf{GLM-4.5}\\\textbf{Base}} &
        \makecell[c]{\textbf{DeepSeek V3.1}\\\textbf{Base}} &
        \makecell[c]{\textbf{DeepSeek V3.2}\\\textbf{Exp Base}} &
        \makecell[c]{\textbf{Kimi-K2}\\\textbf{Base}} \\
        \midrule
        \# Activated Params & - & \glmcell{11B} & 15B & 32B & 37B & 37B & 32B \\
        \# Total Params & - & \glmcell{196B} & 309B & 355B & 671B & 671B & 1043B \\
	        \midrule
	        \rowcolor{section_gray}\multicolumn{8}{l}{\textsc{\textbf{General}}} \\
	        BBH & 3-shot & \glmcell{88.2} & 88.5 & 86.2 & 88.2$\dagger$ & 88.7$\dagger$  & \best{88.7}  \\
	        MMLU & 5-shot & \glmcell{85.8} & 86.7 & 86.1 & 87.4$\dagger$ & \best{87.8}$\dagger$  & 87.8  \\
	        MMLU-Redux & 5-shot & \glmcell{89.2} & \best{90.6} & \na & 90.0$\dagger$ & 90.4$\dagger$  & 90.2  \\
	        MMLU-Pro & 5-shot & \glmcell{62.3} & \best{73.2} & \na & 58.8$\dagger$   & 62.1$\dagger$  & 69.2\\
	        HellaSwag & 10-shot & \glmcell{90.2} & 88.5 & 87.1 & 89.2$\dagger$   & 89.4$\dagger$  & \best{94.6}  \\
	        WinoGrande & 5-shot & \glmcell{79.1} & 83.8 & \na & \best{85.9}$\dagger$ & 85.6$\dagger$  & 85.3  \\
	        GPQA & 5-shot & \glmcell{41.7} & 43.5* & 33.5* & 43.1* & 37.3* & 43.1* \\
	        SuperGPQA & 5-shot & \glmcell{41.0} & 41.1 & - & 42.3$\dagger$  & 43.6$\dagger$  & \best{44.7} \\
	        SimpleQA & 5-shot & \glmcell{31.6} & 20.6 & 30.0 & 26.3$\dagger$ & 27.0$\dagger$  & \best{35.3}  \\
	        \rowcolor{section_gray}\multicolumn{8}{l}{\textsc{\textbf{Mathematics}}} \\
	        GSM8K & 8-shot & \glmcell{88.2} & \best{92.3} & 87.6 & 91.4$\dagger$ & 91.1$\dagger$  & 92.1  \\
	        MATH & 4-shot & \glmcell{66.8} & \best{71.0} & 62.6 & 62.6$\dagger$  & 62.5$\dagger$  & 70.2 \\
	        \rowcolor{section_gray}\multicolumn{8}{l}{\textsc{\textbf{Code}}} \\
	        HumanEval & 3-shot & \glmcell{81.1} & 77.4* & 79.8* & 72.5* & 67.7* & \best{84.8}* \\
	        MBPP & 3-shot & \glmcell{79.4} & 81.0* & 81.6* & 74.6* & 75.6* & \best{89.0}* \\
	        HumanEval+ & 0-shot & \glmcell{72.0} & 70.7 & \na & 64.6$\dagger$  & 67.7$\dagger$  & \na \\
	        MBPP+ & 0-shot & \glmcell{70.6} & 71.4 & \na & \best{72.2}$\dagger$ & 69.8$\dagger$  & \na \\
	        MultiPL-E HumanEval & 0-shot & \glmcell{67.7} & 59.5 & \na & 45.9$\dagger$ & 45.7$\dagger$  & 60.5 \\
	        MultiPL-E MBPP & 0-shot & \glmcell{58.0} & 56.7 & \na & 52.5$\dagger$ & 50.6$\dagger$  & \best{58.8}  \\
	        \rowcolor{section_gray}\multicolumn{8}{l}{\textsc{\textbf{Chinese}}} \\
	        C-EVAL & 5-shot & \glmcell{89.6} & 87.9 & 86.9 & 90.0$\dagger$ & 91.0$\dagger$  & \best{92.5}  \\
	        CMMLU & 5-shot & \glmcell{88.9} & 87.4 & \na & 88.8$\dagger$  & 88.9$\dagger$  & \best{90.9} \\
	        C-SimpleQA & 5-shot & \glmcell{63.2} & 61.5 & 70.1 & 70.9$\dagger$ & 68.0$\dagger$  & \best{77.6}  \\
	        \bottomrule
	    \end{tabular}
            \caption{Pre-training evaluation results. * denotes cases where the original score was unavailable; we report results evaluated under the same test conditions as \ourmodel\ for fair comparison. $\dagger$ indicates Deepseek scores quoted from the MiMo-V2-Flash report \cite{xiaomi2025mimo}.}
	    \label{tab:pretrain_eval_official}
\end{table*}

%% file: src/tables/post-traininig-main-table.tex
\newlength{\stepcolw}
\setlength{\stepcolw}{1.45cm}
\newcolumntype{S}{>{\centering\arraybackslash}m{\stepcolw}}

\newcommand{\stepcell}[1]{\cellcolor{glm_bg}\textcolor{glm_text}{\textbf{#1}}}
\newcommand{\glmtext}[1]{\textcolor{glm_text}{\textbf{#1}}}

\begin{table}[t!]
    \centering
    \resizebox{\linewidth}{!}{
    \begin{tabular}{l|SS ccccc|ccc}
        \toprule
        \rowcolor{header_gray}
        \textbf{Benchmark} &
        \multicolumn{2}{|c}{
          \makecell[c]{
                \\[0.0em]
        {\cellcolor{header_gray}\bfseries\fontsize{12pt}{14pt}\selectfont Step 3.5 Flash}\\[0.2em]
            \scriptsize
            \begin{tabular}{@{}>{\centering\arraybackslash}m{\stepcolw}@{\hspace{10pt}}>{\centering\arraybackslash}m{\stepcolw}@{}}
           \textbf{Vanilla} & \textbf{PaCoRe}
        \end{tabular}
          }
        } &
        \makecell[c]{\textbf{MiniMax}\\\textbf{M2.1}} &
        \makecell[c]{\textbf{MiMo V2}\\\textbf{Flash}} &
        \makecell[c]{\textbf{GLM}\\\textbf{4.7}} &
        \makecell[c]{\textbf{DeepSeek}\\\textbf{V3.2}} &
        \textbf{\textbf{Kimi K2.5}} &
        \makecell[c]{\textbf{Gemini}\\\textbf{3.0 Pro}} &
        \makecell[c]{\textbf{Claude}\\\textbf{Opus}\\\textbf{4.5}} &
        \makecell[c]{\textbf{GPT-5.2}\\\textbf{xHigh}} \\
        \midrule
        \# Activated params &
        \multicolumn{2}{|c}{\cellcolor{glm_bg}\textcolor{glm_text}{\textbf{11B}}} &
        10B & 15B & 32B & 37B & 32B & - & - & - \\
        \# Total params &
        \multicolumn{2}{|c}{\cellcolor{glm_bg}\textcolor{glm_text}{\textbf{196B}}} &
        230B & 309B & 355B & 671B & 1T & - & - & - \\
        \midrule
        \rowcolor{section_gray}
        \multicolumn{11}{l}{\textsc{\textbf{Reasoning}}} \\
        AIME 2025                & \glmcell{97.3} & \glmcell{99.9}  & 83.0 & 95.1* & 95.7  & 93.1   & 96.1 & 95.0 & 92.8 & \best{100.0} \\
        HMMT 2025 Feb.           & \glmcell{98.4} & \glmcell{100.0} & 71.0* & 95.4* & 97.1 & 92.5   & 95.4 & 97.5$\dagger$ & 92.9$\dagger$ & \best{99.4}\\
        HMMT 2025 Nov.           & \glmcell{94.0} & \glmcell{97.8}  & 74.3* & 91.0* & 93.5 & 90.2   & 91.1 & 94.5$\dagger$ & 91.7* & \best{97.1}* \\
        IMO-AnswerBench           & \glmcell{85.4} & \glmcell{88.8}  & 60.4* & 80.9* & 82.0 & 78.3   & 81.8 & 83.3$\dagger$ & 84.0$\dagger$ & \best{86.3}$\dagger$ \\
        LiveCodeBench-v6         & \glmcell{86.4} & \glmcell{88.9}  & 75.4* & 81.6* & 84.9 & 83.3   & 85.0 & \best{90.7}$\dagger$ & 84.8$\dagger$ & 87.7$\dagger$ \\
        CF-Div2-Stepfun-cpp         & \glmcell{86.1} & \glmcell{93.3}  & 59.0* & 46.9* & 74.1* & 81.6* & 73.6* & 83.5* & 72.2* & - \\
        MMLU-Pro                 & \glmcell{84.4} & \glmcell{84.8}  & 88.0 & 84.9 & 84.3 & 85.0     & 87.1 & \best{90.1}$\dagger$ & 89.5$\dagger$ & 87.4$\dagger$ \\
        GPQA-Diamond             & \glmcell{83.5} & \glmcell{85.0}  &  83.0 & 84.1* & 85.7 & 82.4   & 87.6 & 91.9 & 87.0 & \best{92.4} \\
        $\text{HLE}_\text{text}$ & \glmcell{23.1} & \glmcell{27.9} & 22.2 & 22.1 & 24.8 & 25.1      & 31.5 & \best{37.7}$\dagger$ & 30.8$\dagger$ & 35.5$\dagger$ \\
        \midrule

        \rowcolor{section_gray}
        \multicolumn{11}{l}{\textsc{\textbf{Code Agent}}} \\

        SWE Verified       & \glmcell{74.4} & \glmcell{\na} & 74.0 & 73.4 & 73.8 & 73.1 & 76.8 & 76.2 & \best{80.9} & 80.0 \\
        SWE Multilingual   & \glmcell{67.4} & \glmcell{\na} & 72.5 & 71.7 & 66.7 & 70.2 & 73.0 & 65.0$\dagger$ & \best{77.5}$\dagger$ & 72.0$\dagger$ \\
        Terminal-Bench 2.0 & \glmcell{51.0} & \glmcell{\na} & 47.9 & 38.5 & 41.0 & 46.4 & 50.8 & 56.9$\dagger$ & \best{59.3}$\dagger$ & 54.0$\dagger$ \\
        \midrule

        \rowcolor{section_gray}
        \multicolumn{11}{l}{\textsc{\textbf{General Agent}}} \\

        BrowseComp              & \glmcell{51.6} & \glmcell{\na} & 47.4  & 45.4 & 52.0 & 51.4     & \best{60.6} & 37.8$\dagger$ & 37.0$\dagger$ & \na \\
        BrowseComp (w. Ctx Manage)      & \glmcell{69.0} & \glmcell{\na} & 62.0  & 58.3 & 67.5 & 67.6     & \best{74.9} & 59.2$\dagger$ & 57.8$\dagger$ & 65.8 \\
        BrowseComp-ZH           & \glmcell{66.9} & \glmcell{\na} & 47.8* & 51.2* & 66.6 & 65.0    & 62.3* & 66.8* & 62.4* & \best{76.1*} \\
        GAIA                    & \glmcell{84.5} & \glmcell{\na} & 64.3* & 78.2* & 61.9* & 75.1*  & 75.9* & 76.6* & 76.1* & 83.5* \\
        xbench-DeepSearch-2505               & \glmcell{83.7} & \glmcell{\na} & 68.7* & 69.3* & 72.0* & 78.0*  & 76.7* & 78.3* & 77.0* & 83.0* \\
        xbench-DeepSearch-2510               & \glmcell{56.3} & \glmcell{\na} & 43.0* & 44.0* & 52.3* & 55.7*  & 40.0$\dagger$ & 57.7* & 59.3* & \best{67.0*}\\
        \textsc{ResearchRubrics}& \glmcell{65.3} & \glmcell{\na} & 60.2* & 54.3* & 62.0* & 55.8*  & 59.5* & 50.1* &  61.6* & 57.8* \\
        $\tau^2$-Bench          & \glmcell{88.2} & \glmcell{\na} & 86.6* & 84.1* & 87.4 & 85.2*   & 85.4* & 90.7 & \best{92.5} & 85.5* \\

        \midrule
        \rowcolor{section_gray}
        \multicolumn{11}{l}{\textsc{\textbf{General}}} \\
        Arena-Hard-v2.0 & \glmcell{74.0} & \glmcell{93.1} & 63.1* & 68.2* & 73.1* & 66.0*                         & \best{85.8}* & 81.7$\dagger$ & 76.7$\dagger$ & 80.6$\dagger$ \\
        MultiChallenge  & \glmcell{55.7} & \glmcell{60.8} & 50.5* & 44.3* & 67.8* & 57.1*                         & \best{73.6}* & 71.8* & 65.8* & 71.9* \\
        IFBench         & \glmcell{67.4} & \glmcell{56.8} & 70.0 & 64.0$\dagger$  & 68.0$\dagger$ & 61.0$\dagger$ & 72.8* & 70.4$\dagger$ & 58.0$\dagger$ & \best{75.4}$\dagger$ \\

        \midrule
        \rowcolor{section_gray}
        \multicolumn{11}{l}{\textsc{\textbf{Long Context}}} \\
        LongBench v2       & \glmcell{57.5}  & \glmcell{62.0} & 53.9* & 60.6$\dagger$ & 59.1* & 58.4$\dagger$                 & 61.0 & \best{70.0}* & 67.8* & 62.4* \\
        MRCR-8needle       & \glmcell{28.8}  & \glmcell{26.3} & 20.0$\dagger$ & 19.9$\dagger$ & 25.4$\dagger$ & 27.2$\dagger$ & 36.5* & 73.0$\dagger$ & 54.0* & \best{88.2*} \\
        FRAMES-Oracle      & \glmcell{76.5} & \glmcell{77.2} & 76.5* & 78.0* & 75.1* & 80.1*                                  & 77.4* & 79.7* & 85.8* & \best{87.3*} \\
        RepoQA             & \glmcell{88.5} & \glmcell{88.7} & 88.2* & 91.2* & 89.5* & 91.9*                                  & 89.8* & 91.5* & \best{95.7*} & 93.8* \\
        \bottomrule
    \end{tabular}}
    \caption{
    Comparison between Step 3.5 Flash and closed/open models. * denotes cases where the original score was unavailable or inferior to our reproduced result; we therefore report results evaluated under the same test conditions as Step 3.5 Flash for fair comparison. $\dagger$ indicates scores quoted from non-official sources, including technical reports, or independent evaluation platforms. Our evaluation on HLE focuses on the text-only subset. BrowseComp (w. Ctx Manage) denotes the evaluation of BrowseComp with a Context Management enabled.
    }
    \label{fig:post-training-main-table}
\end{table}

%% file: src/05_discussion.tex
\section{Limitations}
\label{sec:limitation}

\paragraph{Token Efficiency.}
Step 3.5 Flash achieves frontier-level intelligence but currently requires longer generation trajectories than Gemini 3.0 Pro to reach comparable quality.
Next step we will prune and compress the thinking for better efficiency while maintaining the same competitive performance.

\paragraph{Efficient Universal Mastery.}
We aim to unify generalist versatility with deep domain expertise. To achieve this efficiently, we are advancing variants of on-policy distillation, allowing the model to internalize expert behaviors with higher sample efficiency.

\paragraph{RL for Open-World Agentic Tasks.}

While Step 3.5 Flash demonstrates competitive performance on academic agentic benchmarks, the next frontier of agentic AI necessitates the application of RL to intricate, expert-level tasks found in professional work, advanced engineering, and scientific research. Solving these challenges is a prerequisite for deploying agents capable of genuine autonomy.

\paragraph{Operational Scope and Constraints.}
Step 3.5 Flash is tailored for coding and work-centric tasks, but may experience reduced stability during distribution shifts. This typically occurs in highly specialized domains or long-horizon, multi-turn dialogues, where the model may exhibit repetitive reasoning, mixed-language outputs, or inconsistencies in time and identity awareness.

%% file: src/tables/appendix-author-list.tex
\section*{Contributors}
The listing of authors is in alphabetical order based on their first names.
\begin{multicols}{4}
Ailin Huang \\
Ang Li \\
Aobo Kong \\
Bin Wang \\
Binxing Jiao \\
Bo Dong \\
Bojun Wang \\
Boyu Chen \\
Brian Li \\
Buyun Ma \\
Chang Su \\
Changxin Miao \\
Changyi Wan \\
Chao Lou \\
Chen Hu \\
Chen Xu \\
Chenfeng Yu \\
Chengting Feng \\
Chengyuan Yao \\
Chunrui Han \\
Dan Ma \\
Dapeng Shi \\
Daxin Jiang \\
Dehua Ma \\
Deshan Sun \\
Di Qi \\
Enle Liu \\
Fajie Zhang \\
Fanqi Wan \\
Guanzhe Huang \\
Gulin Yan \\
Guoliang Cao \\
Guopeng Li \\
Han Cheng \\
Hangyu Guo \\
Hanshan Zhang \\
Hao Nie \\
Haonan Jia \\
Haoran Lv \\
Hebin Zhou \\
Hekun Lv \\
Heng Wang \\
Heung-Yeung Shum \\
Hongbo Huang \\
Hongbo Peng \\
Hongyu Zhou \\
Hongyuan Wang \\
Houyong Chen \\
Huangxi Zhu \\
Huimin Wu \\
Huiyong Guo \\
Jia Wang \\
Jian Zhou \\
Jianjian Sun \\
Jiaoren Wu \\
Jiaran Zhang \\
Jiashu Lv \\
Jiashuo Liu \\
Jiawen Luo \\
Jiayi Fu \\
Jiayu Liu \\
Jie Cheng \\
Jie Luo \\
Jie Yang \\
Jie Zhou \\
Jieyi Hou \\
Jing Bai \\
Jingcheng Hu \\
Jingjing Xie \\
Jingwei Wu \\
Jingyang Zhang \\
Jishi Zhou \\
Junfeng Liu \\
Junzhe Lin \\
Ka Man Lo \\
Kai Liang \\
Kaibo Liu \\
Kaijun Tan \\
Kaiwen Yan \\
Kaixiang Li \\
Kang An \\
Kangheng Lin \\
Lei Yang \\
Liang Lv \\
Liang Zhao \\
Liangyu Chen \\
Lieyu Shi \\
Liguo Tan \\
Lin Lin \\
Lina Chen \\
Luck Ma \\
Mengqiang Ren \\
Michael Li \\
Ming Li \\
Mingliang Li \\
Mingming Zhang \\
Mingrui Chen \\
Mitt Huang \\
Na Wang \\
Peng Liu \\
Qi Han \\
Qian Zhao \\
Qinglin He \\
Qinxin Du \\
Qiuping Wu \\
Quan Sun \\
Rongqiu Yang \\
Ruihang Miao \\
Ruixin Han \\
Ruosi Wan \\
Ruyan Guo \\
Shan Wang \\
Shaoliang Pang \\
Shaowen Yang \\
Shengjie Fan \\
Shijie Shang \\
Shiliang Yang \\
Shiwei Li \\
Shuangshuang Tian \\
Siqi Liu \\
Siye Wu \\
Siyu Chen \\
Song Yuan \\
Tiancheng Cao \\
Tianchi Yue \\
Tianhao Cheng \\
Tianning Li \\
Tingdan Luo \\
Wang You \\
Wei Ji \\
Wei Yuan \\
Wei Zhang \\
Weibo Wu \\
Weihao Xie \\
Wen Sun \\
Wenjin Deng \\
Wenzhen Zheng \\
Wuxun Xie \\
Xiangfeng Wang \\
Xiangwen Kong \\
Xiangyu Liu \\
Xiangyu Zhang \\
Xiaobo Yang \\
Xiaojia Liu \\
Xiaolan Yuan \\
Xiaoran Jiao \\
Xiaoxiao Ren \\
Xiaoyun Zhang \\
Xin Li \\
Xin Liu \\
Xin Wu \\
Xing Chen \\
Xingping Yang \\
Xinran Wang \\
Xu Zhao \\
Xuan He \\
Xuanti Feng \\
Xuedan Cai \\
Xuqiang Zhou \\
Yanbo Yu \\
Yang Li \\
Yang Xu \\
Yanlin Lai \\
Yanming Xu \\
Yaoyu Wang \\
Yeqing Shen \\
Yibo Zhu \\
Yichen Lv \\
Yicheng Cao \\
Yifeng Gong \\
Yijing Yang \\
Yikun Yang \\
Yin Zhao \\
Yingxiu Zhao \\
Yinmin Zhang \\
Yitong Zhang \\
Yixuan Zhang \\
Yiyang Chen \\
Yongchi Zhao \\
Yongshen Long \\
Yongyao Wang \\
Yousong Guan \\
Yu Zhou \\
Yuang Peng \\
Yuanhao Ding \\
Yuantao Fan \\
Yuanwei Lu \\
Yuanzhen Yang \\
Yuchu Luo \\
Yudi Zhao \\
Yue Peng \\
Yueqiang Lin \\
Yufan Lu \\
Yuling Zhao \\
Yunzhou Ju \\
Yurong Zhang \\
Yusheng Li \\
Yuxiang Yang \\
Yuyang Chen \\
Yuzhu Cai \\
Zejia Weng \\
Zetao Hong \\
Zexi Li \\
Zhe Xie \\
Zheng Ge \\
Zheng Gong \\
Zheng Zeng \\
Zhenyi Lu \\
Zhewei Huang \\
Zhichao Chang \\
Zhiguo Huang \\
Zhiheng Hu \\
Zidong Yang \\
Zili Wang \\
Ziqi Ren \\
Zixin Zhang \\
Zixuan Wang \\
\end{multicols}

%% file: src/A_append.tex
\section{Architecture Details}
Table~\ref{tab:arch_hparams} summarizes key architecture hyper-parameters of {\ourmodel}.

\begingroup
\renewcommand{\arraystretch}{1.12}
\begin{table}[!h]
    \centering
    \small
    \setlength{\tabcolsep}{6pt}
    \begin{tabular}{l c}
        \toprule
        \rowcolor{header_gray}
        \textbf{Hyper-Parameter} & \textbf{Value} \\
        \midrule
        \rowcolor{section_gray}\multicolumn{2}{l}{\textsc{\textbf{Backbone}}} \\
        Vocabulary size ($V$) & 128{,}896 \\
        Model width ($d_{\text{model}}$) & 4096 \\
        Transformer blocks & 45 (3 dense + 42 MoE) \\
        \addlinespace[2pt]
        \rowcolor{section_gray}\multicolumn{2}{l}{\textsc{\textbf{MoE FFN}}} \\
        Experts per MoE block & 288 + 1 shared \\
        Routing & top-$k=8$ \\
        Dense FFN hidden size & 11{,}264 \\
        MoE expert hidden size & 1{,}280 \\
        \addlinespace[2pt]
        \rowcolor{section_gray}\multicolumn{2}{l}{\textsc{\textbf{Attention}}} \\
        Hybrid block structure & 3 SWA blocks + 1 full attention block \\
        SWA window size & 512 \\
        KV heads (GQA) & 8 \\
        Query heads (full / SWA) & 64 / 96 \\
        Gate Type & head-wise on output \\
        Head dimension & 128 \\
        RoPE $\theta$ & 10{,}000 \\
        RoPE dims (full / SWA) & 64 / 128 \\
        \addlinespace[2pt]
        \rowcolor{section_gray}\multicolumn{2}{l}{\textsc{\textbf{Multi-Token Prediction}}} \\
        MTP blocks  & 3 (Dense SWA) \\
        \addlinespace[2pt]
        \rowcolor{section_gray}\multicolumn{2}{l}{\textsc{\textbf{Parameter Counts}}} \\
        Total params (backbone) & 196B \\
        Activated params / token (backbone) & 11B \\
        Total params (with MTP3) & 198B \\
        Activated params / token (with MTP3) & 13B \\
        \bottomrule
    \end{tabular}
    \caption{Key architecture hyper-parameters of {\ourmodel}. ``Activated params'' are reported per token and exclude embedding/output matrices.}
    \label{tab:arch_hparams}
\end{table}
\endgroup

\subsection{Head-wise Gated Attention}
\label{appx:gated_attn_details}
Each attention head is assigned a lightweight, input-dependent scalar gate, allowing the model to dynamically modulate information flow across the hybrid layout with negligible computational overhead.

Formally, for a (single) head of dimension $d$, let $\boldsymbol{q}_i,\boldsymbol{k}_j,\boldsymbol{v}_j\in\mathbb{R}^{d}$ denote the query vector at position $i$ and the key and value vectors at position $j$,
the scaled dot-product scores $s$, the corresponding attention weights $\alpha$ and the outputs $\boldsymbol{y}$ are computed as follows:
\begin{equation}
\label{eq:std_attn}
s_{i,j} = \langle \boldsymbol{q}_i,\boldsymbol{k}_j\rangle/\sqrt{d},
\qquad
Z_i = \sum_{j'} \exp(s_{i,j'}),
\qquad
\alpha_{i,j} = \exp(s_{i,j})/Z_i,
\qquad
\boldsymbol{y}_i = \sum_j \alpha_{i,j}\,\boldsymbol{v}_j .
\end{equation}
Given the input representation $\boldsymbol{x}_i$ at position $i$, we compute a head-wise gate $g_i$ to modulate the head output:
\begin{equation}
    \label{eq:head_gated_attn}
    g_i = \sigma(\boldsymbol{w}_{gate}^\top \boldsymbol{x}_i),
    \qquad
    o^{\mathrm{gate}}_{i} = g_i\,\boldsymbol{y}_i,
\end{equation}
where $\sigma(\cdot)$ is the sigmoid function and $\boldsymbol{w}_{gate}$ is a learnable vector.

Head-wise gated attention can be viewed as introducing an \emph{input-dependent sink token}~\cite{openai2025gptoss120bgptoss20bmodel} into the attention mechanism.
Substituting $\sigma(g) = \frac{1}{1+\exp(-g)}$ into Equation \ref{eq:head_gated_attn}, we have
\begin{equation}
    \boldsymbol{o}^{\mathrm{gate}}_i
    = \sum_j \frac{\exp(s_{i,j})}{Z_i + e^{-g_i} Z_i} \boldsymbol{v}_j,
\end{equation}
where $\exp(-g_i)Z_i$ acts as an \emph{input-dependent sink mass} in the softmax normalizer.
As shown in Section~\ref{sec:head_vs_sink}, this adaptive formulation consistently outperforms fixed (input-independent) sink tokens.

\subsection{Speed Benchmark of Attention Enhancements}\label{appx:attn_enhancement_benchmark}

We conduct simulations with MTP-3 to evaluate the latency overheads of the two enhancements under an ideal workload. Table~\ref{tab:attn_enhancement_benchmark} presents the relative increment of theoretical FLOPs and latency. Increasing the number of query heads in SWA slightly raises the FLOPs but has less impact on latency. This is due to a query-to-$kv$ ratio of 12, which keeps SWA in the IO-bound region, even when considering MTP-3. For head-wise gating, neither FLOPs nor latency has noticeable difference because of its lightweight.

\begin{table}[h]
\small
\centering
\begin{tabular}{lcccccc}
\toprule
\multirow{2}{*}{\textbf{Backbone}} & \multirow{2}{*}{\makecell{\textbf{SWA} \\ \textbf{Heads}}} & \multirow{2}{*}{\textbf{Setting}} & \multicolumn{2}{c}{\textbf{Decode (FLOPs / Lat.)}} & \multicolumn{2}{c}{\textbf{Prefill (FLOPs / Lat.)}} \\
\cmidrule(lr){4-5} \cmidrule(lr){6-7}
 & & & 64k & 256k & 64k & 256k \\
\midrule
\multirow{4}{*}{\makecell{\ourmodel \\ ($S3F1$ layout)}}
 & 64 & no gate & 1.00 / 1.00 & 1.00 / 1.00 & 1.00 / 1.00 & 1.00 / 1.00 \\
 & 96 & no gate & 1.02 / 1.01 & 1.01 / 1.00 & 1.08 / 1.06 & 1.04 / 1.03 \\
 & 64 & head-wise & 1.00 / 1.00 & 1.00 / 1.00 & 1.00 / 1.02 & 1.00 / 1.01 \\
 & 96 & head-wise & 1.02 / 1.02 & 1.01 / 1.00 & 1.08 / 1.08 & 1.04 / 1.05 \\
\bottomrule
\end{tabular}
\caption{
Relative increment under different SWA head counts and gating strategies. The metrics are presented as FLOPs / Latency. The baseline configuration (first line) is normalized to 1.0.
}
\label{tab:attn_enhancement_benchmark}
\end{table}

\begin{table}[h]
\small
\centering
\begin{tabular}{llccccc}
\toprule
\multirow{2}{*}{\textbf{Backbone}} & \multirow{2}{*}{\textbf{Layout}} & \multirow{2}{*}{\makecell{\textbf{SWA} \\ \textbf{Heads}}} & \multicolumn{2}{c}{\textbf{Decode}} & \multicolumn{2}{c}{\textbf{Prefill}} \\
\cmidrule(lr){4-5} \cmidrule(lr){6-7}
 & & & 64K & 256K & 64K & 256K \\
\midrule
\multirow{4}{*}{\makecell{\ourmodel}}
 & $S3F1$  & 64 & 1.00 & 1.00 & 1.00 & 1.00 \\
 & \texttt{$S3F1+$Head}                & 96 & 1.02 & 1.01 & 1.08 & 1.04 \\
 & $S1F1$                & 64 & 1.18 & 1.47 & 1.38 & 1.71 \\
 & $FFFF$                & 64 & 1.51 & 2.33 & 2.07 & 3.00 \\
\midrule
\multirow{4}{*}{Internal 30B-A3B}
 & $S3F1$  & 32 & 1.00 & 1.00 & 1.00 & 1.00 \\
 & \texttt{$S3F1+$Head}                  & 48 & 1.02 & 1.01 & 1.05 & 1.02 \\
 & $S1F1$                & 32 & 1.42 & 1.74 & 1.50 & 1.80 \\
 & $FFFF$                & 32 & 2.21 & 3.16 & 2.47 & 3.34 \\
\bottomrule
\end{tabular}
\caption{
Relative FLOPs cost across different backbones and attention patterns. The head count refers to SWA heads. For each backbone, the configuration with minimum FLOPs ($S3F1$ with reduced heads) is the baseline (1.0).
}
\label{tab:rel}
\end{table}

\subsection{Meta Token}\label{app:meta}

Recent literature \cite{gao2025metadata, allenzhu2024physics, fan2025urlsmetadatadiversityposition} has shown both theoretically and empirically that pre-pending structured metadata to pre-training sequences can improve data efficiency and accelerate convergence: by exposing high-level attributes (\eg, modality, language, domain), metadata provides global cues that reduce uncertainty about the upcoming content and thus makes next-token prediction easier.

Motivated by this paradigm, we associate each training example with a metadata string $\mathbf{M}$ in a human-readable format, including content type (\eg, Code, Book, Paper, Web), language (\eg, EN, ZH), domain, and source. We then prepend $\mathbf{M}$ to the original token sequence $\mathbf{x}$, forming a single training sequence $\mathbf{s} = [\mathbf{M}; \mathbf{x}]$. During pre-training, the model is trained to maximize the likelihood of $\mathbf{s}$:
\begin{equation}
\mathcal{L}_{\text{full}}(\theta)
= -\sum_{t=1}^{|\mathbf{s}|} \log P_\theta(s_t \mid \mathbf{s}_{<t}).
\end{equation}

After an initial phase of approximately 3.8T tokens, we keep $\mathbf{M}$ in the context but mask out its positions from the loss while continuing to predict the payload tokens:
\begin{equation}
\mathcal{L}_{\text{mask}}(\theta)
= -\sum_{t=|\mathbf{M}|+1}^{|\mathbf{s}|} \log P_\theta(s_t \mid \mathbf{s}_{<t})
= -\sum_{t=1}^{|\mathbf{x}|} \log P_\theta(x_t \mid \mathbf{M}, \mathbf{x}_{<t}).
\end{equation}

We hypothesize that by this stage the model has already learned to effectively use metadata as a conditioning signal. Masking the metadata loss therefore allocates optimization pressure entirely to the payload tokens, while still benefiting from the explicit conditioning on data characteristics.

\begin{table}[t]
\centering
\renewcommand{\arraystretch}{1.2}

\begin{tabular}{lcc}
\toprule
\rowcolor{header_gray}
\textbf{Hyper-Parameter} & \textbf{100B-A10B} & \textbf{30B-A3B} \\
\midrule
Total Tokens & 250B & 1.4T \\
Optimizer &  \multicolumn{2}{c}{Muon \cite{jordan2024muon} }\\
Peak learning rate & $1.31\times10^{-4}$ & $1.1\times10^{-3}$ \\
Batch-size warmup & - & First 30B tokens \\
\midrule
Layers & 43 & 48 \\
Dimension & 4096 & 2048 \\
Leading Dense Layers & 1 & 1 \\
Routed Experts & 96 & 128 \\
Active Experts & 4 & 8 \\
Shared Experts & 1 & 1 \\
Load Balancing Method &  \multicolumn{2}{c}{Loss Free \cite{wang2024lossfreebalancing}} \\
Attention module & \multicolumn{2}{c}{GQA8} \\
Sequence Length & \multicolumn{2}{c}{4096} \\
Vocab Size & \multicolumn{2}{c}{129280} \\
Batch Size & 8192 & 16384 \\
Weight Decay & \multicolumn{2}{c}{0.1} \\
Partial RoPE & Disabled & Enabled \\
MTP & Enabled & Disabled \\
\bottomrule
\end{tabular}
\caption{Training configuration for the 100B-A10B and the 30B-A3B architecture ablation suite.\label{tab:train_setting_1p4t}}
\end{table}

\begin{table}[t]
\centering
\small
\renewcommand{\arraystretch}{1.12}
\setlength{\tabcolsep}{4pt}
\resizebox{\textwidth}{!}{
\begin{tabular}{l c cccccccccccc}
\toprule
\multirow{2}{*}{\textbf{Layout}} &
\multirow{2}{*}{\makecell{\textbf{SWA} \\ \textbf{Heads}}} &
\multicolumn{12}{c}{\textbf{Pre-train Evaluation}} \\
\cmidrule(lr){3-14}
& &
\textbf{BBH}        &  \textbf{MMLU}       &  \textbf{MMLU-Redux} &  \textbf{MMLU-Pro}   &   \textbf{SimpleQA}   &  \textbf{GSM8K} &  \textbf{MATH} &  \textbf{HumanEval} &  \textbf{MBPP} &  \textbf{C-EVAL} &  \textbf{CMMLU} &  \textbf{Avg.} \\
\midrule

{$FFFF$} & 32 &
\textbf{66.0} & 64.5 & 69.7 & 35.7 & 7.2 & 70.0 & 39.2 & \textbf{48.8} & 53.4 & 69.7 & 70.5 & 54.1 \\

{$S1F1$} & 32 &
64.1 & 64.7 & 69.8 & \textbf{37.7} & 7.5 & 70.1 & 43.9 & 47.0 & 56.2 & 69.8 & 69.8 & 54.6 \\

{$S3F1$} & 32 &
61.7 & 64.2 & 69.4 & 33.7 & \textbf{8.0} & 67.4 & 41.5 & 47.6 & 56.0 & 69.5 & 70.9 & 53.6 \\

\texttt{$S3F1$+Head} & 48 &
65.3 & \textbf{65.9} & \textbf{71.0} & 37.4 & 7.5 & \textbf{72.2} & \textbf{44.5} & \textbf{48.8} & \textbf{58.6} & \textbf{70.2} & \textbf{71.0} & \textbf{55.7} \\

\bottomrule
\end{tabular}
}
\caption{Pre-training evaluation results for hybrid attention layout ablations ($W{=}512$) on 30B-A3B.
$F$ denotes full attention and $S$ denotes SWA; $S3F1$ indicates three $S$ and one $F$ in the hybrid layout.
\texttt{$S3F1$+Head} increases the number of SWA heads from 32 to 48.}
\label{tab:swa_pretrain_results}
\end{table}

\subsection{Pre-training Ablations Details}
\label{sec:arch_ablation}

We conduct controlled pre-training ablations to isolate the effects of (i) different hybrid attention layout and (ii) sink tokens versus head-wise gated attention.

\noindent\textbf{Hybrid attention layout.}
We adopt a \textbf{30B-A3B MoE} architecture to evaluate the downstream impact of different hybrid attention layout under a fixed token budget. The training follows a strict, multi-stage pipeline: a 30B-token warmup phase, followed by 1T tokens of main pre-training, a 300B-token cooldown phase, and an additional 100B-token long-context specialization stage—totaling approximately 1.4T tokens. Supervised fine-tuning (SFT) is then performed on a 0.1× downsampled dataset. Full training details are provided in Table~\ref{tab:train_setting_1p4t}.

\noindent\textbf{Gate vs. sink (scaled setting).}
We pre-train a \textbf{100B-A10B MoE} model for $\sim$250B tokens to compare sink tokens and head-wise gating under a larger-scale regime.

Pre-training results of the architectural ablations are presented in Tables \ref{tab:gate_vs_sink_100b} and \ref{tab:swa_pretrain_results}.
We employ the evaluation protocols detailed in Section \ref{sec:evals}. Specifically, GPQA \cite{rein2023gpqa} is evaluated using 5-shot prompting, while HumanEval \cite{chen2021evaluating} and MBPP \cite{austin2021program} utilize 3-shot prompting.

The post-training results in Table \ref{tab:swa_sft_results} are aggregated as follows:
\begin{itemize}[leftmargin=*,nosep]
\item \textbf{Reasoning:} The average of MMLU-Pro~\cite{wang2024mmlupro}, GPQA-Diamond~\cite{rein2023gpqa}, LiveCodeBench v6~\cite{jain2024livecodebench}, and LiveBench\cite{white2025livebenchchallengingcontaminationlimitedllm}.
\item \textbf{Math:} The average of AIME 2024~\cite{AIME}, AIME 2025~\cite{AIME25}, HMMT 2025 Feb.~\cite{balunovic2025matharena}, and CNMO 2024\footnote{https://www.cms.org.cn/Home/comp/comp/cid/12.html}.
\item \textbf{Code:} The average of CF-Div2-Stepfun and LiveCodeBench v6~\cite{jain2024livecodebench}.
\item \textbf{Sci:} Represented by GPQA-Diamond~\cite{rein2023gpqa}.
\item \textbf{General:} The average of IFEval~\cite{zhou2023instructionfollowingevaluationlargelanguage}, IFBench\cite{pyatkin2025generalizing}, WildBench\cite{lin2024wildbenchbenchmarkingllmschallenging}, Arena-Hard~\cite{arenahard2024}, and MultiChallenge~\cite{sirdeshmukh2025multichallengerealisticmultiturnconversation}.
\item \textbf{LongCtx:} The average of six benchmark-level averages: (i) the average score across context lengths 8k-128k on RULER~\cite{hsieh2024ruler}, (ii) the average score over the Short and Medium subsets of LongBench v2~\cite{bai2024longbenchv2}, (iii) the average score across context lengths 8k-128k on HELMET~\cite{yen2024helmet}, (iv) GSM-Infinite~\cite{zhou2025gsminfinite}, (v) the overall score on FRAMES~\cite{krishna2025fact}, and (vi) the overall score on RepoQA~\cite{liu2024repoqaevaluatinglongcontext}.

\end{itemize}

Tables~\ref{tab:swa_sft_results} and~\ref{tab:swa_pretrain_results} show that the vanilla $S3F1$ layout underperforms the full-attention baseline on general pre-training benchmarks and consistently degrades SFT quality (\eg, BBH: $-4.3$; SFT Avg: $-0.7$). Increasing the number of SWA query heads substantially closes this gap (e.g., MMLU-Pro: $+3.7$; SFT Reasoning: $+0.4$), with only a minor regression on SFT Code ($-0.6$), while matching or exceeding the full-attention baseline on several metrics. Table~\ref{tab:gate_vs_sink_100b} further demonstrates that head-wise gated attention yields an average improvement from 62.5 to 64.4 ($+1.9$) on the sink token metric.

\section{Detail Analysis of Localized Activation Blow-up}\label{app:clip}

To investigate the root cause of the localized activation blow-up, we analyze the tokens that trigger the largest expert activations across all layers, and identify two distinct large activation patterns: (1) Specific lexical items, such as special tokens and punctuation, commonly elicit large but not dramatic activations, particularly in the shallower layers. This pattern is not recognized as a failure mode by us, as there is no rapid increment and it may serve as an internal mechanism for semantic modeling~\cite{sun2024massive,an2025systematic}. Another pattern is that (2) some high-frequency bi-grams trigger extremely large activations on the first token, which represents the failure mode we are investigating. The pattern is triggered by several factors:
The frequency of a bi-gram's occurrence is sufficiently high, and the MoE FFN is fine-grained enough, allowing an expert to specialize in that bi-gram without being regulated by the load balancing mechanism. This specialization serves as a shortcut: once the expert is activated, the output becomes deterministic, and other networks no longer influence the prediction. While finding shortcuts is a reasonable approach to minimizing loss, in a MoE model with a pre-norm architecture~\cite{radford2019language,xiong2020layer}, there is a straightforward, pathological solution for achieving such deterministic predictions, as outlined next. The model's final representation is the sum of the outputs from all layers, followed by a RMSNorm. This can be expressed as a combination of the outputs from the experts and the attention layers:
\begin{equation}
\boldsymbol{h}_{\text{final}} = \text{RMSNorm}(\underbrace{\text{expert}_{\text{outlier}}}_{\boldsymbol{h}_{\text{outlier}}} + \underbrace{\sum_{l=1}^L \text{attn}_l + \sum_{\substack{l,e\\(l,e)\text{ is not a outlier}}} \text{expert}_{l,e}}_{\boldsymbol{h}_{\text{others}}}),
\end{equation}
where attn, MoE, expert represent the output hidden states of their respective modules, while \(L\) and \(E\) denote the number of layers and experts, respectively. The straightforward solution is to boundlessly enlarge $\text{expert}_{\text{outlier}}$, then
\begin{equation}
\text{RMSNorm}(\boldsymbol{h}_{\text{final}}) = \lim_{c\rightarrow \infty}\text{RMSNorm}(c \cdot \hat{\boldsymbol{h}}_\text{outlier} + \boldsymbol{h}_{\text{others}}) = \text{RMSNorm}(\boldsymbol{h}_\text{outlier}),
\end{equation}
where we decouple $\boldsymbol{h}_\text{outlier}$ to the magnitude $c$ and the unit vector $\hat{\boldsymbol{h}}_\text{outlier}$ denoting the direction.

SwiGLU~\cite{shazeer2020gluvariantsimprovetransformer}, the expert architecture in \ourmodel, provides a way to generate large outputs, even when the weight decay effectively suppresses the weight norms. SwiGLU is defined as follows:
\begin{equation}
\text{SwiGLU}(\boldsymbol{x}) = \boldsymbol{W}_{\text{down}}\left(\text{SiLU}(\boldsymbol{W}_{\text{gate}}\boldsymbol{x})\cdot\boldsymbol{W}_{\text{up}}\boldsymbol{x}\right).
\end{equation}
We analyze the activation norms of $\boldsymbol{W}_{\text{gate}}\boldsymbol{x}$ and $\boldsymbol{W}_{\text{up}}\boldsymbol{x}$ and find no significant differences between outlier experts and normal experts. However, the element-wise product produces abnormal outputs, which have
\begin{equation}
\|\text{SiLU}(\boldsymbol{W}_{\text{gate}}\boldsymbol{x})\|\cdot\|\boldsymbol{W}_{\text{up}}\boldsymbol{x}\|\approx\|\text{SiLU}(\boldsymbol{W}_{\text{gate}}\boldsymbol{x})\cdot\boldsymbol{W}_{\text{up}}\boldsymbol{x}\|,
\end{equation}
in outlier experts. It can be achieved only if $\text{SiLU}(\boldsymbol{W}_{\text{gate}}\boldsymbol{x})$ and $\boldsymbol{W}_{\text{up}}\boldsymbol{x}$ are highly aligned and concentrate on a very limited number of dimensions. Consequently, only a limited number of rows from \(\boldsymbol{W}_{\text{up}}\) are utilized due to the sparse input. This observation leads us to prefer activation clipping over weight clipping, as activation's numerical property directly contribute to the blow-up and the sparsity, and activation clipping can promptly address these issues. Besides, activation clipping has negligible negative effects, as well-behaved activations rarely exceed the threshold.

When using the Muon optimizer, gated linear units, such as SwiGLU, are susceptible to logit explosion. This vulnerability arises from similar mechanisms that cause explosion in attention, as reported in \cite{kimi_k2}.
For an outlier expert specialized to some specific bi-gram, hidden states routed to it are expected to be closely aligned to its router embedding. We validate this by inputting the router embedding into a outlier expert and directly predicting outputs based on this expert's output. The predicted distribution aligns with that of the real data and the entire network's performance. Combined with the overly single training target (to predict the second token in the bi-gram), we argue that gradients w.r.t. the outlier expert's parameters, $\boldsymbol{W}_{\text{gate}}$, $\boldsymbol{W}_{\text{up}}$ and $\boldsymbol{W}_{\text{down}}$, are not only abnormally low rank (denoted as $r$), but also consistently point in a direction that emphasizes the magnitude as analyzed in the first factor, without rotation. Let the update matrices of a parameter matrix $\boldsymbol{W}\in\mathbb{R}^{N\times N}$ to be
\begin{equation}
\Delta \boldsymbol{W} = \sum_i \sigma_i \boldsymbol{u}_i\boldsymbol{v}_i^\top =  \underbrace{\sum_{i=1}^r \sigma_i \boldsymbol{u}_i\boldsymbol{v}_i^\top}_{\text{low rank signal}} + \underbrace{\sum_{j=r+1}^{N} \sigma_j \boldsymbol{u}_j\boldsymbol{v}_j^\top}_{\text{noise}}
\end{equation}
Accumulating updates over optimization steps will rapidly increase the singular value of the low-rank signals, resulting in an explosion of the weight parameter. In the GLU structure, $\|\text{SiLU}(\boldsymbol{W}_{\text{gate}}\boldsymbol{x})\cdot\boldsymbol{W}_{\text{up}}\boldsymbol{x}\|$ squares the spectral norm in our strong alignment case, making the progress more sharp. Additionally, Muon completely eliminates the influence of gradient magnitudes. During the blow-up process, RMSNorm reduces the gradients of large inputs. When using the Adam optimizer, its $\epsilon$ acts as a threshold to filter out small gradients during the learning rate adaptation, which can hinder the progress. In contrast, Muon consistently and effectively orthogonalizes the gradients, resulting in more aggressive updates.

\section{Step Pre-training Data Foundation}
\label{sec:step-data}

\subsection{Knowledge Data Construction}

\subsubsection{StepCrawl}
\label{subsubsec:stepcrawl}

Beyond standard web-scale datasets (\eg, CommonCrawl), we develop \textbf{StepCrawl}, an in-house crawling and curation system designed to acquire \emph{high-quality} and \emph{diverse} tokens at scale.
StepCrawl serves as a primary data source for both high-signal web pages and document-like content (notably PDFs), which frequently contain long-form, high-information-density material.

\noindent A key component of StepCrawl is a site and URL selection layer powered by a WebOrganizer-style model~\cite{weborganizer}.
We adapt the capabilities introduced in WebOrganizer and further fine-tune a version tailored to our pipeline.
During crawling, each fetched web page is analyzed by this model, forming a lightweight LM-in-the-loop feedback cycle that (i) filters SEO-driven and other low-utility pages, and (ii) guides crawl-budget allocation by balancing site categories (e.g., preventing disproportionate crawling of tool and e-commerce sites) to preserve corpus diversity and reduce topical skew.
In practice, StepCrawl processes on the order of $\sim$1B pages per day under this quality- and diversity-aware scheduling policy.

\noindent All crawling activities strictly adhere to \texttt{robots.txt} and site-specific access policies.
The collected content is subsequently passed through a multi-stage filtering process (quality scoring, deduplication, and sanitization), ensuring that only high-utility and policy-compliant data are retained for training.

\subsubsection{Quality Refinement and Stratification}

\paragraph{Quality stratification.}
Inspired by Nemotron-CC~\cite{su2025nemotroncctransformingcommoncrawl}-style quality bucketing, we divide the internal web data into quality tiers and sample preferentially from higher tiers.
We label each document using an ensemble of six lightweight scorers/classifiers and ensemble the tier assignments across scorers.
In the final recipe, we keep High/Medium-High/Medium and discard Medium-Low/Low, which substantially improves token efficiency in ablations.
For book and paper corpora, we apply the same stratification but restrict retention to High/Medium-High tiers exclusively during the annealing stage to maximize diversity.
In addition to the shared six-scorer ensemble, we integrate additional domain-specific filters targeting STEM and knowledge-dense content, and down-sample overrepresented domains to ensure balanced representation.

\paragraph{Embedding-based cluster rebalancing.}
We leverage embedding-based corpus balancing as a principled way to further reduce redundancy and mitigate distribution skew.
Specifically, we embed large-scale Chinese/English web data, run k-means clustering (100k+ clusters), and down-sample clusters with disproportionate mass.
In ablations, this cluster-level rebalancing in the cooldown stage improves a broad set of benchmarks.

\paragraph{Knowledge-Intensive Mining and Augmentation.}
We construct a dedicated knowledge subset using a lightweight two-stage pipeline built on the shared embedding representation described above.
First, a curated inventory of high-value entities, concepts, and relations is used to retrieve knowledge-dense documents and passages from the full corpus in embedding space; these candidates are ranked by a knowledge-density model and simple coverage heuristics.
Second, for a portion of the retrieved content, we apply targeted transformations such as controlled rephrasing and QA synthesis to improve learnability.
The resulting samples are mixed back into the training mixture to increase effective knowledge signal density.
We observe consistent gains from this pipeline in ablations, while a detailed causal analysis of its benefits is left for future work.

\subsection{Code Data}

\subsubsection{Pure-Code}
\label{sec:pure_code}

\noindent We refine our internal programming dataset using a modified version of the OpenCoder filtering rules~\cite{huang2025opencoderopencookbooktoptier}, introducing a calibrated relaxation to balance data quality and diversity.
In our pipeline, applying OpenCoder filters generates a set of ``hits'' for each document, where each hit represents a violation of a heuristic rule (signaling potential noise).
We categorize the corpus by these hit counts: \texttt{hit0} for clean documents (zero violations), \texttt{hit1} for one violation, and so on.

\noindent Our internal ablations reveal a clear quality-diversity trade-off: strict filtering (e.g., \texttt{hit0}-only) over-prunes the corpus, while no filtering introduces excessive noise.
We find that the \texttt{hit0--6} configuration (accepting documents with up to 6 violations) yields the best overall benchmark performance, retaining a wider variety of high-signal code compared to the original strict constraints.

\subsubsection{PR/Issue/Commit Data}
\label{sec:pr_issue_data}

\noindent To enhance software engineering capabilities, we construct a comprehensive dataset from GitHub repositories with over 10 stars, comprising PRs, issues, and commits.
We apply strict filtering on repository popularity and content quality, and use LLMs to generate missing issue descriptions, resulting in a 5-million-sample foundation.
From this, we derive four training subsets:

\noindent \textbf{(1) Base PR/Issue/Commit Data:}
We crawl data via GHArchive and GitHub API, including full commit histories.
We extract changes and validate a small portion of samples against \texttt{git diff} ground truth, then filter to 20+ mainstream languages (\eg, Python, Java, C++).
We strictly deduplicate against SWE-Bench Verified~\cite{swe_verified} and SWE-Bench Multilingual~\cite{yang2025swesmith} to prevent leakage.

\noindent \textbf{(2) Concatenated PR-Dialogue Data (90B tokens):}
We generate 90B tokens of code-editing training data by applying two Agentless-inspired templates~\cite{xia2024agentless}:
(1) \textit{File localization}: Given a problem description and repository structure, identify target file paths;
(2) \textit{Code repair}: Given a problem description and file content, generate precise modifications via SEARCH/REPLACE blocks.

\noindent We integrate this 90B code-editing data into two training phases with phase-specific masking strategies.
In the annealing stage of pre-training, only template scaffolding is masked; in mid-training, the data is converted to chat dialogs with user prompts masked.
Internal ablations show consistent gains over SWE-Bench Verified and SWE-Bench Multilingual in the cooldown stage and mid-training.

\noindent \textbf{(3) Rewritten Reasoning-Oriented Data (12B tokens):}
From the Python subset of our base dataset, we derive bug-fix samples via LLM change-type annotation.
We apply two concise rewriting strategies:
(1) \textit{Reasoning reconstruction}: an LLM reconstructs the PR author’s problem-solving process (problem analysis, root cause identification, solution design, and code implementation), injected into PR-Dialogue format.
Hallucinated/inconsistent traces are filtered via rule-based and LLM verification.
(2) \textit{Active Reading notebooks}: PR/issue/commit data is converted into structured learning outlines (motivation, root causes, design decisions, insights), then synthesized into coherent technical notes.
These rewritten datasets ($\sim$12B tokens) are incorporated during mid-training, yielding further gains on SWE-Bench Verified.

\noindent \textbf{(4) Environment-based Seed Data.}
We curate executable environments derived from raw PR, issue, and commit records using the environment building pipeline described in Appendix~\ref{sec:code-agent}.
Candidate samples are rigorously filtered to ensure test-patch inclusion and validated via strict rule-based criteria to guarantee environmental reproducibility.
Furthermore, selected issues undergo targeted rewriting to augment data quality and coverage.
The resulting dataset comprises hundreds of thousands of seed samples, including problem descriptions, code changes, and test functions, and serves as the foundational bedrock for enhancing agentic coding capabilities, driving significant performance gains in downstream agent tasks.

\subsection{Mathematics \& STEM Data}

To enhance reasoning capabilities and elicit intelligence from knowledge, we curate a large-scale mathematics and STEM dataset.
Beyond the standard Common Crawl data used in prior works~\cite{shao2024deepseekmathpushinglimitsmathematical,zhou2025megamathpushinglimitsopen}, we leverage our in-house {StepCrawl} system to harvest a massive scale of additional mathematics-related data.
Specifically, we implement a filtering pipeline inspired by MegaMath~\cite{zhou2025megamathpushinglimitsopen}, utilizing an ensemble of internal classifiers alongside FineMath~\cite{allal2025smollm2smolgoesbig}.
This allows us to retain hundreds of billions of mathematics-related tokens distinct from Common Crawl.
We further collect a diverse 100M-sample educational dataset encompassing exercises, quizzes, and instructional content.
This collection bridges the gap between academic theory and professional application, covering domains from K-12 mathematics/physics/chemistry and humanities to adult vocational exams (CPA, Legal).
Early-stage experiments confirm that this problem-solving data is crucial for optimizing token efficiency during pre-training.

\subsection{Data Infrastructure}
\label{sec:data-infra}

\noindent Our data construction and curation pipeline runs on a high-throughput in-house data infrastructure system designed for large-scale deduplication, mining, and model-inference filtering.
We operate hybrid CPU/GPU clusters with distributed frameworks such as \texttt{Spark} and \texttt{Ray} to execute both large-volume processing (e.g., minhash-based deduplication) and model-driven curation workloads (e.g., embedding generation and classifier/LM inference), backed by a storage layer spanning object storage (OSS), \texttt{HDFS}, and \texttt{JuiceFS} for efficient reads/writes of raw corpora and intermediate artifacts.

\subsection{Data Ablations Setting}

To rigorously assess data quality and the impact of curation strategies, we conduct an extensive ablation suite using the 30B-A3B MoE architecture trained with the Muon optimizer, consistent with the mainline settings (Table~\ref{tab:train_setting_1p4t}).
Adhering to a strict token efficiency protocol, we set a fixed training budget for all experiments.
Models are evaluated on the comprehensive benchmarks listed in Section~\ref{sec:evals}, alongside a series of carefully designed held-out compression (perplexity) test sets.
We observe that compression metrics often provide a more direct measure of knowledge capacity, offering signals complementary to mainstream benchmarks.

Internal experiments on the 30B-A3B MoE model demonstrate its superior performance and stability compared to smaller proxies.
While smaller models are computationally cheaper, they often fail to capture the nuances of complex reasoning and lack the capacity to memorize long-tail patterns, leading to an artificial bias towards data repetition.
Empirically, the 30B-A3B size offers stronger stability and better fidelity to full-scale trends.

\section{Post Training Details}
This section describes the post-training process that refines the base model into a high-performance agentic system, covering SFT with rigorous data processing and quality control, followed by large-scale RL to further improve reasoning, tool use, and generalization.

\subsection{SFT Details}
\subsubsection{SFT Data Processing Pipeline}
Across all domains, we apply a unified data processing pipeline that emphasizes answer verifiability, reasoning quality, and execution realism. To ensure overall data integrity, the aggregated dataset undergoes a strict two-stage filtration process:
\begin{enumerate}
\item \textbf{Rule-based Filtering:} We eliminate low-quality data exhibiting degenerate patterns, such as infinite repetition, harmful content, and personally identifiable information.
\item \textbf{Model-based Filtering:} We utilize specialized models to detect and filter out linguistically inconsistent data. By identifying and removing samples with unnatural language mixing, we significantly refine the dataset's linguistic purity and overall quality.
\item \textbf{Decontamination:} We conduct comprehensive benchmark decontamination to prevent test set leakage. This involves both exact matching (with digit masking to catch numerical modifications) and $N$-gram matching.

\end{enumerate}
This process yields a final refined dataset of 871k samples, totaling 7.23B tokens.The detailed distribution of the SFT data is presented in Table \ref{tab:sft_data_statistics}.

\vspace{-1.5cm}

\subsection{RL Details and Ablations}
This section details the large-scale RL post-training, covering data curation, asynchronous search-agent training, and ablations on dense and MoE models.

\vspace{-0.5cm}
\subsubsection{Data Curation}
We curate the RL training dataset by aggregating problems from open-source collections and competition archives spanning competitive coding, STEM, and synthetic data for general RLVR training.
To prevent data contamination, we strictly exclude problems from competitions held during 2024–2026. The dataset is further augmented with:
(i) synthetic arithmetic problems involving 11–13 digit integers;
(ii) a generator–validator pipeline that synthesizes additional test cases for coding tasks; and
(iii) synthetic environments for general reasoning tasks, such as puzzle and instruction following.

We apply a two-stage filtering process. First, deterministic rule-based pruning removes prompts containing images, external links, or open-ended requirements without a unique final answer. Second, an accuracy-based filter excludes trivial or degenerate problems. During training, each batch is constructed by sampling from different domains according to predefined sampling probabilities.

\subsubsection{Reward System}
\label{sec:rl_reward}

\paragraph{Verifiable Rewards.}

For STEM tasks, we employ gpt-oss-120b~\cite{openai2025gptoss120bgptoss20bmodel} as the verifier model, using the following structured prompt (originally in Chinese) to rigorously assess final-answer correctness. For coding tasks, we utilize sandboxes to validate code execution against test cases with soft reward.

\begin{MyBox}{}
\footnotesize
\linespread{0.8}\selectfont
You are a strict grader. Below you are given the problem, the student's answer, and the reference answer. Please determine whether the student's answer is correct according to the rules below.

\textbf{Grading procedure:}

1. \textbf{Overall check:} If the student's response is incomplete, lacks a clear final answer, or contains repeated content multiple times $\rightarrow$ mark as incorrect.

2. \textbf{Final-answer match:} Extract the student's \textbf{explicit final answer} and compare it with the reference answer:
   \begin{itemize}[itemsep=0pt, parsep=0pt, topsep=0pt]
      \item If they are exactly equivalent semantically or mathematically $\rightarrow$ proceed to process check.
      \item If numerical computation is involved and the discrepancy is solely due to rounding $\rightarrow$ proceed to process check.
      \item Otherwise $\rightarrow$ mark as incorrect.
   \end{itemize}

3. \textbf{Process check:} Carefully verify each reasoning step:
   \begin{itemize}[itemsep=0pt, parsep=0pt, topsep=0pt]
      \item If there are errors, contradictions, obvious irrelevance to the problem, or the student merely copies the prompt without a substantive solution $\rightarrow$ mark as incorrect.
      \item If the solution process is correct, clear, and consistent $\rightarrow$ mark as correct.
   \end{itemize}

4. \textbf{Format requirements:} If the problem requires a specific format (\textit{e.g.}, units, step-by-step answers, or explicit equations) and the student does not satisfy it $\rightarrow$ mark as incorrect.

5. \textbf{Multiple sub-questions:} If the problem contains multiple sub-questions, the student must answer \emph{all} of them correctly to be marked correct.

6. \textbf{Other cases:} If the above rules do not cover the situation, make an overall judgment from the perspective of whether the student truly knows how to solve the problem.

\textbf{Output requirement:}

Your final output must be \emph{strictly} one of the following:
\begin{itemize}[itemsep=0pt, parsep=0pt, topsep=0pt]
  \item \texttt{<correct> True </correct>}
  \item \texttt{<correct> False </correct>}
\end{itemize}

Now begin:

\texttt{<question>}\\
\{question\}\\
\texttt{</question>}

\texttt{<student\_answer>}\\
\{student\_answer\}\\
\texttt{</student\_answer>}

\texttt{<reference\_answer>}\\
\{reference\_answer\}\\
\texttt{</reference\_answer>}
\end{MyBox}

\subsubsection{RL Ablation Details} \label{sec:rl_ab}
\paragraph{MIS-PO vs. GSPO.}

\begin{figure}[t!]
    \centering
    \begin{subfigure}[b]{1.0\linewidth}
        \centering
        \includegraphics[width=1.0\linewidth]{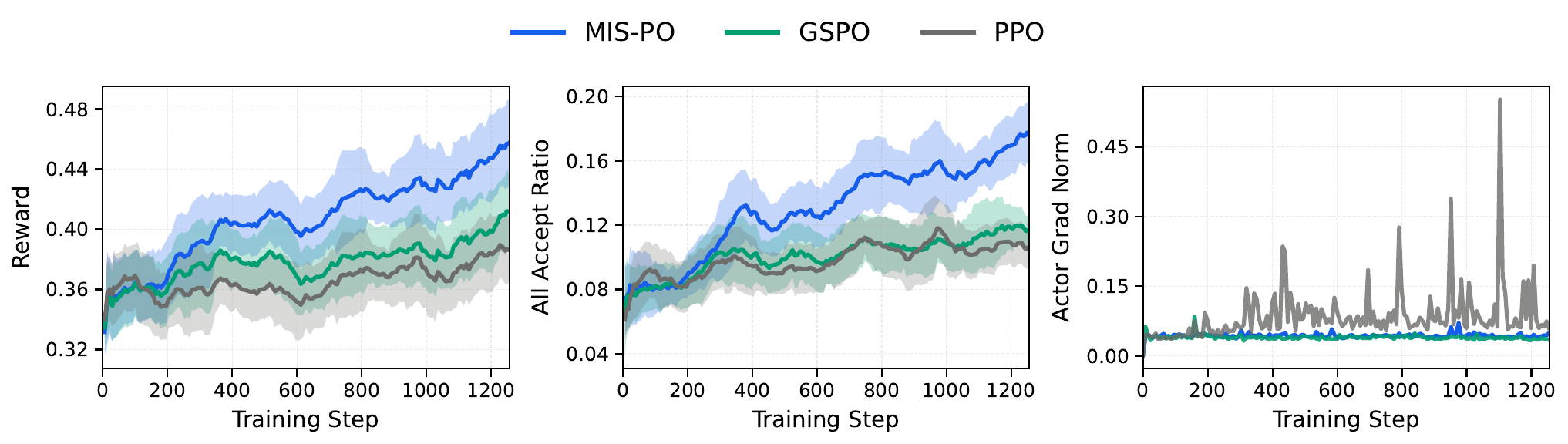}
        \caption{\textbf{Comparison on the dense model.} While GSPO also effectively reduces the variance of the actor gradient norm, its efficiency is inferior to that of MIS-PO. Under the same iteration budget, MIS-PO achieves higher rewards and all acceptance ratio.}
        \label{fig:mispo_vs_gspo_dense}
    \end{subfigure}

    \vspace{1em}

    \begin{subfigure}[b]{1.0\linewidth}
        \centering
        \includegraphics[width=1.0\linewidth]{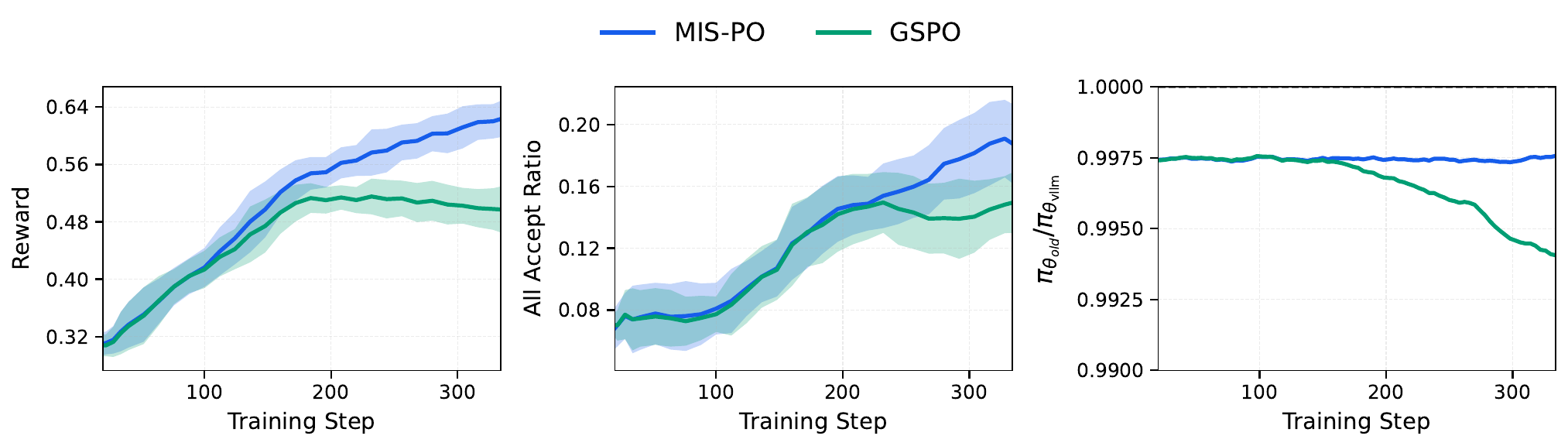}
        \caption{\textbf{Comparison on the MoE model.}
        \textbf{(1) Efficiency:} MIS-PO demonstrates superior sample efficiency, achieving higher rewards with accelerated convergence, whereas GSPO plateaus around iteration 200.
        \textbf{(2) Stability:} GSPO exhibits an increasing training-inference discrepancy during training, quantified by the density ratio $\pi_{\theta_\text{old}}/\pi_{\theta_\text{vllm}}$ (where $\pi_{\theta_\text{vllm}}$ is the rollout policy in the inference backend and $\pi_{\theta_\text{old}}$ is the pre-update policy snapshot in the training backend). Conversely, MIS-PO consistently maintains this discrepancy within a stable range.}
        \label{fig:mispo_vs_gspo_moe}
    \end{subfigure}

    \caption{\textbf{Performance comparison between MIS-PO and GSPO.} The top figure (a) shows results on the dense model, and the bottom figure (b) shows results on the MoE model. MIS-PO consistently outperforms GSPO in both efficiency and stability across different architectures.}
    \label{fig:mispo_vs_gspo_main}
\end{figure}

To rigorously validate the effectiveness of our method, we benchmark MIS-PO against GSPO~\cite{zheng2025group} on both Dense and MoE architectures. We select GSPO as the primary baseline because it represents a competitive strategy for reducing the gradient variance inherent in importance sampling. In our implementation, we extend the original GSPO estimator to the actor-critic setting by integrating its Generalized Importance Sampling mechanism into the actor loss. Specifically, we replace the standard token-level importance sampling ratio with the geometric mean of trajectory-level ratios. The resulting actor loss is formulated as follows ($\gamma=\lambda=1$):
\begin{gather}
    r_\tau(\theta) = \left(\prod_{t=0}^{T-1}\frac{\pi_\theta(a_t|s_t)}{\pi_{\theta_\text{old}}(a_t|s_t)}\right)^{\frac 1T} \\
    \hat{A}_t = \hat{R} - V_\phi(s_t) \\
    \mathcal{L}^{\text{GSPO}}_\text{actor} = -\mathbb{E}_{\tau \sim \pi_{\theta_\text{vllm}}} \left[ \mathbb{I}(x_t) \cdot \mathbb{I}(\bar{\rho}(\tau)) \cdot  \min (r_\tau(\theta)\hat{A}_t, \text{clip}(r_\tau(\theta),1-\epsilon,1+\epsilon)\hat{A}_t) \right]
    \label{eq:gspo_loss}
\end{gather}
To ensure a fair comparison, we apply the same token- and sample-level masking strategies used in MIS-PO to exclude data with significant training–inference mismatches. Regarding the clip ratio $\epsilon$, we conduct a grid search over $\{1,2,3,4\}\times10^{-4}$. We adopt $\epsilon = 10^{-4}$ for all experiments primarily because it achieves the best benchmark performance after 200 RL training steps. Additionally, we observe that this setting yields a clip fraction of approximately 15\%, consistent with the original GSPO~\cite{zheng2025group}.

Figure~\ref{fig:mispo_vs_gspo_main} presents the comparative results. Empirically, MIS-PO demonstrates superior sample efficiency and scalability compared to GSPO. Crucially, MIS-PO effectively constrains the training-inference mismatch within a stable range. This stability proves particularly critical for the large-scale RL training of MoE models, where the baseline GSPO fails to maintain consistent convergence.

\begin{figure}[thbp]
    \centering
    \includegraphics[width=1.0\linewidth]{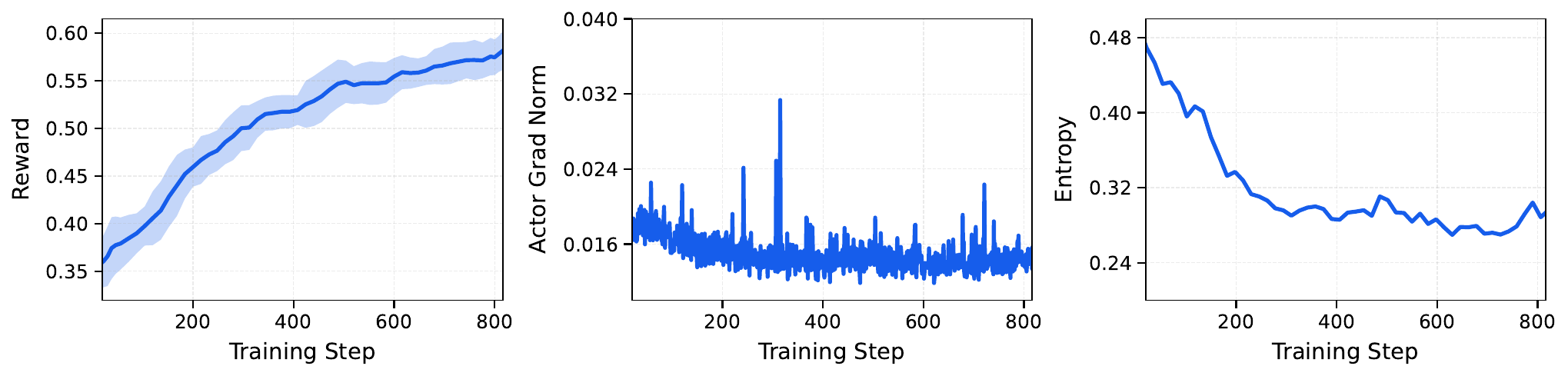}
    \caption{\textbf{Extended training dynamics of MIS-PO on the MoE model.} The metrics include Reward (left), Actor Gradient Norm (middle), and Entropy (right). Notably, the middle panel displays the raw gradient norm without smoothing or downsampling to highlight the stability of the optimization.}
    \label{fig:extend_mispo_moe}
\end{figure}

\paragraph{Extended Training Dynamics on MoE.}
To further validate the scalability of our method, we conduct an extended training run of MIS-PO on the MoE model using a challenging dataset. As illustrated in Figure~\ref{fig:extend_mispo_moe}, the model maintains a continuous upward trend in rewards, stable actor gradient norms, and well-controlled entropy levels. These results empirically confirm that MIS-PO is reliability for large-scale MoE off-policy RL training.

\subsubsection{Search Agent}
Regarding the training architecture, the early client–server one-step off-policy framework is severely bottlenecked by long-tail latency: approximately 5\% of samples accounted for roughly 80\% of the generation cost. However, our observations indicate that the policy exhibits strong robustness to staleness, maintaining stable performance even with a latency of approximately 20 steps. Consequently, we adopt the FullyAsync paradigm, decoupling generation and updates into a completely asynchronous process. Furthermore, to minimize inference overhead during multi-turn interactions, we implement sticky scheduling, where the same session is consistently dispatched to the same node to maximize KV-cache reuse. Overall, this configuration achieves an approximate $10\times$ efficiency gain while maintaining training stability.

Throughout the training process, the FullyAsync paradigm demonstrates robust stability, evidenced by a sustained increase in rewards and a Truncated Importance Sampling (TIS) truncation rate maintained within a controllable range, thereby indicating limited policy drift induced by asynchrony. Notably, we observe that distinct from the limited scalability of ``RL from zero'' regarding training budgets, injecting task-relevant knowledge and tool-use priors during the mid-training phase elicited significantly higher performance gains and a more stable emergence of capabilities during the RL.

\definecolor{tg}{HTML}{228B22}
\definecolor{hg}{gray}{0.9}
\definecolor{sg}{gray}{0.95}

\newcommand{\gain}[1]{\textcolor{glm_text}{\scriptsize~$\mkern-2mu\blacktriangle$#1}}

\begin{table*}[h]
    \centering
    \small
    \setlength{\tabcolsep}{2pt}
    \setlength{\extrarowheight}{3pt}
    \resizebox{\textwidth}{!}{
    \begin{tabular}{l | c c c c c c}
        \toprule
        \rowcolor{hg} \textbf{Model} & \makecell{\textbf{BrowseComp}} & \makecell{\textbf{BrowseComp-ZH \ \ \ \ \ \ }}&\makecell{\textbf{GAIA\ \ \ \ \ \ \ \ \ }} & \makecell{\textbf{xbench}\\\small \textbf{DeepSearch-2505}} & \makecell{\textbf{xbench}\\\small \textbf{DeepSearch-2510}} & \makecell{\textbf{Avg Gain}} \\
        \midrule
        \rowcolor{sg}\multicolumn{7}{l}{\textsc{\textbf{Agent $\Delta$avg@3 (Metric: Pass Rate \%)}}} \\
        Step 3.5 Flash* & 1.5 \gain{50.1}  & 25.0 \gain{41.9} & 17.0 \gain{67.5} & 26.0 \gain{57.7} & 11.3 \gain{42.7} & \best{52.0} \\
        Kimi K2-Thinking* & 3.6 \gain{37.9}  & 23.8 \gain{38.5} & 18.8 \gain{36.6} & 28.7 \gain{39.3} & 14.3 \gain{27.0} & 35.9 \\
        Kimi K2.5* & 7.4 \gain{53.2}  & 40.3 \gain{22.0} & 26.7 \gain{49.2} & 36.0 \gain{40.3} & 19.7 \gain{36.6} & 40.2 \\
        DeepSeek V3.2         & 8.1 \gain{43.3}  & 41.2 \gain{23.8} & 23.4 \gain{51.7} & 35.7 \gain{41.3} & 18.7 \gain{30.6} & 38.1 \\
        GLM-4.7               & 3.4 \gain{48.6}  & 30.2 \gain{36.4} & 19.6 \gain{26.5} & 29.7 \gain{34.6} & 19.3 \gain{23.4} & 33.9 \\
        MiniMax M2.1          & 1.3 \gain{46.1}  & 10.1 \gain{37.7} & 15.4 \gain{30.9} & 18.7 \gain{46.6} & 6.0 \gain{36.3}  & 39.5 \\
        MiMo-V2 Flash         & 0.9 \gain{44.5}  & 12.9 \gain{38.3} & 12.9 \gain{42.3} & 19.7 \gain{49.6} & 6.3 \gain{13.7}  & 37.7 \\
        Gemini 3.0 Pro           & 25.2 \gain{12.6} & \na              & 32.1 \gain{44.5} & 45.0 \gain{32.0} & \na              & 29.7 \\
        Claude Sonnet 4.5      & 1.4 \gain{22.7}  & 21.2 \gain{19.6} & 16.2 \gain{54.7} & 24.7 \gain{42.6} & 7.3 \gain{37.7}  & 35.5 \\
        \bottomrule
    \end{tabular}
    }
    \caption{\textbf{Impact of Tool Usage on Agent Performance.} Each cell displays the \textbf{Baseline Score} (internal knowledge only) followed by the \gain{\textbf{Performance Gain}} achieved by enabling search tools. The final score is the sum of both values. \textbf{Avg Gain} highlights the model's ability to leverage external information to improve results. Models marked with * denote tool results measured under a 256K setting; the setting for other models is unspecified.}
    \label{tab:tool_gain}

\end{table*}

\vspace{+1cm}
\noindent\textbf{Discussion.}
To rigorously evaluate agentic competence isolated from parametric memorization, we focus on the \emph{tool-usage gain}, defined as:
\[
\Delta_{\text{tool}} = \text{Score}_{\text{with tools}} - \text{Score}_{\text{no tools}}
\]
This metric decouples the model's inherent knowledge from its ability to dynamically leverage external tools. As detailed in Table~\ref{tab:tool_gain}, \textbf{Step 3.5 Flash} demonstrates the most robust capability to leverage external information, achieving the highest average gain ($52.0$) and leading significantly on complex benchmarks such as GAIA and xbench-DeepSearch.

This distinction is critical because high absolute scores on benchmarks like BrowseComp can sometimes stem from strong internalized knowledge rather than effective search strategies. A smaller $\Delta_{\text{tool}}$ in a high-performing model may ambiguously indicate either high efficiency (the model already ``knows'' the answer) or a failure to effectively utilize tools to improve results. Conversely, a large $\Delta_{\text{tool}}$ explicitly signals the model's proficiency in bridging knowledge gaps through retrieval.
Therefore, we argue that future optimization should not merely chase higher absolute scores (``benchmark grinding''), but should aim to maximize this $\Delta_{\text{tool}}$ in long-context, evidence-critical scenarios. This ensures the agent is truly mastering the \emph{process} of information retrieval and reasoning, rather than overfitting to static knowledge or benchmark artifacts.

\subsection{Tool-integrated Reasoning and Parallel Reasoning}

In this section, we introduce two primary methodologies for test-time scaling in Step 3.5 Flash: tool-integrated reasoning and parallel reasoning.
\paragraph{Tool-integrated Reasoning}
For complex reasoning tasks, we integrate the model with a Python interpreter to facilitate tool-assisted reasoning. In this framework, the model operates within a sandbox to iteratively think and execute code for computational, simulation, and visualization purposes. In our experiments, we evaluate on AIME 2025, HMMT 2025, IMO-AnswerBench, GPQA, HLE$_{\text{text}}$, and ARC-AGI-1 with a 100-turn limit.
As shown in Table \ref{tab:tir-results}, tool-integrated reasoning significantly enhances performance across challenging mathematics, STEM, and puzzle benchmarks, highlighting the advanced agentic reasoning capabilities of Step 3.5 Flash.

\begin{table}[h]
\centering
\begin{tabular}{lcc}
\toprule
\rowcolor{header_gray}
\textbf{Benchmark} & \textbf{Step 3.5 Flash} & \textbf{Step 3.5 Flash w. Python} \\
\midrule
AIME 2025           & 97.3 & \textbf{99.8} (+2.5) \\
HMMT 2025 Feb.      & 98.4 & \textbf{98.7} (+0.3) \\
HMMT 2025 Nov.      & 94.0 & \textbf{98.0} (+4.0) \\
IMO-AnswerBench      & 85.4 & \textbf{86.7} (+1.3) \\
GPQA-Diamond        & 83.5 & \textbf{84.4} (+0.9) \\
HLE$_{\text{text}}$ & 23.1 & \textbf{26.5} (+3.4) \\
ARC-AGI-1           & 54.8 & \textbf{56.5} (+1.7) \\
\bottomrule
\end{tabular}
\caption{Comparison of Step 3.5 Flash and Step 3.5 Flash w. Python.}
\label{tab:tir-results}
\end{table}

\paragraph{Tool-integrated Parallel Reasoning}
We present a preliminary exploration of extending PaCoRe to a multi-turn interactive environment.
By design, PaCoRe preserves the standard LLM message interface. This compatibility allows for seamless integration into existing agentic frameworks that utilize multi-turn tool interaction.
To adapt PaCoRe to this setting, we implement a state-aware input serialization protocol as shown in Table \ref{tab:pacore-tool-template}.

We evaluate this approach on the GPQA and HLE$_{\text{text}}$ benchmarks using Step 3.5 Flash equipped with a Python interpreter. As shown in Table~\ref{tab:tool-results}, extending parallel reasoning to these agentic loops yields significant performance improvements over the standard reasoning baseline. These findings demonstrate that PaCoRe effectively generalizes to environments requiring interactive feedback, highlighting a promising avenue for agentic test-time scaling.

\begin{table}[h]
\centering
\begin{tabular}{lcc}
\toprule
\rowcolor{header_gray}
\textbf{Benchmark w. Python} & \textbf{Step 3.5 Flash} & \textbf{Step 3.5 Flash + PaCoRe} \\
\midrule
GPQA-Diamond        & 84.4 & \textbf{85.7} (+1.3) \\
HLE$_{\text{text}}$ & 26.5 & \textbf{28.2} (+1.7) \\
\bottomrule
\end{tabular}
\caption{Comparison of Step 3.5 Flash w. Python and the same model with PaCoRe test-time scaling.}
\label{tab:tool-results}
\end{table}

\input{src/tables/app_pacore_prompt_template_table}

\section{Detailed Evaluation Protocols and Prompts}
\label{sec:post_training_eval_appendix}

This section provides the implementation details for our evaluation suite. We outline the specific prompt templates, few-shot configurations, and the judge models employed across different benchmarks. For complex metrics, such as those used in long-context or reasoning tasks, we also detail the underlying calculation logic and scoring criteria to ensure reproducibility. In the templates provided below, \texttt{\{question\}} denotes the placeholder for the textual problem description, while other placeholders (e.g., \texttt{\{test\}}, \texttt{\{context\}}) represent task-specific information.

\subsection{Evaluation Details of Pre-trained Models}

\subsubsection{General language understanding and reasoning benchmarks}

\paragraph{BBH.} We use the official CoT-prompts~\footnote{https://github.com/suzgunmirac/BIG-Bench-Hard/tree/main/cot-prompts} of BBH~\cite{suzgun2022challengingbbh}, with only "Q:" and "A:" replaced by "Problem:" and "Solution:" as follows:

\begin{MyBox}{}

    Problem:\\
    \{question\}\\
    \\
    Solution:\\
\end{MyBox}

\paragraph{MMLU.} We use the official evaluation metric of MMLU~\cite{hendrycks2020mmlu} with 5-shot. We employ the following task-specific system prompt:

\begin{MyBox}{}

    The following are multiple choice questions (with answers) about \{category\}.
\end{MyBox}

The corresponding question prompt is structured as follows:

\begin{MyBox}{}

    \{question\}\\
    Answer:
\end{MyBox}

\paragraph{MMLU-Redux.} We use the official evaluation metric of MMLU-Redux~\cite{gema2024donewithmmlu} with 5-shot. and employ the following question prompt:

\begin{MyBox}{}

    Answer the question and place the option (A/B/C/D...) inside \textbackslash boxed\{\}.\\
    \{question\}
\end{MyBox}

\paragraph{MMLU-Pro.} \label{sec:MMLU_pro_pretrain_evaluation} We follow the official evaluation metric of MMLU-Pro~\cite{wang2024mmlupro} with 5-shot. All evaluations use the following system prompt:

\begin{MyBox}{}

    The following are multiple choice questions (with answers) about \{category\}. Think step by step and then output the answer in the format of "The answer is (X)" at the end.
\end{MyBox}

The question prompt is structured as follows, with a deliberate trailing space after the final period:

\begin{MyBox}{}

    Question: \{question\}\\
    Answer: Let's think step by step.
\end{MyBox}

Notably, we observe that a subset of the original MMLU-Pro dataset (470 out of 12,102 questions) contained an inconsistent leading space before the ground-truth options. We explicitly remove these spaces to mitigate potential formatting bias and ensure evaluation consistency.

\paragraph{HellaSwag.} We use the official evaluation metric of HellaSwag~\cite{zellers2019hellaswag} with 10-shot. We employ the following question prompt:

\begin{MyBox}{}

    Question: \{question\}\\
    A. \{option\_0\}\\
    B. \{option\_1\}\\
    C. \{option\_2\}\\
    D. \{option\_3\}\\
    Answer:
\end{MyBox}

\paragraph{WinoGrande.} We use the official evaluation metric of WinoGrande~\cite{sakaguchi2019winogrande} with 5-shot. The question prompt is structured to present the binary choices clearly:

\begin{MyBox}{}

    Question: \{question\}\\
    Options:\\
    A. \{option\_0\}\\
    B. \{option\_1\}\\
    \\
    Answer:
\end{MyBox}

\paragraph{GPQA.} We use the official evaluation metric of GPQA~\cite{rein2023gpqa} with 5-shot. The question prompt is structured to present the choices clearly:

\begin{MyBox}{}
\small
    Question: \{question\}\\
    Options:\\
    A. \{option\_0\}\\
    B. \{option\_1\}\\
    C. \{option\_2\}\\
    D. \{option\_3\}\\
    \\
    Answer: Let’s think step by step.
\end{MyBox}

\paragraph{SuperGPQA.} We use the official evaluation metric of SuperGPQA~\cite{du2025supergpqa} with 5-shot. The question prompt follows a Chain-of-Thought (CoT) structure, where each few-shot example includes a step-by-step derivation leading to the final answer:

\begin{MyBox}{}

    Question:\\
    \{question\}\\
    \\
    Answer: Let's think step by step.
\end{MyBox}

\paragraph{SimpleQA.} We use the official evaluation metric of SimpleQA~\cite{openai2024simpleqa} with 5-shot. As SimpleQA requires open-ended short answers, we employ an LLM-based judgement for evaluation, specifically using gpt-oss-120b\cite{openai2025gptoss120bgptoss20bmodel} as the judge model. The question prompt is formatted as a concise query:

\begin{MyBox}{}

    Question: \{question\} Answer:
\end{MyBox}

\subsubsection{Mathematics reasoning benchmarks}

\paragraph{GSM8K.} We use the official evaluation metric of GSM8K~\cite{cobbe2021gsm8k} with 8-shot. The question prompt is designed to elicit CoT reasoning by using the following template:

\begin{MyBox}{}

    Q: \{question\}\\
    A: Let’s think step by step.
\end{MyBox}

\paragraph{MATH.} We use the official evaluation metric of MATH~\cite{hendrycks2021math} with 4-shot. The question prompt is structured with explicit problem and solution delimiters:

\begin{MyBox}{}

    Problem:\\
    \{question\}\\
    \\[\baselineskip]
    Solution:
\end{MyBox}

\subsubsection{Coding benchmarks}
\paragraph{HumanEval.} We use the official evaluation metric of HumanEval\cite{chen2021evaluatinglargelanguagemodels} with 3-shot. The question prompt is structured with three ground-truth examples to provide contextual guidance for code generation:

\begin{MyBox}{}
\small
\begin{verbatim}
# Below are the ground-truth solutions:

def add_two_numbers(a, b):
""" Given two numbers a and b, return the sum of a and b. """
    # get the sum of a and b
    sum_of_a_and_b = a + b
    return sum_of_a_and_b

def reverse_list(some_list: list) -> list:
    """ Given a list, return a reversed copy of the list. """
    new_list = []
    # iterate over the list
    for item in some_list:
        # insert item into new list
        new_list.insert(0, item)
    return new_list

def fast_reverse_list(some_list: list) -> list:
    """ Given a list, return a reversed copy of the list. Be fast! """
    # use faster built-in reverse
    some_list.reverse()
    return some_list
{question}
\end{verbatim}
\end{MyBox}

\paragraph{MBPP.} We follow the official evaluation metric of MBPP\cite{austin2021programsynthesislargelanguage} with 3-shot.

\paragraph{HumanEval+.} We follow the official evaluation metric of HumanEval+~\cite{liu2023evalplus} with 3-shot.

\paragraph{MBPP+.} We use the official evaluation metric of MBPP+~\cite{liu2023evalplus} with zero-shot. We employ a structured instruction prompt that specifies the task requirements and includes a sample test case for alignment:

\begin{MyBox}{}

You are an expert Python programmer, and here is your task: \\
  \{question\} \\
  Your code should pass the test: \\
  \{test\} \\
  Here is the corresponding code: \\
  \textasciigrave\textasciigrave\textasciigrave python
\end{MyBox}

\paragraph{MultiPL-E.} We use the official evaluation metric of MultiPL-E~\cite{cassano2022multipl} with zero-shot. We follow the official test cases to judge the generated code.

\subsubsection{Chinese understanding benchmarks}

\paragraph{C-Eval.} We use the official evaluation metric of C-Eval~\cite{huang2023ceval} and add a 5-shot setting. We employ the following system prompt:

\begin{CJK*}{UTF8}{gbsn}
\begin{MyBox}{}

    \raggedright
    你是一个中文人工智能助手，以下是中国关于 \{category\}考试的单项选择题，请选出其中的正确答案。
\end{MyBox}
\end{CJK*}

The corresponding question prompt is structured as follows:

\begin{CJK*}{UTF8}{gbsn}
\begin{MyBox}{}

    \{question\} \\
    答案：
\end{MyBox}
\end{CJK*}

\paragraph{CMMLU.} We use the official evaluation metric of CMMLU~\cite{li2023cmmlu} and add a 5-shot setting. We employ the following system prompt:

\begin{CJK*}{UTF8}{gbsn}
\begin{MyBox}{}

    \raggedright
    你是一个中文人工智能助手，以下是中国关于 \{category\}考试的单项选择题，请选出其中的正确答案。
\end{MyBox}
\end{CJK*}

The corresponding question prompt is structured as follows:

\begin{CJK*}{UTF8}{gbsn}
\begin{MyBox}{}

    \{question\} \\
    答案：
\end{MyBox}
\end{CJK*}

\paragraph{C-SimpleQA.}  We use the official evaluation metric and LLM-based judgement protocols of Chinese SimpleQA~\cite{he2024chinesesimpleqa}. We add a 5-shot setting and use gpt-oss-120b\cite{openai2025gptoss120bgptoss20bmodel} as the judge model. We employ the following question prompt:

\begin{CJK*}{UTF8}{gbsn}
\begin{MyBox}{}

    问题：\{question\} \\
    答案：
\end{MyBox}
\end{CJK*}

\subsection{Evaluation Details of Post-Trained Models}
\label{sec:post_training_eval_appendix}
In this section, we detail the evaluation protocols used to assess the post-trained models across a diverse set of agentic tasks.
Our evaluations span both code-centric and general-purpose agent settings, covering software engineering, terminal interaction, deep search, research workflows, and real-world tool use.
We report standardized metrics under carefully controlled environments and inference budgets to ensure fair, stable comparisons across benchmarks.

\subsubsection{Reasoning benchmarks}

\paragraph{CF-Div2-Stepfun.} Recent studies and advanced benchmarks emphasize the critical need to evaluate models on fresh, competition-level problem~\cite{zheng2025livecodebench,anonymous2026autocode}. We evaluate the competitive programming capabilities of our model using a custom CodeForces Div. 2 Benchmark~\footnote{https://huggingface.co/datasets/stepfun-ai/CF-Div2-Stepfun}. The benchmark comprises 53 problems sourced from official CodeForces Div.2 contests held between September 2024 and February 2025. We develop an offline evaluation framework that utilizes a local grading mechanism as an alternative to real-time online submissions. We try to construct test cases similar to the original test cases. Specifically, we first generate enough small-scale test cases for evaluation correctness coverage, then add randomized data for large-scale testing. Finally, we performed adversarial construction of edge cases by analyzing common error patterns and "hacked" submissions from actual users. Some edge cases are also auto-generated by the stress testing technique, which keeps generating countless test cases until one can distinguish failed submissions from correct submissions. To validate the reliability of this benchmark, we run both correct and representative failed submissions selected from the original contests. Our evaluator correctly identifies 100\% of the accepted submissions as "Passed", while 92.45\% of the failed submissions are accurately flagged.
~\label{cf-div2-step}

\begin{table}[H]
    \centering
    \small
    \setlength{\tabcolsep}{8pt}
    \begin{tabular}{l ccc c}
        \toprule
        & \multicolumn{3}{c}{\textbf{Accuracy (avg@8)}} & \textbf{Codeforces C++} \\
        \cmidrule(lr){2-4}
        \textbf{Model} & \textbf{C++} & \textbf{Python} & \textbf{Java} & \textbf{pass@8 Rating} \\
        \midrule
        \ourmodel & \textbf{86.1\%} & \textbf{81.5\%} & 77.1\% & \textbf{2489} \\
        \addlinespace[2pt]
        Deepseek V3.2 & 81.6\% & 66.5\% & 80.7\% & 2319 \\
        GLM-4.7 & 74.1\% & 63.0\% & 70.5\% & 2156 \\
        Kimi K2-Thinking & 67.9\% & 60.4\% & 58.5\% & 1976 \\
        Minimax-M2.1 & 59.0\% & 46.4\% & 58.0\% & 1869 \\
        Mimo-V2 Flash & 46.9\% & 43.6\% & 39.6\% & 1658 \\
        \midrule
        Gemini 3.0 Pro & 83.5\% & 74.1\% & \textbf{81.6}\% & 2397 \\
        Claude Opus 4.5 & 72.2\% & 68.4\% & 68.9\% & 2100 \\
        \bottomrule
    \end{tabular}
    \caption{Full evaluation results of variable models in CF-Div2-Stepfun.}
    \label{tab:codeforces-full}
\end{table}

We sample 8 responses for each problem and report the average accuracy. The user prompt utilized for this process is:

\begin{MyBox}{}

You are a coding expert. Given a competition-level coding problem, you need to write a \{LANGUAGE\} program to solve it. You may start by outlining your thought process. In the end, please provide the complete code in a code block enclosed with \`{}\`{}\`~\`{}\`{}\`{}.
\newline
\{question\}
\end{MyBox}

The compilation and execution commands for C++, Python, Java are given below:

\begin{MyBox}{}
\small
g++ -std=c++20 -fno-asm -fsanitize=bounds -fno-sanitize-recover=bounds --static -O2 -DONLINE\_JUDGE -o code.exe code.cpp

./code.exe
\end{MyBox}

\begin{MyBox}{}
\small
python3 code.py
\end{MyBox}

\begin{MyBox}{}
\small
javac -J-Xmx544m \{JAVA\_CLASS\_NAME\}.java

java -XX:+UseSerialGC -Xmx544m -Xss64m -DONLINE\_JUDGE \{JAVA\_CLASS\_NAME\}
\end{MyBox}

To maintain consistency with competitive programming norms and avoid the inconsistent overhead associated with JIT "warm-up" periods, we use the standard Python interpreter with a double time limit rather than PyPy\footnote{https://pypy.org/}. We apply this same double time limit to all Java submissions.

While the Table~\ref{tab:codeforces-full} reports raw accuracy, we recognize that problem difficulty varies significantly. Therefore, rating scores provide more robust metrics. Although frameworks like CodeELO~\cite{quan2025codeelobenchmarkingcompetitionlevelcode} can calculate competitive ratings, current top-tier models perform so effectively in Division 2 contests that their ratings may result in statistical outliers. Furthermore, we adopt a simplified rating calculation that disregards submission time penalties by assuming all solutions are submitted at the onset of the contest. While this approach deviates from empirical competitive scenarios and may result in ratings that are not directly comparable to human participants, it provides a standardized benchmark for consistent cross-model comparison.

\paragraph{LiveCodeBench-v6.} We use the official evaluation method of LiveCodeBench\cite{jain2024livecodebench}. We employ the following system prompt:

\begin{MyBox}{}

    You are an expert Python programmer. You will be given a question (problem specification) and will generate a correct Python program that matches the specification and passes all tests.
\end{MyBox}

The corresponding question prompt is structured as follows:

\begin{MyBox}{}

    \#\#\# Question: \\
    \{question\} \\
    \#\#\# Format: You will use the following starter code to write the solution to the problem and enclose your code within delimiters. \\
    \textasciigrave\textasciigrave\textasciigrave{} python \\
    \{starter\_code\} \\
    \textasciigrave\textasciigrave\textasciigrave\\
    \#\#\# Answer: (use the provided format with backticks)
\end{MyBox}

\paragraph{AIME 2025.} We use the official evaluation method of AIME 2025\cite{AIME25} with repeat@64. We employ the following question prompt:

\begin{MyBox}{}

    Answer the question and place the answer inside \textbackslash boxed \{\} with MathTeX format. \\
    \{question\}
\end{MyBox}

\paragraph{HMMT 2025 Feb./Nov.} We use the official evaluation method of HMMT 2025~\cite{hmmt25} with repeat@64. We employ the following question prompt:

\begin{MyBox}{}
    Answer the question and place the answer inside \textbackslash boxed \{\} with MathTeX format. \\
    \{question\}
\end{MyBox}

\paragraph{IMO-AnswerBench.} We use the official evaluation method of IMO-AnswerBench\cite{luong2025towards} with repeat@64. We employ the following question prompt:

\begin{MyBox}{}

    Answer the question and place the answer inside \textbackslash boxed \{\} with MathTeX format. \\
    \{question\}
\end{MyBox}

\paragraph{MMLU-Pro.} We use the official evaluation method of MMLU-Pro~\cite{wang2024mmlupro}. The processing of dataset remains consistent with our pre-training MMLU-Pro evaluation methodology (see Appendix \ref{sec:MMLU_pro_pretrain_evaluation} for details).

\begin{MyBox}{}

Answer the question and place the option (A/B/C/D...) inside \textbackslash boxed\{\}.\newline
\newline
\{question\}
\end{MyBox}

\paragraph{GPQA-Diamond.} We use the official evaluation method of GPQA-Diamond~\cite{rein2023gpqa}. We employ the following question prompt:

\begin{MyBox}{}

Answer the question and place the option (A/B/C/D...) inside \textbackslash boxed\{\}.\newline
\newline
\{question\}
\end{MyBox}

\paragraph{\text{HLE}$_\text{text}$.} We use the official evaluation metric and LLM-based judgement protocols of HLE. We use gpt-oss-120b~\cite{openai2025gptoss120bgptoss20bmodel} as the judge model.

\subsubsection{Code Agent benchmarks}\label{sec:code-agent}

\paragraph{SWE-Bench.}
SWE-Bench Verified~\cite{swe_verified} is a high-quality subset of the original SWE-bench dataset, consisting of 500 software engineering tasks rigorously validated by human expert developers to ensure reliable and accurate evaluation. SWE-Bench Multilingual extends the original benchmark to a diverse set of 300 real-world software engineering tasks across 9 programming languages.

We test the software engineering agent ability of \ourmodel on SWE-Bench Verified and SWE-Bench Multilingual using our internal agent infrastructure, which is built upon the described session-router architecture. For each evaluation instance, we provision a containerized session orchestrated via Kubernetes. We then perform environment initialization specific to SWE-Bench, which includes removing future commits to prevent data leakage, as well as configuring network proxies and critical system settings.
Regarding the agent scaffold, we adopted the OpenHands~\cite{wang2024openhands} CodeAct Agent framework, which is widely used in the research community. We enabled a default suite of four tools: execute\_bash, str\_replace\_editor, finish, and think. The max interactive turns is set to 350.

Given the resource-intensive nature of compiled languages, we allocate 12GB of memory for the multilingual setting, whereas the verified instances are restricted to a 4GB limit. In evaluations, the tool execution timeout is set to 1200s, and the model inference parameters are: temperature=1, top-p=0.95. Following the above settings, \ourmodel reach 74.4\% on SWE-Bench Verified, and 67.4\% on SWE-Bench Multilingual benchmark with an average score of 4 repeat of runnings. We also cross-evaluate \ourmodel on other popular agent scaffolds: SWE-Agent~\cite{yang2024swe} with the original agent pipeline settings achieving 74.2\% accuracy on SWE-Bench Verified, and standard Claude Code~\footnote{https://github.com/anthropics/claude-code} environment scoring 72.0\% with an extended time limit of 4 hours for each instance and no time limit for single tool execution.

\paragraph{Terminal-Bench 2.0.}

We test the Terminal-Bench benchmark~\cite{terminalbench2025} within remote task-independent containers. We limit the container memory to 16GB. We have deployed an internal Artifactory repository and update the default package sources for all Docker containers. During session creation and dependency installation of the testing phases, the system will retry multiple times if an error occurs. To streamline the system-agent interaction, we modify the Terminus 2 framework so that it automatically interrupts timed-out commands and prevents subsequent commands in the same round from executing, returning a timeout warning to the agent. Accordingly, we modify the command duration control part of the original system prompt:

\begin{MyBox}{}

Keystroke duration sets the command hard timeout. The system automatically interrupts timed-out commands and prevents subsequent commands in the same round from executing. You can simply continue with your next round - no special action is required.
\end{MyBox}

During inference, we cap the model's single-turn output at 64k and the maximum context window at 256k for all interactions. The thinking process will be preserved in the multi-round history. If the model output exceeds the 256k context window limit, we execute a pruning context management: Keep the problem statement and the last 50\% of history before retrying. We use the inference parameters of top-p=0.95 and temperature=1. The interaction protocol is primarily conducted using XML-formatted structured responses. The agent is limited to 200 interaction rounds and will proceed directly to the testing phase once this limit is reached. The total time limit for interaction and testing is 6 hours.

To ensure consistency, we verify and refine each task's checker against its problem statement~\footnote{https://huggingface.co/datasets/zai-org/terminal-bench-2-verified}, which improved overall accuracy by approximately 1.5\%. Each task is executed across 8 trials. Notably, 88.6\% of successful trajectories are completed within 30 interactions. The final pass@8 stands at 67/89, with an avg@8 of 50.98\%. Our agent achieves a 100\% success rate across all 8 trials in 23 out of 89 tasks. In the successful trajectories, 9.41\% of the runs triggered history pruning to manage context limits.

\begingroup
\begin{table}[H]
    \centering
    \small
    \begin{tabular}{l c c c c c}
        \toprule
        \rowcolor{header_gray}
        \textbf{Setting} & \textbf{Max Output} & \textbf{Max Round} & \textbf{Timeout} & \textbf{Context Management} & \textbf{Avg@8} \\
        \midrule
        Baseline & 64k & 200 & 6h & \checkmark & \textbf{50.98\%} \\

        Limit 16k & 16k & 200 & 6h & \checkmark & 48.03\% \\
        Limit 16k w/o Pruning & 16k & 200 & 6h & $\times$ & 45.22\% \\

        Rounds 100 & 64k & 100 & 6h & \checkmark & 50.42\% \\
        Timeout 2h & 64k & 200 & 2h & \checkmark & 49.72\% \\
        \bottomrule
    \end{tabular}
    \caption{Ablation study of inference constraints on Terminal-Bench 2.0.}
    \label{tab:ablation_study}
\end{table}
\endgroup

The ablation study shows that Limit 16k causes the largest performance drop because the model’s long reasoning for complex tasks often exhausts the token limit before it can output the terminal commands. The further decline to 45.22\% when disabling context management under the 16k limit. Meanwhile, Rounds 100 has minimal impact as most tasks finish early. The Timeout 2h decrease reflects that certain tasks involving model training, heavy compilation, or complex environment configuration require more time to complete.

\subsubsection{General Agent benchmarks}

\paragraph{Deep Search.}

We evaluate our agent's deep search capabilities on multiple benchmarks \ (e.g, \textbf{BrowseComp~\cite{wei2025browsecomp}, BrowseComp-ZH~\cite{zhou2025browsecompzhbenchmarkingwebbrowsing}, GAIA~\cite{mialon2023gaiabenchmarkgeneralai}, xbench-DeepSearch~\cite{chen2025xbench}}). The results reported in Table~\ref{fig:post-training-main-table} are based on the avg@3 metric; GPT-5.2 xHigh uses avg@1. The agent is equipped with a core toolset including:
\begin{itemize}
    \item \textbf{search}: Executes multiple search queries in parallel.
    \item \textbf{visit}: Analyzes the content of the webpage to answer specific questions based on LLM.
    \item \textbf{google\_scholar}: Search for academic articles and technical literature.
    \item \textbf{python\_interpreter}: Runs Python code for calculations and data analysis.
     \item \textbf{file}: Downloads and saves files from direct URLs.
\end{itemize}

During inference, we employ a 256k-token context window with no limit on the maximum generation length. Inference is conducted with top-p = 0.95, temperature = 1.0, and presence penalty = 1.1, allowing for an execution budget of up to 400 steps.

The detailed system prompts for the agent and the LLM judge are consistent with the configurations provided in the GitHub repository associated with~\cite{hu2025stepdeepresearchtechnicalreport}.

\paragraph{BrowseComp (w. Ctx Manage).}

The BrowseComp (w. Ctx Manage) result of 69.0 reported in Table~\ref{fig:post-training-main-table} corresponds to the discard-all methodology evaluated on the full BrowseComp dataset. This approach, same as DeepSeek V3.2~\cite{deepseekai2024deepseekv32}, is triggered when the context length exceeds predefined thresholds, at which point the agent discards its entire context and reinitializes the operational loop. Under a maximum iteration constraint of 1000 steps, this strategy employs a context length threshold of 72k tokens for BrowseComp and 41k tokens for BrowseComp-ZH.

We also evaluate various context management strategies on a subset of 200 instances from BrowseComp, including Summary, Keep-first\&last$K$, Discard-all, and Multi-agent orchestration. As shown in Table~\ref{tab:context_manager}, our model demonstrates robust adaptability across these diverse paradigms. Among single-agent strategies, Discard-all yields a competitive $66.0\%$ accuracy. We posit that Discard-all functions as a test-time pass@$k$ strategy, forcing the model to re-reason from scratch until a self-verified path is found. The performance follows a clear hierarchy: Multi-agent ranks highest by leveraging a master agent to decompose tasks and dispatch specialized agents for parallel reasoning, followed by Discard-all, Keep-first\&last$K$ and Summary —closely aligns with the increase in real steps. This alignment reflects a direct trade-off between inference cost (number of steps) and accuracy, suggesting that intensive context management effectively converts increased computation into superior performance.

\begin{table}[h]
\centering

\begin{tabular}{lcc}
\toprule
\rowcolor{header_gray}
\textbf{Method} & \textbf{Accuracy (\%)} & \textbf{Real Steps} \\
\midrule
Step 3.5 Flash & 49.5 & 86 \\
+ Summary & 57.0  & 131 \\
+ Keep-first\&lastK & 58.0  & 244 \\
+ Discard-all & 66.0  & 302 \\
+ Multi-Agent & 68.5 & 721 \\
\bottomrule
\end{tabular}
\caption{Evaluation results of context manager methods.}
\label{tab:context_manager}
\end{table}

\paragraph{\textsc{ResearchRubrics}.}

To evaluate deep research capabilities, we utilize the \textsc{ResearchRubrics}~\cite{sharma2025researchrubrics} benchmark. This dataset comprises 101 domain-diverse research tasks, each accompanied by 20--43 expert-written, fine-grained scoring criteria that assess factual accuracy, reasoning soundness, and clarity. We benchmark performance against two representative system families: commercial agent systems and ReAct agents.

For commercial agents, we collect reports via their official web interfaces (captured Dec 2--15, 2025) under default configurations. As shown in Table~\ref{tab:commercial_systems}, the leading commercial system (Gemini DeepResearch) achieves an aggregated score of 63.69.

\begin{table}[h]
    \centering
    \renewcommand{\arraystretch}{1.2}
    \begin{tabular}{lc}
        \toprule
        \rowcolor{header_gray}
        \textbf{Agent System} & \textbf{Score} \\
        \midrule
        Gemini DeepResearch & 63.69 \\
        OpenAI DeepResearch & 60.67 \\
        Kimi Researcher & 53.67 \\
        MiniMax Agent Pro & 51.85 \\
        Qwen DeepResearch & 49.24 \\
        \bottomrule
    \end{tabular}
    \caption{Performance of Commercial Agent Systems on the \textsc{ResearchRubrics} benchmark.}
    \label{tab:commercial_systems}
\end{table}

For ReAct agents, detailed performance comparisons are presented in Table~\ref{fig:post-training-main-table}. Our model achieves a score of \textbf{65.3}, surpassing the complex, proprietary commercial baselines. Notably, when evaluating Gemini 3.0 Pro within our standardized ReAct framework, we observe a score of 50.1. We attribute this performance gap to insufficient search depth when addressing open-ended research questions; the model tends to rely on internal parametric knowledge rather than perform extensive external retrieval. Consequently, the generated reports lack comprehensiveness, failing to adequately cover the user's implicit criteria.

We standardize the execution environment for ReAct agents with a maximum of 30 reasoning turns and a per-turn output limit of 16k tokens. For inference parameters, other API-based models use their default settings, and our model is configured with a temperature of $1$ and top-p=$0.95$. All outputs are subsequently appraised by an LLM judge using a ternary grading for each criterion. To support the end-to-end research workflow, our ReAct framework provides access to the following suite of tools:

\begin{itemize}
    \item \textbf{batch\_web\_surfer}: For concurrent web searching and multi-page browsing.
    \item \textbf{file}: For robust file operations, including reading, writing, and iterative editing.
    \item \textbf{file\_parser}: For converting files into Markdown format.
    \item \textbf{shell}: For interactive command execution and environment interaction.
    \item \textbf{todo}: For dynamic task state management and tracking.
    \item \textbf{tmux}: For simulating a multiplexed terminal environment with persistent sessions and scrollback history.
\end{itemize}

\paragraph{$\tau^2$-Bench.}

$\tau^2$-Bench\cite{tau2bench2025} is an agentic benchmark that evaluates general tool-use capability in three customer service domains: airline, retail, telecom. We evaluate Step 3.5 Flash using the official settings in the original codebase. Specifically, we use the default LLM agent framework and set the temperature to 1.0, top-p to 0.95, max sequence length to 256K. The user model is set to GPT-4.1 with 0.0 temperature to ensure a stable interaction during evaluation. For the airline domain, since it has incorrect ground truth answers, we use the fixed version from Claude Opus 4.5 to ensure evaluation reliability~\footnote{https://github.com/sierra-research/tau2-bench/pulls/chrisgorgo}. For the retail and telecom domains, we also follow Claude Opus 4.5 to include a general prompt addendum to the user prompt to avoid failure modes from the user ending the interaction incorrectly~\footnote{https://github.com/anthropics/model-cards/tree/main/claude-opus-4-5-20251101/tau2}. We report an average score of 8 runs to ensure stable evaluation results.

\subsubsection{General benchmarks}

\paragraph{Arena-Hard-v2.0.} We use the official evaluation metric of Arena-Hard-v2.0~\cite{arenahard2024} and use GPT-4.1~\cite{openai2024gpt4technicalreport} as the judge model.

\paragraph{MultiChallenge.} We use the official evaluation metric of MultiChallenge~\cite{sirdeshmukh2025multichallengerealisticmultiturnconversation} with o3-mini~\cite{o3-mini} as the judge model. This follows findings from the GPT-5\cite{gpt5} release that GPT-4o~\cite{openai2024gpt4technicalreport} frequently mis-scores complex responses, leading to underestimated results.

\paragraph{IFBench.} We use the official evaluation method of IFBench~\cite{pyatkin2025generalizing}.

\subsubsection{Long Context benchmarks}

\paragraph{LongBench v2.} We use the official evaluation method of LongBench v2~\cite{bai2024longbenchv2}.

\paragraph{MRCR-8needle.} For MRCR-8needle~\cite{vodrahalli2024michelangelo} benchmark, we report the Area Under Curve (AUC) metric, following the protocol established by ContextArena~\footnote{https://contextarena.ai/}. Specifically, we use the \textbf{AUC@128k} metric, which provides a single holistic score summarizing performance across context lengths up to 131,072 tokens.

The AUC is calculated by plotting the average retrieval accuracy for each context bin (ranging from 8k to 128k) against the bin's maximum context length. We apply the trapezoidal rule on a linear scale to measure the area under the resulting curve, which is then normalized by the total context width (128k minus the initial bin size) to yield a percentage score between 0\% and 100\%. This metric effectively penalizes performance degradation as difficulty increases with longer context sequences.

\paragraph{FRAMES-Oracle.} We use the official evaluation metric of FRAMES\cite{krishna2025fact}. Since our focus is on long-context capabilities, we specifically report results for the \textbf{Oracle Prompt} subset. In this setting, the model is provided with the question alongside all ground-truth Wikipedia articles used during human annotation. This configuration serves as an upper bound for model performance, simulating a perfect retrieval system that delivers all relevant context to the model.

\paragraph{RepoQA.} We use the official evaluation method of REPOQA~\cite{liu2024repoqaevaluatinglongcontext}.

\input{src/internal-benchmark}

%% file: src/tables/app_pacore_prompt_template_table.tex
\begin{table}[h]
    \centering
    \scriptsize
    \resizebox{0.99\linewidth}{!}{
        \begin{tabular}{l}
            \toprule
            \rowcolor{header_gray}
            \textbf{Panel A: Standard User Query} (Last role: \texttt{user}) \\
            \midrule
            You are given a problem and a list of reference responses. Your job is to analyze these references and provide your own response. \\
            \\
            Original Problem: \\
            \textcolor{blue}{ \{\{ original\_content \}\} } \\
            \\
            Reference Responses: \\
            \textit{Note: Some references may contain <tool\_call> tags indicating tool calls the reference intended to make.} \\
            \textit{These tool calls have NOT been executed - they are shown only as reference for your analysis.} \\
            \textcolor{blue}{ \{\% for response in ref\_responses \%\} } \\
            Reference \textcolor{blue}{\{\{ loop.index \}\}}: \\
            \textcolor{blue}{ \{\{ response \}\} } \\
            \textcolor{blue}{ \{\% endfor \%\} } \\
            \\
            Now, based on the original problem and reference responses above, please provide your own comprehensive solution. \\
            \midrule
            \textbf{Panel B: Tool Observation} (Last role: \texttt{tool}) \\
            \midrule
            You are given a tool response and a list of reference responses analyzing it. Your job is to analyze these references and provide your own response. \\
            \\
            Original Tool Response: \\
            \textcolor{blue}{ \{\{ original\_content \}\} } \\
            \\
            Reference Responses: \\
            \textit{[Same preamble regarding unexecuted tool calls as in Panel A]} \\
            \textcolor{blue}{ \{\% for response in ref\_responses \%\} } \\
            Reference \textcolor{blue}{\{\{ loop.index \}\}}: \\
            \textcolor{blue}{ \{\{ response \}\} } \\
            \textcolor{blue}{ \{\% endfor \%\} } \\
            \\
            Now, based on the original tool response and reference responses above, please provide your own comprehensive analysis and next steps. \\
            \bottomrule
        \end{tabular}
    }
    \caption{\textbf{Input serialization templates for Tool-integrated PaCoRe.} We introduce distinct templates to handle the initial user query (Panel A) and subsequent tool observations (Panel B). Note that \texttt{tool\_calls} within reference branches are serialized as text for analysis.}
    \label{tab:pacore-tool-template}
\end{table}

%% file: src/internal-benchmark.tex
\subsection{Internal Evaluation - Benchmarks and Methodology}
\subsubsection{Data Analysis Benchmark}
To reliably assess \ourmodel’s ability to perform practical data-analysis tasks in the Claude Code environment, we develop an internal Data Analysis Benchmark for evaluating end-to-end analytical problem solving under realistic business constraints. The benchmark is constructed by systematically distilling senior practitioners’ tacit expertise into a rubric-grounded evaluation suite. This approach captures the ambiguity and contextual nuance of real-world analytics while ensuring consistent evaluation through standardized rubrics and verifiable ground-truth artifacts.

The benchmark is constructed using an expert-driven, rubric-based protocol to ensure domain authenticity and scoring reliability. Ten senior data analytics leaders from major Chinese internet companies, each with over 15 years of experience, contributed real-world business cases through structured interviews that elicited core analytical patterns and decision logic. This process yields representative tasks paired with expert-endorsed solution strategies.

Interview materials are normalized into machine-consumable tasks, each comprising a problem statement, a CSV dataset, a reference analysis, and a weighted checklist-style scoring rubric. The resulting benchmark contains 50 items spanning diverse analytical intents, with an average of 26.9 rubric items per task. Quality is ensured through iterative expert review, aligning task definitions, data, reference solutions, and evaluation criteria to improve validity and reproducibility.

We further implement a unified end-to-end evaluation framework covering task execution, automated scoring, and report synthesis. The framework supports code-based, research-oriented, and text-based analyses within a single pipeline, enabling scalable and reproducible evaluation across heterogeneous environments with low integration overhead.

\paragraph{Evaluation Method.}
Each task is evaluated by a model-based evaluator that scores generated outputs against expert-defined rubrics, with results averaged over 3 identical runs to reduce stochastic variance and ensure reliable, comparable cross-model evaluation.

\begin{table}[ht]
\centering
\captionsetup{position=bottom}
\begin{tabular}{l r}
\toprule
\rowcolor{header_gray}
\textbf{Model} & \textbf{Avg@3(\%)} \\
\midrule
Claude Opus 4.5   & 45.0 \\
\ourmodel    & 39.6 \\
GPT-5.2           & 39.3 \\
Gemini 3.0 Pro      & 33.6 \\
Deepseek V3.2     & 27.9 \\
\bottomrule
\end{tabular}
\caption{Evaluation Results on the Data Analysis Benchmark}
\label{tab:data_analysis_bmk_results}
\end{table}

\paragraph{Evaluation Results.}
Table~\ref{tab:data_analysis_bmk_results} presents the results on the Data Analysis Benchmark. Claude Opus 4.5 ranks first overall, while \ourmodel achieves a strong second place (39.58\%) and remains very close to GPT-5.2 (39.31\%). Its competitive performance may be partly related to relatively good adaptation to the Claude Code environment. In addition, \ourmodel demonstrates a favorable speed–capability trade-off, maintaining solid analytical quality while delivering faster responses. The results position \ourmodel as a highly efficient and competitive option for real-world data analysis tasks.

\subsubsection{Consulting and Recommendations Benchmark}
To rigorously evaluate \ourmodel in real-world advisory scenarios, we curate a benchmark of 500 diverse queries sourced from authentic social platforms such as Reddit, Stack Exchange, and various community forums. These queries represent authentic user intent across everyday life, academic learning, entertainment, and professional workplace contexts.

Here, we implement an "Anchor-Based" scoring framework to evaluate candidate models. In this process, we first utilize leading models, including GPT-5.2, Claude Opus 4.5, and DeepSeek V3.2 to generate independent responses for each query. These high-level outputs are then synthesized and refined by human experts to create a Reference Response as Ground Truth. This reference serves as a high-quality "Anchor" with a standardized performance value of 88/100.

We then measure the performance of the models across four critical dimensions, applying a rigorous scoring rubric, including Usefulness, Logic, Instruction Following, and Tone. Usefulness assesses whether the model delivers a ready-to-use solution that meaningfully resolves the task with expert-level depth, actionable steps, and feasible recommendations. Logic evaluates factual accuracy and structural soundness, checking for hallucinations, incorrect citations, invalid conclusions, or causal and temporal inconsistencies, as well as overall coherence and argument flow. Instruction Following measures adherence to both explicit constraints (\textit{e.g.}, formatting, length, and stated requirements) and implicit contextual expectations embedded in the user query. Tone assesses communicative quality, including appropriateness of language and register, clarity in unpacking complex reasoning, and calibrated expression that avoids overconfidence while clearly signaling uncertainty when appropriate.

We employ a Hybrid LLM-as-a-Judge system. Recognizing that different frontier models have distinct evaluative strengths, we assign specific scoring responsibilities as follows: Logic, Instruction Following, and Usefulness: These three dimensions are evaluated by GPT-5.2, leveraging its industry-leading capabilities in factual verification, constraint checking, and objective problem-solving. Tone: This dimension is evaluated by Claude Opus 4.5, utilizing its superior nuance in linguistic style, emotional calibration, and "human-like" resonance.
Judge reliability is validated through an alignment study with human experts, yielding a high Pearson correlation between AI- and human-assigned scores. Final scores are computed using equal weighting across the four dimensions (25\% each), ensuring a balanced assessment that jointly reflects technical correctness and communicative quality.

\begin{table}[htbp]
  \centering
  \captionsetup{position=bottom}
  \begin{tabular}{lccccc}
    \toprule
    \rowcolor{header_gray}
    \textbf{Model} & \textbf{Average} & \textbf{Usefulness} & \textbf{Logic} & \textbf{Tone} & \textbf{Instruction-following} \\
    \midrule
    GPT-5.2 & 77.8\% & 77.2\% & 81.9\% & 73.0\% & 79.6\% \\
    Kimi K2.5 & 72.2\% & 77.1\% & 62.1\% & 72.7\% & 77.3\% \\
    Gemini 3.0 Pro & 70.6\% & 73.9\% & 61.7\% & 72.3\% & 74.4\% \\
    \rowcolor{gray!20}
    \ourmodel & 70.5\% & 73.3\% & 62.1\% & 72.4\% & 74.2\% \\
    Deepseek V3.2 & 70.3\% & 72.5\% & 64.4\% & 71.2\% & 72.9\% \\
    GLM-4.7 & 70.3\% & 73.5\% & 61.5\% & 72.5\% & 73.6\% \\
    Claude Opus 4.5 & 68.5\% & 69.7\% & 66.5\% & 65.9\% & 72.1\% \\
    Mimo-V2 Flash & 67.9\% & 71.5\% & 58.0\% & 70.6\% & 71.4\% \\
    Minimax M2.1 & 67.1\% & 70.7\% & 60.1\% & 67.2\% & 70.4\% \\
    \bottomrule
  \end{tabular}
  \caption{Evaluation results on the Consulting and Recommendations Benchmark}
  \label{tab:consulting_rec_bmk}
\end{table}

\paragraph{Evaluation Results.}
Table~\ref{tab:consulting_rec_bmk} shows that \ourmodel achieves an average Score of 70.5\% on the Consulting and Recommendations Benchmark, securing the 4th position overall. \ourmodel matches Gemini 3.0 Pro performance across all dimensions, achieving comparable Pro-level scores (70.5\% vs. 70.6\%) while offering substantially lower inference cost and latency. Unlike many fast models that trade speed for degraded reasoning quality, \ourmodel surpasses larger models in the Logic dimension, reducing hallucinations and logical failures and making it well suited for automated consulting workflows where factual integrity is critical.

\subsubsection{\ourmodel + Step-GUI}

To validate \ourmodel's efficacy in real-world agentic scenarios, we evaluate on \textbf{AndroidDaily Hard}~\cite{yan2025step}, a challenging benchmark designed for Chinese mobile application environments. This benchmark comprises compositional tasks spanning e-commerce transactions, multimedia interactions, and daily mobile operations, offering a naturalistic testbed for assessing GUI agent capabilities in complex, multi-step workflows representative of production deployments.

We empirically investigate two architectural instantiations: (1) \textbf{Step-GUI}~\cite{yan2025step}, a lightweight on-device agent (Edge Only) that executes tasks autonomously using local computational resources, and (2) \textbf{\ourmodel + Step-GUI}, an edge-cloud collaborative framework wherein \ourmodel functions as a cloud-based reasoning orchestrator that synthesizes high-level task plans, decomposes them into executable primitives via the GUI-MCP protocol, and delegates low-level control to the on-device Step-GUI agent. This hierarchical architecture exploits the complementary strengths of cloud-scale reasoning and edge efficiency: \ourmodel's 11B active parameters enable sophisticated multi-step planning and contextual understanding, while Step-GUI ensures low-latency action execution and privacy-preserving local control.

\paragraph{Quantitative Results.} The edge-cloud collaborative paradigm achieves a success rate of \textbf{57.0\%} on AndroidDaily Hard, substantially outperforming the edge-only baseline (\textbf{40.0\%}). This result suggests that combining strong cloud-side reasoning with efficient edge execution is an effective strategy for navigating deployment constraints in multi-round agent interactions.

\paragraph{Architectural Generalization.} Critically, this collaborative pattern extends beyond mobile ecosystems to heterogeneous platforms including desktop computers and automotive infotainment systems. By decoupling cognitive orchestration (cloud) from embodied execution (edge), the framework establishes a scalable paradigm for deploying sophisticated agents in resource-constrained industrial environments—directly aligned with \ourmodel's design objective of redefining the efficiency frontier for production-grade agentic systems. The results underscore that effective real-world agents require not only advanced reasoning capabilities but also architectures that harmonize computational distribution across infrastructure tiers.

%% file: references.bib
@misc{liu2025moeparallelfoldingheterogeneous,
      title={MoE Parallel Folding: Heterogeneous Parallelism Mappings for Efficient Large-Scale MoE Model Training with Megatron Core}, 
      author={Dennis Liu and Zijie Yan and Xin Yao and Tong Liu and Vijay Korthikanti and Evan Wu and Shiqing Fan and Gao Deng and Hongxiao Bai and Jianbin Chang and Ashwath Aithal and Michael Andersch and Mohammad Shoeybi and Jiajie Yao and Chandler Zhou and David Wu and Xipeng Li and June Yang},
      year={2025},
      eprint={2504.14960},
      archivePrefix={arXiv},
      primaryClass={cs.LG},
      url={https://arxiv.org/abs/2504.14960}, 
}

@misc{guo2025sonicmoeacceleratingmoeio,
      title={SonicMoE: Accelerating MoE with IO and Tile-aware Optimizations}, 
      author={Wentao Guo and Mayank Mishra and Xinle Cheng and Ion Stoica and Tri Dao},
      year={2025},
      eprint={2512.14080},
      archivePrefix={arXiv},
      primaryClass={cs.LG},
      url={https://arxiv.org/abs/2512.14080}, 
}

@misc{narayanan2021efficientlargescalelanguagemodel,
      title={Efficient Large-Scale Language Model Training on GPU Clusters Using Megatron-LM}, 
      author={Deepak Narayanan and Mohammad Shoeybi and Jared Casper and Patrick LeGresley and Mostofa Patwary and Vijay Anand Korthikanti and Dmitri Vainbrand and Prethvi Kashinkunti and Julie Bernauer and Bryan Catanzaro and Amar Phanishayee and Matei Zaharia},
      year={2021},
      eprint={2104.04473},
      archivePrefix={arXiv},
      primaryClass={cs.CL},
      url={https://arxiv.org/abs/2104.04473}, 
}

@misc{lepikhin2020gshardscalinggiantmodels,
      title={GShard: Scaling Giant Models with Conditional Computation and Automatic Sharding}, 
      author={Dmitry Lepikhin and HyoukJoong Lee and Yuanzhong Xu and Dehao Chen and Orhan Firat and Yanping Huang and Maxim Krikun and Noam Shazeer and Zhifeng Chen},
      year={2020},
      eprint={2006.16668},
      archivePrefix={arXiv},
      primaryClass={cs.CL},
      url={https://arxiv.org/abs/2006.16668}, 
}

@misc{suzgun2022challengingbbh,
  title={Challenging BIG-Bench Tasks and Whether Chain-of-Thought Can Solve Them},
  author={Mirac Suzgun and Nathan Scales and Nathanael Sch{\"a}rli and Sebastian Gehrmann and Yi Tay and Hyung Won Chung and Aakanksha Chowdhery and Quoc V. Le and Ed H. Chi and Denny Zhou and Jason Wei},
  year={2022},
  eprint={2210.09261},
  archivePrefix={arXiv},
  primaryClass={cs.CL},
  url={https://arxiv.org/abs/2210.09261},
}

@misc{hendrycks2020mmlu,
  title={Measuring Massive Multitask Language Understanding},
  author={Dan Hendrycks and Collin Burns and Steven Basart and Andy Zou and Mantas Mazeika and Dawn Song and Jacob Steinhardt},
  year={2020},
  eprint={2009.03300},
  archivePrefix={arXiv},
  primaryClass={cs.CY},
  url={https://arxiv.org/abs/2009.03300},
}

@misc{gema2024donewithmmlu,
  title={Are We Done with MMLU?},
  author={Aryo Pradipta Gema and Joshua Ong Jun Leang and Giwon Hong and Alessio Devoto and Alberto Carlo Maria Mancino and Rohit Saxena and Xuanli He and Yu Zhao and Xiaotang Du and Mohammad Reza Ghasemi Madani and Claire Barale and Robert McHardy and Joshua Harris and Jean Kaddour and Emile van Krieken and Pasquale Minervini},
  year={2024},
  eprint={2406.04127},
  archivePrefix={arXiv},
  primaryClass={cs.CL},
  url={https://arxiv.org/abs/2406.04127},
}

@misc{wang2024mmlupro,
  title={MMLU-Pro: A More Robust and Challenging Multi-Task Language Understanding Benchmark},
  author={Yubo Wang and Xueguang Ma and Ge Zhang and Yuansheng Ni and Abhranil Chandra and Shiguang Guo and Weiming Ren and Aaran Arulraj and Xuan He and Ziyan Jiang and Tianle Li and Max Ku and Kai Wang and Alex Zhuang and Rongqi Fan and Xiang Yue and Wenhu Chen},
  year={2024},
  eprint={2406.01574},
  archivePrefix={arXiv},
  primaryClass={cs.CL},
  url={https://arxiv.org/abs/2406.01574},
}

@misc{zellers2019hellaswag,
  title={HellaSwag: Can a Machine Really Finish Your Sentence?},
  author={Rowan Zellers and Ari Holtzman and Yonatan Bisk and Ali Farhadi and Yejin Choi},
  year={2019},
  eprint={1905.07830},
  archivePrefix={arXiv},
  primaryClass={cs.CL},
  url={https://arxiv.org/abs/1905.07830},
}

@misc{sakaguchi2019winogrande,
  title={WinoGrande: An Adversarial Winograd Schema Challenge at Scale},
  author={Keisuke Sakaguchi and Ronan Le Bras and Chandra Bhagavatula and Yejin Choi},
  year={2019},
  eprint={1907.10641},
  archivePrefix={arXiv},
  primaryClass={cs.CL},
  url={https://arxiv.org/abs/1907.10641},
}

@misc{du2025supergpqa,
  title={SuperGPQA: Scaling LLM Evaluation across 285 Graduate Disciplines},
  author={Xinrun Du and Yifan Yao and Kaijing Ma and Bingli Wang and Tianyu Zheng and King Zhu and Minghao Liu and Yiming Liang and Xiaolong Jin and Zhenlin Wei and others},
  year={2025},
  eprint={2502.14739},
  archivePrefix={arXiv},
  primaryClass={cs.CL},
  url={https://arxiv.org/abs/2502.14739},
}

@misc{openai2024simpleqa,
  title={SimpleQA},
  author={OpenAI},
  year={2024},
  howpublished={\url{https://github.com/openai/simple-evals}},
}

@misc{cobbe2021gsm8k,
  title={Training Verifiers to Solve Math Word Problems},
  author={Karl Cobbe and Vineet Kosaraju and Mohammad Bavarian and Mark Chen and Heewoo Jun and Lukasz Kaiser and Matthias Plappert and Jerry Tworek and Jacob Hilton and Reiichiro Nakano and Christopher Hesse and John Schulman},
  year={2021},
  eprint={2110.14168},
  archivePrefix={arXiv},
  primaryClass={cs.LG},
  url={https://arxiv.org/abs/2110.14168},
}

@misc{hendrycks2021math,
  title={Measuring Mathematical Problem Solving With the MATH Dataset},
  author={Dan Hendrycks and Collin Burns and Saurav Kadavath and Akul Arora and Steven Basart and Eric Tang and Dawn Song and Jacob Steinhardt},
  year={2021},
  eprint={2103.03874},
  archivePrefix={arXiv},
  primaryClass={cs.LG},
  url={https://arxiv.org/abs/2103.03874},
}

@misc{liu2023evalplus,
  title={Is Your Code Generated by ChatGPT Really Correct? Rigorous Evaluation of Large Language Models for Code Generation},
  author={Jiawei Liu and Chunqiu Steven Xia and Yuyao Wang and Lingming Zhang},
  year={2023},
  eprint={2305.01210},
  archivePrefix={arXiv},
  primaryClass={cs.SE},
  url={https://arxiv.org/abs/2305.01210},
}

@misc{cassano2022multipl,
  title={MultiPL-E: A Scalable and Extensible Approach to Benchmarking Neural Code Generation},
  author={Federico Cassano and John Gouwar and Daniel Nguyen and Sydney Nguyen and Luna Phipps-Costin and Donald Pinckney and Ming-Ho Yee and Yangtian Zi and Carolyn Jane Anderson and Molly Q Feldman and Arjun Guha and Michael Greenberg and Abhinav Jangda},
  year={2022},
  eprint={2208.08227},
  archivePrefix={arXiv},
  primaryClass={cs.LG},
  url={https://arxiv.org/abs/2208.08227},
}

@misc{huang2023ceval,
  title={C-Eval: A Multi-Level Multi-Discipline Chinese Evaluation Suite for Foundation Models},
  author={Yuzhen Huang and Yuzhuo Bai and Zhihao Zhu and Junlei Zhang and Jinghan Zhang and Tangjun Su and Junteng Liu and Chuancheng Lv and Yikai Zhang and Jiayi Lei and Yao Fu and Maosong Sun and Junxian He},
  year={2023},
  eprint={2305.08322},
  archivePrefix={arXiv},
  primaryClass={cs.CL},
  url={https://arxiv.org/abs/2305.08322},
}

@misc{li2023cmmlu,
  title={CMMLU: Measuring massive multitask language understanding in Chinese},
  author={Haonan Li and Yixuan Zhang and Fajri Koto and Yifei Yang and Hai Zhao and Yeyun Gong and Nan Duan and Timothy Baldwin},
  year={2023},
  eprint={2306.09212},
  archivePrefix={arXiv},
  primaryClass={cs.CL},
  url={https://arxiv.org/abs/2306.09212},
}

@misc{he2024chinesesimpleqa,
  title={Chinese SimpleQA: A Chinese Factuality Evaluation for Large Language Models},
  author={Yancheng He and Shilong Li and Jiaheng Liu and Yingshui Tan and Weixun Wang and Hui Huang and Xingyuan Bu and Hangyu Guo and Chengwei Hu and Boren Zheng and others},
  year={2024},
  eprint={2411.07140},
  archivePrefix={arXiv},
  primaryClass={cs.CL},
  url={https://arxiv.org/abs/2411.07140},
}

@misc{rein2023gpqa,
  title={GPQA: A Graduate-Level Google-Proof Q\&A Benchmark},
  author={David Rein and Betty Li Hou and Asa Cooper Stickland and Jackson Petty and Richard Yuanzhe Pang and Julien Dirani and Julian Michael and Samuel R. Bowman},
  year={2023},
  eprint={2311.12022},
  archivePrefix={arXiv},
  primaryClass={cs.AI},
  url={https://arxiv.org/abs/2311.12022},
}

@article{terminalbench2025,
  title={Terminal-Bench: Benchmarking Agents on Hard, Realistic Tasks in Command Line Interfaces},
  author={Merrill, Mike A and Shaw, Alexander G and Carlini, Nicholas and Li, Boxuan and Raj, Harsh and Bercovich, Ivan and Shi, Lin and Shin, Jeong Yeon and Walshe, Thomas and Buchanan, E Kelly and others},
  journal={arXiv preprint arXiv:2601.11868},
  year={2026}
}

@misc{wei2025browsecomp,
  title={BrowseComp: A Simple Yet Challenging Benchmark for Browsing Agents},
  author={Jason Wei and Zhiqing Sun and Spencer Papay and Scott McKinney and Jeffrey Han and Isa Fulford and Hyung Won Chung and Alex Tachard Passos and William Fedus and Amelia Glaese},
  year={2025},
  eprint={2504.12516},
  archivePrefix={arXiv},
  primaryClass={cs.CL},
  url={https://arxiv.org/abs/2504.12516},
}

@misc{tau2bench2025,
  title={tau2-bench},
  author={Sierra Research},
  year={2025},
  howpublished={\url{https://github.com/sierra-research/tau2-bench}},
}

@misc{phan2025hle,
  title={Humanity's Last Exam},
  author={Long Phan and Tony CY Pang and Adam Wecker and Yifan Xiong and Dan Hendrycks and others},
  year={2025},
  eprint={2501.14249},
  archivePrefix={arXiv},
  primaryClass={cs.AI},
  url={https://arxiv.org/abs/2501.14249},
}

@misc{jordan2024muon,
  author={Keller Jordan and Yuchen Jin and Vlado Boza and Jiacheng You and Franz Cesista and Laker Newhouse and Jeremy Bernstein},
  title={Muon: An optimizer for hidden layers in neural networks},
  year={2024},
  url={https://kellerjordan.github.io/posts/muon/}
}

@inproceedings{rajbhandari2020zero,
  author={Rajbhandari, Samyam and Rasley, Jeff and Ruwase, Olatunji and He, Yuxiong},
  booktitle={SC20: International Conference for High Performance Computing, Networking, Storage and Analysis}, 
  title={ZeRO: Memory optimizations Toward Training Trillion Parameter Models}, 
  year={2020},
  volume={},
  number={},
  pages={1-16},
  doi={10.1109/SC41405.2020.00024}}

@inproceedings{Ansel_PyTorch_2_Faster_2024,
author = {Ansel, Jason and Yang, Edward and He, Horace and Gimelshein, Natalia and Jain, Animesh and Voznesensky, Michael and Bao, Bin and Bell, Peter and Berard, David and Burovski, Evgeni and Chauhan, Geeta and Chourdia, Anjali and Constable, Will and Desmaison, Alban and DeVito, Zachary and Ellison, Elias and Feng, Will and Gong, Jiong and Gschwind, Michael and Hirsh, Brian and Huang, Sherlock and Kalambarkar, Kshiteej and Kirsch, Laurent and Lazos, Michael and Lezcano, Mario and Liang, Yanbo and Liang, Jason and Lu, Yinghai and Luk, CK and Maher, Bert and Pan, Yunjie and Puhrsch, Christian and Reso, Matthias and Saroufim, Mark and Siraichi, Marcos Yukio and Suk, Helen and Suo, Michael and Tillet, Phil and Wang, Eikan and Wang, Xiaodong and Wen, William and Zhang, Shunting and Zhao, Xu and Zhou, Keren and Zou, Richard and Mathews, Ajit and Chanan, Gregory and Wu, Peng and Chintala, Soumith},
booktitle = {29th ACM International Conference on Architectural Support for Programming Languages and Operating Systems, Volume 2 (ASPLOS '24)},
doi = {10.1145/3620665.3640366},
month = apr,
publisher = {ACM},
title = {{PyTorch 2: Faster Machine Learning Through Dynamic Python Bytecode Transformation and Graph Compilation}},
url = {https://docs.pytorch.org/assets/pytorch2-2.pdf},
year = {2024}
}

@article{megatron-lm,
  title={Megatron-LM: Training Multi-Billion Parameter Language Models Using Model Parallelism},
  author={Shoeybi, Mohammad and Patwary, Mostofa and Puri, Raul and LeGresley, Patrick and Casper, Jared and Catanzaro, Bryan},
  journal={arXiv preprint arXiv:1909.08053},
  year={2019}
}

@article{sharma2025researchrubrics,
  title = {ResearchRubrics: A Benchmark of Prompts and Rubrics For Evaluating Deep Research Agents},
  author = {Sharma, Manasi and Zhang, Chen Bo Calvin and others},
  journal = {arXiv preprint arXiv:2511.07685},
  year = {2025}
}

@misc{amsel2025polarexpress,
  title         = {The Polar Express: Optimal Matrix Sign Methods and Their Application to the Muon Algorithm},
  author        = {Amsel, Noah and Persson, David and Musco, Christopher and Gower, Robert M.},
  year          = {2025},
  eprint        = {2505.16932},
  archivePrefix = {arXiv},
  primaryClass  = {cs.LG},
  doi           = {10.48550/arXiv.2505.16932},
  url           = {https://arxiv.org/abs/2505.16932}
}

@misc{bernstein2024oldoptimizernewnorm,
  title         = {Old Optimizer, New Norm: An Anthology},
  author        = {Bernstein, Jeremy and Newhouse, Laker},
  year          = {2024},
  eprint        = {2409.20325},
  archivePrefix = {arXiv},
  primaryClass  = {cs.LG},
  doi           = {10.48550/arXiv.2409.20325},
  url           = {https://arxiv.org/abs/2409.20325}
}

@misc{xiao2026mimov2flashtechnicalreport,
      title={MiMo-V2-Flash Technical Report}, 
      author={LLM-Core Xiaomi},
      year={2026},
      eprint={2601.02780},
      archivePrefix={arXiv},
      primaryClass={cs.CL},
      url={https://arxiv.org/abs/2601.02780}, 
}

@article{fedus2021switchtransformers,
  title        = {Switch Transformers: Scaling to Trillion Parameter Models with Simple and Efficient Sparsity},
  author       = {Fedus, William and Zoph, Barret and Shazeer, Noam},
  journal      = {arXiv preprint arXiv:2101.03961},
  year         = {2021},
  doi          = {10.48550/arXiv.2101.03961},
  url          = {https://arxiv.org/abs/2101.03961}
}

@article{qiu2025demonsdetail,
  title        = {Demons in the Detail: On Implementing Load Balancing Loss for Training Specialized Mixture-of-Expert Models},
  author       = {Qiu, Zihan and Huang, Zeyu and Zheng, Bo and Wen, Kaiyue and Wang, Zekun and Men, Rui and Titov, Ivan and Liu, Dayiheng and Zhou, Jingren and Lin, Junyang},
  journal      = {arXiv preprint arXiv:2501.11873},
  year         = {2025},
  doi          = {10.48550/arXiv.2501.11873},
  url          = {https://arxiv.org/abs/2501.11873}
}

@article{yang2025qwen3,
  title        = {Qwen3 Technical Report},
  author       = {Yang, An and Li, Anfeng and Yang, Baosong and Zhang, Beichen and Hui, Binyuan and Zheng, Bo and Yu, Bowen and Gao, Chang and Huang, Chengen and Lv, Chenxu and Zheng, Chujie and Liu, Dayiheng and Zhou, Fan and Huang, Fei and Hu, Feng and Ge, Hao and Wei, Haoran and Lin, Huan and Tang, Jialong and Yang, Jian and Tu, Jianhong and Zhang, Jianwei and Yang, Jianxin and Yang, Jiaxi and Zhou, Jing and Zhou, Jingren and Lin, Junyang and Dang, Kai and Bao, Keqin and Yang, Kexin and Yu, Le and Deng, Lianghao and Li, Mei and Xue, Mingfeng and Li, Mingze and Zhang, Pei and Wang, Peng and Zhu, Qin and Men, Rui and Gao, Ruize and Liu, Shixuan and Luo, Shuang and Li, Tianhao and Tang, Tianyi and Yin, Wenbiao and Ren, Xingzhang and Wang, Xinyu and Zhang, Xinyu and Ren, Xuancheng and Fan, Yang and Su, Yang and Zhang, Yichang and Zhang, Yinger and Wan, Yu and Liu, Yuqiong and Wang, Zekun and Cui, Zeyu and Zhang, Zhenru and Zhou, Zhipeng and Qiu, Zihan},
  journal      = {arXiv preprint arXiv:2505.09388},
  year         = {2025},
  doi          = {10.48550/arXiv.2505.09388},
  url          = {https://arxiv.org/abs/2505.09388}
}

@article{wang2024lossfreebalancing,
  title        = {Auxiliary-Loss-Free Load Balancing Strategy for Mixture-of-Experts},
  author       = {Wang, Lean and Gao, Huazuo and Zhao, Chenggang and Sun, Xu and Dai, Damai},
  journal      = {arXiv preprint arXiv:2408.15664},
  year         = {2024},
  doi          = {10.48550/arXiv.2408.15664},
  url          = {https://arxiv.org/abs/2408.15664}
}

@article{deepseek2024deepseekv3,
  title        = {DeepSeek-V3 Technical Report},
  author       = {{DeepSeek-AI}},
  journal      = {arXiv preprint arXiv:2412.19437},
  year         = {2024},
  doi          = {10.48550/arXiv.2412.19437},
  url          = {https://arxiv.org/abs/2412.19437}
}

@misc{dai2024deepseekmoeultimateexpertspecialization,
      title={DeepSeekMoE: Towards Ultimate Expert Specialization in Mixture-of-Experts Language Models}, 
      author={Damai Dai and Chengqi Deng and Chenggang Zhao and R. X. Xu and Huazuo Gao and Deli Chen and Jiashi Li and Wangding Zeng and Xingkai Yu and Y. Wu and Zhenda Xie and Y. K. Li and Panpan Huang and Fuli Luo and Chong Ruan and Zhifang Sui and Wenfeng Liang},
      year={2024},
      eprint={2401.06066},
      archivePrefix={arXiv},
      primaryClass={cs.CL},
      url={https://arxiv.org/abs/2401.06066}, 
}

@misc{openai2025gptoss120bgptoss20bmodel,
  title = {GPT-OSS-120B \& GPT-OSS-20B Model Card},
  author = {OpenAI},
  year = {2025},
  eprint = {2508.10925},
  archivePrefix = {arXiv},
  primaryClass = {cs.CL},
  url = {https://arxiv.org/abs/2508.10925}
}

@misc{qiu2025gatedattentionlargelanguage,
      title={Gated Attention for Large Language Models: Non-linearity, Sparsity, and Attention-Sink-Free}, 
      author={Zihan Qiu and Zekun Wang and Bo Zheng and Zeyu Huang and Kaiyue Wen and Songlin Yang and Rui Men and Le Yu and Fei Huang and Suozhi Huang and Dayiheng Liu and Jingren Zhou and Junyang Lin},
      year={2025},
      eprint={2505.06708},
      archivePrefix={arXiv},
      primaryClass={cs.CL},
      url={https://arxiv.org/abs/2505.06708}, 
}

@misc{step3blog,
      title={Step3: Cost-Effective Multimodal Intelligence}, 
      author={StepFun Team},
      url={https://stepfun.ai/research/step3}, 
}

@misc{su2024roformer,
  title={Roformer: Enhanced transformer with rotary position embedding},
  author={Su, Jianlin and Ahmed, Murtadha and Lu, Yu and Pan, Shengfeng and Bo, Wen and Liu, Yunfeng},
  journal={Neurocomputing},
  volume={568},
  pages={127063},
  year={2024},
  publisher={Elsevier}
}

@misc{xiong2024effective,
  title={Effective long-context scaling of foundation models},
  author={Xiong, Wenhan and Liu, Jingyu and Molybog, Igor and Zhang, Hejia and Bhargava, Prajjwal and Hou, Rui and Martin, Louis and Rungta, Rashi and Sankararaman, Karthik Abinav and Oguz, Barlas and others},
  booktitle={Proceedings of the 2024 Conference of the North American Chapter of the Association for Computational Linguistics: Human Language Technologies (Volume 1: Long Papers)},
  pages={4643--4663},
  year={2024}
}

@misc{commoncrawl,
  title        = {Common Crawl},
  author       = {{Common Crawl}},
  howpublished = {\url{https://commoncrawl.org}}
}

@misc{su2025nemotroncctransformingcommoncrawl,
      title={Nemotron-CC: Transforming Common Crawl into a Refined Long-Horizon Pretraining Dataset},
      author={Dan Su and Kezhi Kong and Ying Lin and Joseph Jennings and Brandon Norick and Markus Kliegl and Mostofa Patwary and Mohammad Shoeybi and BryanCatanzaro},
      year={2025},
      eprint={2412.02595},
      archivePrefix={arXiv},
      primaryClass={cs.CL},
      url={https://arxiv.org/abs/2412.02595},
}

@misc{huang2025opencoderopencookbooktoptier,
      title={OpenCoder: The Open Cookbook for Top-Tier Code Large Language Models}, 
      author={Siming Huang and Tianhao Cheng and J. K. Liu and Jiaran Hao and Liuyihan Song and Yang Xu and J. Yang and Jiaheng Liu and Chenchen Zhang and Linzheng Chai and Ruifeng Yuan and Zhaoxiang Zhang and Jie Fu and Qian Liu and Ge Zhang and Zili Wang and Yuan Qi and Yinghui Xu and Wei Chu},
      year={2025},
      eprint={2411.04905},
      archivePrefix={arXiv},
      primaryClass={cs.CL},
      url={https://arxiv.org/abs/2411.04905}, 
}

@inproceedings{jimenez2024swebench,
    title={{SWE}-bench: Can Language Models Resolve Real-world Github Issues?},
    author={Carlos E Jimenez and John Yang and Alexander Wettig and Shunyu Yao and Kexin Pei and Ofir Press and Karthik R Narasimhan},
    booktitle={The Twelfth International Conference on Learning Representations},
    year={2024},
    url={https://openreview.net/forum?id=VTF8yNQM66}
}

@misc{yang2025swesmith,
    title={SWE-smith: Scaling Data for Software Engineering Agents},
    author={John Yang and Kilian Lieret and Carlos E. Jimenez and Alexander Wettig and Kabir Khandpur and Yanzhe Zhang and Binyuan Hui and Ofir Press and Ludwig Schmidt and Diyi Yang},
    year={2025},
    eprint={2504.21798},
    archivePrefix={arXiv},
    primaryClass={cs.SE},
    url={https://arxiv.org/abs/2504.21798},
}

@article{xia2024agentless,
  title={Agentless: Demystifying llm-based software engineering agents},
  author={Xia, Chunqiu Steven and Deng, Yinlin and Dunn, Soren and Zhang, Lingming},
  journal={arXiv preprint arXiv:2407.01489},
  year={2024}
}

@misc{zhou2025megamathpushinglimitsopen,
      title={MegaMath: Pushing the Limits of Open Math Corpora}, 
      author={Fan Zhou and Zengzhi Wang and Nikhil Ranjan and Zhoujun Cheng and Liping Tang and Guowei He and Zhengzhong Liu and Eric P. Xing},
      year={2025},
      eprint={2504.02807},
      archivePrefix={arXiv},
      primaryClass={cs.CL},
      url={https://arxiv.org/abs/2504.02807}, 
}

@misc{allal2025smollm2smolgoesbig,
      title={SmolLM2: When Smol Goes Big -- Data-Centric Training of a Small Language Model}, 
      author={Loubna Ben Allal and Anton Lozhkov and Elie Bakouch and Gabriel Martín Blázquez and Guilherme Penedo and Lewis Tunstall and Andrés Marafioti and Hynek Kydlíček and Agustín Piqueres Lajarín and Vaibhav Srivastav and Joshua Lochner and Caleb Fahlgren and Xuan-Son Nguyen and Clémentine Fourrier and Ben Burtenshaw and Hugo Larcher and Haojun Zhao and Cyril Zakka and Mathieu Morlon and Colin Raffel and Leandro von Werra and Thomas Wolf},
      year={2025},
      eprint={2502.02737},
      archivePrefix={arXiv},
      primaryClass={cs.CL},
      url={https://arxiv.org/abs/2502.02737}, 
}

@misc{shao2024deepseekmathpushinglimitsmathematical,
      title={DeepSeekMath: Pushing the Limits of Mathematical Reasoning in Open Language Models}, 
      author={Zhihong Shao and Peiyi Wang and Qihao Zhu and Runxin Xu and Junxiao Song and Xiao Bi and Haowei Zhang and Mingchuan Zhang and Y. K. Li and Y. Wu and Daya Guo},
      year={2024},
      eprint={2402.03300},
      archivePrefix={arXiv},
      primaryClass={cs.CL},
      url={https://arxiv.org/abs/2402.03300}, 
}

@misc{guo2026swefactoryautomatedfactoryissue,
      title={SWE-Factory: Your Automated Factory for Issue Resolution Training Data and Evaluation Benchmarks}, 
      author={Lianghong Guo and Yanlin Wang and Caihua Li and Wei Tao and Pengyu Yang and Jiachi Chen and Haoyu Song and Duyu Tang and Zibin Zheng},
      year={2026},
      eprint={2506.10954},
      archivePrefix={arXiv},
      primaryClass={cs.SE},
      url={https://arxiv.org/abs/2506.10954}, 
}

@misc{zhang2026docksmithscalingreliablecoding,
      title={DockSmith: Scaling Reliable Coding Environments via an Agentic Docker Builder}, 
      author={Jiaran Zhang and Luck Ma and Yanhao Li and Fanqi Wan and Di Qi and Xu Zhao and Jieyi Hou and Zhe Xie and Mengqiang Ren and Xin Wu and Zhewei Huang and Liangyu Chen and Yingwei Ma and Qi Han and Xiangyu Zhang},
      year={2026},
      eprint={2602.00592},
      archivePrefix={arXiv},
      primaryClass={cs.AI},
      url={https://arxiv.org/abs/2602.00592}, 
}

@misc{weborganizer,
      title={Organize the Web: Constructing Domains Enhances Pre-Training Data Curation}, 
      author={Alexander Wettig and Kyle Lo and Sewon Min and Hannaneh Hajishirzi and Danqi Chen and Luca Soldaini},
      year={2025},
      eprint={2502.10341},
      archivePrefix={arXiv},
      primaryClass={cs.CL},
      url={https://arxiv.org/abs/2502.10341}, 
}

@article{chen2021evaluating,
  title={Evaluating large language models trained on code},
  author={Chen, Mark and Tworek, Jerry and Jun, Heewoo and Yuan, Qiming and Pinto, Henrique Ponde De Oliveira and Kaplan, Jared and Edwards, Harri and Burda, Yuri and Joseph, Nicholas and Brockman, Greg and others},
  journal={arXiv preprint arXiv:2107.03374},
  year={2021}
}

@misc{cai2025fastmtpacceleratingllminference,
      title={FastMTP: Accelerating LLM Inference with Enhanced Multi-Token Prediction}, 
      author={Yuxuan Cai and Xiaozhuan Liang and Xinghua Wang and Jin Ma and Haijin Liang and Jinwen Luo and Xinyu Zuo and Lisheng Duan and Yuyang Yin and Xi Chen},
      year={2025},
      eprint={2509.18362},
      archivePrefix={arXiv},
      primaryClass={cs.LG},
      url={https://arxiv.org/abs/2509.18362}, 
}

@misc{gemmateam2024gemmaopenmodelsbased,
      title={Gemma: Open Models Based on Gemini Research and Technology}, 
      author={Team, Gemma and Mesnard, Thomas and Hardin, Cassidy and Dadashi, Robert and Bhupatiraju, Surya and Pathak, Shreya and Sifre, Laurent and Rivi{\`e}re, Morgane and Kale, Mihir Sanjay and Love, Juliette and others},
      year={2024},
      eprint={2403.08295},
      archivePrefix={arXiv},
      primaryClass={cs.CL},
      url={https://arxiv.org/abs/2403.08295}, 
}

@misc{deepseekai2024deepseekv32,
      title={DeepSeek-V3.2-Exp: Boosting Long-Context Efficiency with DeepSeek Sparse Attention}, 
      author={DeepSeek-AI},
      year={2025},
}

@misc{austin2021programsynthesislargelanguage,
      title={Program Synthesis with Large Language Models}, 
      author={Jacob Austin and Augustus Odena and Maxwell Nye and Maarten Bosma and Henryk Michalewski and David Dohan and Ellen Jiang and Carrie Cai and Michael Terry and Quoc Le and Charles Sutton},
      year={2021},
      eprint={2108.07732},
      archivePrefix={arXiv},
      primaryClass={cs.PL},
      url={https://arxiv.org/abs/2108.07732}, 
}

@misc{chen2021evaluatinglargelanguagemodels,
      title={Evaluating Large Language Models Trained on Code}, 
      author={Mark Chen and Jerry Tworek and Heewoo Jun and Qiming Yuan and Henrique Ponde de Oliveira Pinto and Jared Kaplan and Harri Edwards and Yuri Burda and Nicholas Joseph and Greg Brockman and Alex Ray and Raul Puri and Gretchen Krueger and Michael Petrov and Heidy Khlaaf and Girish Sastry and Pamela Mishkin and Brooke Chan and Scott Gray and Nick Ryder and Mikhail Pavlov and Alethea Power and Lukasz Kaiser and Mohammad Bavarian and Clemens Winter and Philippe Tillet and Felipe Petroski Such and Dave Cummings and Matthias Plappert and Fotios Chantzis and Elizabeth Barnes and Ariel Herbert-Voss and William Hebgen Guss and Alex Nichol and Alex Paino and Nikolas Tezak and Jie Tang and Igor Babuschkin and Suchir Balaji and Shantanu Jain and William Saunders and Christopher Hesse and Andrew N. Carr and Jan Leike and Josh Achiam and Vedant Misra and Evan Morikawa and Alec Radford and Matthew Knight and Miles Brundage and Mira Murati and Katie Mayer and Peter Welinder and Bob McGrew and Dario Amodei and Sam McCandlish and Ilya Sutskever and Wojciech Zaremba},
      year={2021},
      eprint={2107.03374},
      archivePrefix={arXiv},
      primaryClass={cs.LG},
      url={https://arxiv.org/abs/2107.03374}, 
}

@article{allenzhu2024physics,
  title={Physics of Language Models: Part 3.3, Knowledge Capacity Scaling Laws},
  author={Allen-Zhu, Zeyuan and Li, Yuanzhi},
  journal={arXiv preprint arXiv:2404.05405},
  year={2024}
}

@misc{fan2025urlsmetadatadiversityposition,
      title={Beyond URLs: Metadata Diversity and Position for Efficient LLM Pretraining}, 
      author={Dongyang Fan and Diba Hashemi and Sai Praneeth Karimireddy and Martin Jaggi},
      year={2025},
      eprint={2511.21613},
      archivePrefix={arXiv},
      primaryClass={cs.CL},
      url={https://arxiv.org/abs/2511.21613}, 
}

@inproceedings{gao2025metadata,
  title={Metadata Conditioning Accelerates Language Model Pre-training},
  author={Gao, Tianyu and Wettig, Alexander and He, Luxi and Dong, Yihe and Malladi, Sadhika and Chen, Danqi},
  booktitle={International Conference on Machine Learning (ICML)},
  year={2025}
}

@misc{liu2024repoqaevaluatinglongcontext,
      title={RepoQA: Evaluating Long Context Code Understanding}, 
      author={Jiawei Liu and Jia Le Tian and Vijay Daita and Yuxiang Wei and Yifeng Ding and Yuhan Katherine Wang and Jun Yang and Lingming Zhang},
      year={2024},
      eprint={2406.06025},
      archivePrefix={arXiv},
      primaryClass={cs.SE},
      url={https://arxiv.org/abs/2406.06025}, 
}

@misc{hsieh2024ruler,
  title         = {RULER: What's the Real Context Size of Your Long-Context Language Models?},
  author        = {Cheng-Ping Hsieh and Simeng Sun and Samuel Kriman and Shantanu Acharya and Dima Rekesh and Fei Jia and Yang Zhang and Boris Ginsburg},
  year          = {2024},
  eprint        = {2404.06654},
  archivePrefix = {arXiv},
  primaryClass  = {cs.CL},
  doi           = {10.48550/arXiv.2404.06654},
  url           = {https://arxiv.org/abs/2404.06654}
}

@misc{bai2024longbenchv2,
  title         = {LongBench v2: Towards Deeper Understanding and Reasoning on Realistic Long-context Multitasks},
  author        = {Yushi Bai and Shangqing Tu and Jiajie Zhang and Hao Peng and Xiaozhi Wang and Xin Lv and Shulin Cao and Jiazheng Xu and Lei Hou and Yuxiao Dong and Jie Tang and Juanzi Li},
  year          = {2024},
  eprint        = {2412.15204},
  archivePrefix = {arXiv},
  primaryClass  = {cs.CL},
  doi           = {10.48550/arXiv.2412.15204},
  url           = {https://arxiv.org/abs/2412.15204}
}

@misc{yen2024helmet,
  title         = {HELMET: How to Evaluate Long-Context Language Models Effectively and Thoroughly},
  author        = {Howard Yen and Tianyu Gao and Minmin Hou and Ke Ding and Daniel Fleischer and Peter Izsak and Moshe Wasserblat and Danqi Chen},
  year          = {2024},
  eprint        = {2410.02694},
  archivePrefix = {arXiv},
  primaryClass  = {cs.CL},
  doi           = {10.48550/arXiv.2410.02694},
  url           = {https://arxiv.org/abs/2410.02694}
}

@misc{zhou2025gsminfinite,
  title         = {GSM-Infinite: How Do Your LLMs Behave over Infinitely Increasing Context Length and Reasoning Complexity?},
  author        = {Yang Zhou and Hongyi Liu and Zhuoming Chen and Yuandong Tian and Beidi Chen},
  year          = {2025},
  eprint        = {2502.05252},
  archivePrefix = {arXiv},
  primaryClass  = {cs.CL},
  doi           = {10.48550/arXiv.2502.05252},
  url           = {https://arxiv.org/abs/2502.05252}
}

@article{austin2021program,
  title={Program synthesis with large language models},
  author={Austin, Jacob and Odena, Augustus and Nye, Maxwell and Bosma, Maarten and Michalewski, Henryk and Dohan, David and Jiang, Ellen and Cai, Carrie and Terry, Michael and Le, Quoc and others},
  journal={arXiv preprint arXiv:2108.07732},
  year={2021}
}

@inproceedings{AIME,
 author = {MAA},
 booktitle = {American Invitational Mathematics Examination - AIME},
 title = {American Invitational Mathematics Examination - AIME},
 url = {https://maa.org/math-competitions/american-invitational-mathematics-examination-aime},
 year = {2024}
}

@inproceedings{AIME25,
 author = {MAA},
 booktitle = {American Invitational Mathematics Examination - AIME},
 title = {American Invitational Mathematics Examination - AIME},
 url = {https://maa.org/math-competitions/american-invitational-mathematics-examination-aime},
 year = {2025}
}

@article{balunovic2025matharena,
  title={Matharena: Evaluating llms on uncontaminated math competitions},
  author={Balunovi{\'c}, Mislav and Dekoninck, Jasper and Petrov, Ivo and Jovanovi{\'c}, Nikola and Vechev, Martin},
  journal={arXiv preprint arXiv:2505.23281},
  year={2025}
}

@misc{zhou2023instructionfollowingevaluationlargelanguage,
      title={Instruction-Following Evaluation for Large Language Models}, 
      author={Jeffrey Zhou and Tianjian Lu and Swaroop Mishra and Siddhartha Brahma and Sujoy Basu and Yi Luan and Denny Zhou and Le Hou},
      year={2023},
      eprint={2311.07911},
      archivePrefix={arXiv},
      primaryClass={cs.CL},
      url={https://arxiv.org/abs/2311.07911}, 
}

@misc{arenahard2024,
    title = {From Live Data to High-Quality Benchmarks: The Arena-Hard Pipeline},
    url = {https://lmsys.org/blog/2024-04-19-arena-hard/},
    author = {Tianle Li and Wei-Lin Chiang and Evan Frick and Lisa Dunlap and Banghua Zhu and Joseph E. Gonzalez and Ion Stoica},
    month = {April},
    year = {2024}
}

@article{DBLP:journals/corr/abs-1904-10509,
  author       = {Rewon Child and
                  Scott Gray and
                  Alec Radford and
                  Ilya Sutskever},
  title        = {Generating Long Sequences with Sparse Transformers},
  journal      = {CoRR},
  volume       = {abs/1904.10509},
  year         = {2019},
  url          = {http://arxiv.org/abs/1904.10509},
  eprinttype    = {arXiv},
  eprint       = {1904.10509},
  bibsource    = {dblp computer science bibliography, https://dblp.org}
}

@article{fan2025megascience,
  title={MegaScience: Pushing the Frontiers of Post-Training Datasets for Science Reasoning},
  author={Fan, Run-Ze and Wang, Zengzhi and Liu, Pengfei},
  year={2025},
  journal={arXiv preprint arXiv:2507.16812},
  url={https://arxiv.org/abs/2507.16812}
}

@misc{li2023camel,
      title={CAMEL: Communicative Agents for "Mind" Exploration of Large Scale Language Model Society},
      author={Guohao Li and Hasan Abed Al Kader Hammoud and Hani Itani and Dmitrii Khizbullin and Bernard Ghanem},
      year={2023},
      eprint={2303.17760},
      archivePrefix={arXiv},
      primaryClass={cs.AI}
}

@misc{ye2025limoreasoning,
      title={LIMO: Less is More for Reasoning},
      author={Yixin Ye and Zhen Huang and Yang Xiao and Ethan Chern and Shijie Xia and Pengfei Liu},
      year={2025},
      eprint={2502.03387},
      archivePrefix={arXiv},
      primaryClass={cs.CL},
      url={https://arxiv.org/abs/2502.03387}, 
}

@inproceedings{
yang2025multiverse,
title={Multiverse: Your Language Models Secretly Decide How to Parallelize and Merge Generation},
author={Xinyu Yang and Yuwei An and Hongyi Liu and Tianqi Chen and Beidi Chen},
booktitle={The Thirty-ninth Annual Conference on Neural Information Processing Systems},
year={2025},
url={https://openreview.net/forum?id=r9YDEErKXU}
}

@inproceedings{muennighoff2025s1,
  title={s1: Simple test-time scaling},
  author={Muennighoff, Niklas and Yang, Zitong and Shi, Weijia and Li, Xiang Lisa and Fei-Fei, Li and Hajishirzi, Hannaneh and Zettlemoyer, Luke and Liang, Percy and Cand{\`e}s, Emmanuel and Hashimoto, Tatsunori B},
  booktitle={Proceedings of the 2025 Conference on Empirical Methods in Natural Language Processing},
  pages={20286--20332},
  year={2025}
}

@misc{zoph2022stmoedesigningstabletransferable,
      title={ST-MoE: Designing Stable and Transferable Sparse Expert Models}, 
      author={Barret Zoph and Irwan Bello and Sameer Kumar and Nan Du and Yanping Huang and Jeff Dean and Noam Shazeer and William Fedus},
      year={2022},
      eprint={2202.08906},
      archivePrefix={arXiv},
      primaryClass={cs.CL},
      url={https://arxiv.org/abs/2202.08906}, 
}

@InProceedings{pmlr-v162-du22c,
  title = 	 {{GL}a{M}: Efficient Scaling of Language Models with Mixture-of-Experts},
  author =       {Du, Nan and Huang, Yanping and Dai, Andrew M and Tong, Simon and Lepikhin, Dmitry and Xu, Yuanzhong and Krikun, Maxim and Zhou, Yanqi and Yu, Adams Wei and Firat, Orhan and Zoph, Barret and Fedus, Liam and Bosma, Maarten P and Zhou, Zongwei and Wang, Tao and Wang, Emma and Webster, Kellie and Pellat, Marie and Robinson, Kevin and Meier-Hellstern, Kathleen and Duke, Toju and Dixon, Lucas and Zhang, Kun and Le, Quoc and Wu, Yonghui and Chen, Zhifeng and Cui, Claire},
  booktitle = 	 {Proceedings of the 39th International Conference on Machine Learning},
  pages = 	 {5547--5569},
  year = 	 {2022},
  editor = 	 {Chaudhuri, Kamalika and Jegelka, Stefanie and Song, Le and Szepesvari, Csaba and Niu, Gang and Sabato, Sivan},
  volume = 	 {162},
  series = 	 {Proceedings of Machine Learning Research},
  month = 	 {17--23 Jul},
  publisher =    {PMLR},
  url = 	 {https://proceedings.mlr.press/v162/du22c.html}
}

@inproceedings{xiong2020layer,
  title={On layer normalization in the transformer architecture},
  author={Xiong, Ruibin and Yang, Yunchang and He, Di and Zheng, Kai and Zheng, Shuxin and Xing, Chen and Zhang, Huishuai and Lan, Yanyan and Wang, Liwei and Liu, Tieyan},
  booktitle={International conference on machine learning},
  pages={10524--10533},
  year={2020},
  organization={PMLR}
}

@article{radford2019language,
  title={Language Models are Unsupervised Multitask Learners},
  author={Radford, Alec and Wu, Jeff and Child, Rewon and Luan, David and Amodei, Dario and Sutskever, Ilya},
  year={2019}
}

@misc{shazeer2020gluvariantsimprovetransformer,
      title={GLU Variants Improve Transformer}, 
      author={Noam Shazeer},
      year={2020},
      eprint={2002.05202},
      archivePrefix={arXiv},
      primaryClass={cs.LG},
      url={https://arxiv.org/abs/2002.05202}, 
}

@article{li2024numinamath,
  title={Numinamath: The largest public dataset in ai4maths with 860k pairs of competition math problems and solutions},
  author={Li, Jia and Beeching, Edward and Tunstall, Lewis and Lipkin, Ben and Soletskyi, Roman and Huang, Shengyi and Rasul, Kashif and Yu, Longhui and Jiang, Albert Q and Shen, Ziju and others},
  year={2024}
}

@misc{ji2025amthinkingv1advancingfrontierreasoning,
      title={AM-Thinking-v1: Advancing the Frontier of Reasoning at 32B Scale}, 
      author={Yunjie Ji and Xiaoyu Tian and Sitong Zhao and Haotian Wang and Shuaiting Chen and Yiping Peng and Han Zhao and Xiangang Li},
      year={2025},
      eprint={2505.08311},
      archivePrefix={arXiv},
      primaryClass={cs.CL},
      url={https://arxiv.org/abs/2505.08311}, 
}

@misc{swe_verified,
  title={Introducing {SWE}-bench Verified
We’re releasing a human-validated subset of SWE-bench that more},
  author={OpenAI},
  year={2024},
  url = {https://openai.com/index/introducing-swe-bench-verified/}
}

@misc{luong2025robustmathematicalreasoning,
      title={Towards Robust Mathematical Reasoning}, 
      author={Thang Luong and Dawsen Hwang and Hoang H. Nguyen and Golnaz Ghiasi and Yuri Chervonyi and Insuk Seo and Junsu Kim and Garrett Bingham and Jonathan Lee and Swaroop Mishra and Alex Zhai and Clara Huiyi Hu and Henryk Michalewski and Jimin Kim and Jeonghyun Ahn and Junhwi Bae and Xingyou Song and Trieu H. Trinh and Quoc V. Le and Junehyuk Jung},
      year={2025},
      eprint={2511.01846},
      archivePrefix={arXiv},
      primaryClass={cs.CL},
      url={https://arxiv.org/abs/2511.01846}, 
}

@misc{hmmt25,
  title={HMMT 2025 Feb.},
  author={HMMT},
  year={2025},
  url = {https://www.hmmt.org/www/archive/282}
}

@misc{gpt5,
  title={Introducing GPT-5},
  author={OpenAI},
  year={2025},
  url = {https://openai.com/index/introducing-gpt-5/}
}

@misc{o3-mini,
    title={OpenAI o3-mini},
    author={OpenAI},
    year={2025},
    url={https://openai.com/index/openai-o3-mini/}
}

@article{kimi_k2,
  title={Kimi k2: Open agentic intelligence},
  author={Team, Kimi and Bai, Yifan and Bao, Yiping and Chen, Guanduo and Chen, Jiahao and Chen, Ningxin and Chen, Ruijue and Chen, Yanru and Chen, Yuankun and Chen, Yutian and others},
  journal={arXiv preprint arXiv:2507.20534},
  year={2025}
}

@article{liu2025webexplorer,
  title={Webexplorer: Explore and evolve for training long-horizon web agents},
  author={Liu, Junteng and Li, Yunji and Zhang, Chi and Li, Jingyang and Chen, Aili and Ji, Ke and Cheng, Weiyu and Wu, Zijia and Du, Chengyu and Xu, Qidi and others},
  journal={arXiv preprint arXiv:2509.06501},
  year={2025}
}

@article{li2025websailor,
  title={WebSailor: Navigating Super-human Reasoning for Web Agent},
  author={Li, Kuan and Zhang, Zhongwang and Yin, Huifeng and Zhang, Liwen and Ou, Litu and Wu, Jialong and Yin, Wenbiao and Li, Baixuan and Tao, Zhengwei and Wang, Xinyu and others},
  journal={arXiv preprint arXiv:2507.02592},
  year={2025}
}

@article{li2025simulating,
  title={Simulating environments with reasoning models for agent training},
  author={Li, Yuetai and Inan, Huseyin A and Yue, Xiang and Chen, Wei-Ning and Wutschitz, Lukas and Kulkarni, Janardhan and Poovendran, Radha and Sim, Robert and Rajmohan, Saravan},
  journal={arXiv preprint arXiv:2511.01824},
  year={2025}
}

@misc{minimax_m2p1,
  title={MiniMax-M2.1},
  author={MiniMax Team},
  year={2025},
  url = {https://huggingface.co/MiniMaxAI/MiniMax-M2.1}
}

@inproceedings{li2024eagle, 
	author = {Yuhui Li and Fangyun Wei and Chao Zhang and Hongyang Zhang}, 
	title = {{EAGLE}: Speculative Sampling Requires Rethinking Feature Uncertainty}, 
	booktitle = {International Conference on Machine Learning},
	year = {2024}
}

@misc{yu2025dapoopensourcellmreinforcement,
  title         = {{DAPO}: An Open-Source {LLM} Reinforcement Learning System at Scale},
  author        = {Yu, Qiying and Zhang, Zheng and Zhu, Ruofei and Yuan, Yufeng and Zuo, Xiaochen and Yue, Yu and Fan, Tiantian and Liu, Gaohong and Liu, Lingjun and Liu, Xin and Lin, Haibin and Lin, Zhiqi and Ma, Bole and Sheng, Guangming and Tong, Yuxuan and Zhang, Chi and Zhang, Mofan and Zhang, Wang and Zhu, Hang and Zhu, Jinhua and Chen, Jiaze and Chen, Jiangjie and Wang, Chengyi and Yu, Hongli and Dai, Weinan and Song, Yuxuan and Wei, Xiangpeng and Zhou, Hao and Liu, Jingjing and Ma, Wei-Ying and Zhang, Ya-Qin and Yan, Lin and Qiao, Mu and Wu, Yonghui and Wang, Mingxuan},
  year          = {2025},
  eprint        = {2503.14476},
  archivePrefix = {arXiv},
  primaryClass  = {cs.LG},
  url           = {https://arxiv.org/abs/2503.14476}
}

@misc{anthropic_agent_workflow,
  title = {Building effective agents},
  howpublished = {\url{https://www.anthropic.com/engineering/building-effective-agents}},
}

@misc{codex_agent_workflow,
  title = {Unrolling the Codex agent loop},
  howpublished = {\url{https://openai.com/index/unrolling-the-codex-agent-loop/}},
}

@inproceedings{10.5555/3618408.3619203,
author = {Leviathan, Yaniv and Kalman, Matan and Matias, Yossi},
title = {Fast inference from transformers via speculative decoding},
year = {2023},
publisher = {JMLR.org},
booktitle = {Proceedings of the 40th International Conference on Machine Learning},
articleno = {795},
numpages = {13},
location = {Honolulu, Hawaii, USA},
series = {ICML'23}
}

@inproceedings{
sun2024massive,
title={Massive Activations in Large Language Models},
author={Mingjie Sun and Xinlei Chen and J Zico Kolter and Zhuang Liu},
booktitle={First Conference on Language Modeling},
year={2024},
url={https://openreview.net/forum?id=F7aAhfitX6}
}

@misc{cohere2025commandaenterprisereadylarge,
      title={Command A: An Enterprise-Ready Large Language Model}, 
      author={Team Cohere and : and Aakanksha and Arash Ahmadian and Marwan Ahmed and others},
      year={2025},
      eprint={2504.00698},
      archivePrefix={arXiv},
      primaryClass={cs.CL},
      url={https://arxiv.org/abs/2504.00698}, 
}

@inproceedings{
xiao2024efficient,
title={Efficient Streaming Language Models with Attention Sinks},
author={Guangxuan Xiao and Yuandong Tian and Beidi Chen and Song Han and Mike Lewis},
booktitle={The Twelfth International Conference on Learning Representations},
year={2024},
url={https://openreview.net/forum?id=NG7sS51zVF}
}

@inproceedings{
gu2025when,
title={When Attention Sink Emerges in Language Models: An Empirical View},
author={Xiangming Gu and Tianyu Pang and Chao Du and Qian Liu and Fengzhuo Zhang and Cunxiao Du and Ye Wang and Min Lin},
booktitle={The Thirteenth International Conference on Learning Representations},
year={2025},
url={https://openreview.net/forum?id=78Nn4QJTEN}
}

@article{xiong2025dyspec,
  title={DySpec: Faster speculative decoding with dynamic token tree structure},
  author={Xiong, Yunfan and Zhang, Ruoyu and Li, Yanzeng and Zou, Lei},
  journal={World Wide Web},
  volume={28},
  number={3},
  pages={36},
  year={2025},
  publisher={Springer}
}

@article{wang2025opt,
  title={Opt-tree: Speculative decoding with adaptive draft tree structure},
  author={Wang, Jikai and Su, Yi and Li, Juntao and Xia, Qingrong and Ye, Zi and Duan, Xinyu and Wang, Zhefeng and Zhang, Min},
  journal={Transactions of the Association for Computational Linguistics},
  volume={13},
  pages={188--199},
  year={2025},
  publisher={MIT Press 255 Main Street, 9th Floor, Cambridge, Massachusetts 02142, USA~…}
}

@inproceedings{
an2025systematic,
title={Systematic Outliers in Large Language Models},
author={Yongqi An and Xu Zhao and Tao Yu and Ming Tang and Jinqiao Wang},
booktitle={The Thirteenth International Conference on Learning Representations},
year={2025},
url={https://openreview.net/forum?id=rLX7Vyyzus}
}

@inproceedings{10.5555/3524938.3525416,
author = {Katharopoulos, Angelos and Vyas, Apoorv and Pappas, Nikolaos and Fleuret, Fran\c{c}ois},
title = {Transformers are RNNs: fast autoregressive transformers with linear attention},
year = {2020},
publisher = {JMLR.org},
booktitle = {Proceedings of the 37th International Conference on Machine Learning},
articleno = {478},
numpages = {10},
series = {ICML'20}
}

@inproceedings{schlag2021linear,
  title={Linear transformers are secretly fast weight programmers},
  author={Schlag, Imanol and Irie, Kazuki and Schmidhuber, J{\"u}rgen},
  booktitle={International conference on machine learning},
  pages={9355--9366},
  year={2021},
  organization={PMLR}
}

@inproceedings{ainslie-etal-2023-gqa,
    title = "{GQA}: Training Generalized Multi-Query Transformer Models from Multi-Head Checkpoints",
    author = "Ainslie, Joshua  and
      Lee-Thorp, James  and
      de Jong, Michiel  and
      Zemlyanskiy, Yury  and
      Lebron, Federico  and
      Sanghai, Sumit",
    editor = "Bouamor, Houda  and
      Pino, Juan  and
      Bali, Kalika",
    booktitle = "Proceedings of the 2023 Conference on Empirical Methods in Natural Language Processing",
    month = dec,
    year = "2023",
    address = "Singapore",
    publisher = "Association for Computational Linguistics",
    url = "https://aclanthology.org/2023.emnlp-main.298/",
    doi = "10.18653/v1/2023.emnlp-main.298",
    pages = "4895--4901",
}

@inproceedings{10.5555/3692070.3694435,
author = {You, Haoran and Fu, Yichao and Wang, Zheng and Yazdanbakhsh, Amir and Lin, Yingyan (Celine)},
title = {When linear attention meets autoregressive decoding: towards more effective and efficient linearized large language models},
year = {2024},
publisher = {JMLR.org},
booktitle = {Proceedings of the 41st International Conference on Machine Learning},
articleno = {2365},
numpages = {17},
location = {Vienna, Austria},
series = {ICML'24}
}

@misc{gemmateam2025gemma3technicalreport,
      title={Gemma 3 Technical Report}, 
      author={Gemma Team and Aishwarya Kamath and Johan Ferret and others},
      year={2025},
      eprint={2503.19786},
      archivePrefix={arXiv},
      primaryClass={cs.CL},
      url={https://arxiv.org/abs/2503.19786}, 
}

@article{Beltagy2020Longformer,
  title={Longformer: The Long-Document Transformer},
  author={Iz Beltagy and Matthew E. Peters and Arman Cohan},
  journal={arXiv:2004.05150},
  year={2020},
}

@article{hu2025openreasonerzero,
  title   = {{Open-Reasoner-Zero}: An Open Source Approach to Scaling Up Reinforcement Learning on the Base Model},
  author  = {Hu, Jingcheng and Zhang, Yinmin and Han, Qi and Jiang, Daxin and Zhang, Xiangyu and Shum, Heung-Yeung},
  journal = {arXiv preprint arXiv:2503.24290},
  year    = {2025}
}

@article{guha2025openthoughts,
  title   = {OpenThoughts: Data Recipes for Reasoning Models},
  author  = {Guha, Etash and others},
  journal = {arXiv preprint arXiv:2506.04178},
  year    = {2025}
}

@article{albalak2025bigmath,
  title   = {Big-Math: A Large-Scale, High-Quality Math Dataset for Reinforcement Learning in Language Models},
  author  = {Albalak, Alon and others},
  journal = {arXiv preprint arXiv:2502.17387},
  year    = {2025}
}

@article{mitra2024orcamath,
  title   = {Orca-Math: Unlocking the Potential of SLMs in Grade School Math},
  author  = {Mitra, Arindam and Khanpour, Hamed and Rosset, Corby and Awadallah, Ahmed},
  journal = {arXiv preprint arXiv:2402.14830},
  year    = {2024}
}

@misc{aslawliet2024olympiads,
  title        = {Olympiads},
  author       = {aslawliet},
  year         = {2024},
  howpublished = {\url{https://huggingface.co/datasets/aslawliet/olympiads}},
  note         = {Hugging Face dataset}
}

@misc{aslawliet2024cnk12,
  title        = {CN-K12},
  author       = {aslawliet},
  year         = {2024},
  howpublished = {\url{https://huggingface.co/datasets/aslawliet/cn-k12}},
  note         = {Hugging Face dataset of Chinese K-12 math problems}
}

@misc{openr12025math220k,
  title        = {OpenR1-Math-220k},
  author       = {{Open-R1 Team}},
  year         = {2025},
  howpublished = {\url{https://huggingface.co/datasets/open-r1/OpenR1-Math-220k}},
  note         = {Open-source distilled math reasoning dataset}
}

@article{limo2025lessismore,
  title   = {Less is More for Reasoning: Semi-Parametric Math Reasoners},
  author  = {{LIMO Authors}},
  journal = {arXiv preprint arXiv:2502.03387},
  year    = {2025}
}

@misc{muennighoff2025s1simpletesttimescaling,
  title        = {s1: Simple Test-Time Scaling},
  author       = {Muennighoff, Niklas and others},
  year         = {2025},
  eprint       = {2501.19393},
  archivePrefix = {arXiv},
  primaryClass = {cs.CL},
  howpublished = {\url{https://arxiv.org/abs/2501.19393}}
}

@article{deepmath2025deepmath103k,
  title   = {DeepMath-103K: A Large-Scale, Challenging Math QA Benchmark},
  author  = {He, X. and others},
  journal = {arXiv preprint arXiv:2504.11456},
  year    = {2025}
}

@misc{sirdeshmukh2025multichallengerealisticmultiturnconversation,
      title={MultiChallenge: A Realistic Multi-Turn Conversation Evaluation Benchmark Challenging to Frontier LLMs}, 
      author={Ved Sirdeshmukh and Kaustubh Deshpande and Johannes Mols and Lifeng Jin and Ed-Yeremai Cardona and Dean Lee and Jeremy Kritz and Willow Primack and Summer Yue and Chen Xing},
      year={2025},
      eprint={2501.17399},
      archivePrefix={arXiv},
      primaryClass={cs.CL},
      url={https://arxiv.org/abs/2501.17399}, 
}

@article{jain2024livecodebench,
    title={LiveCodeBench: Holistic and Contamination Free Evaluation of Large Language Models for Code},
    author={Jain, Naman and Han, King and Gu, Alex and Li, Wen-Ding and Yan, Fanjia and Zhang, Tianjun and Wang, Sida and Solar-Lezama, Armando and Sen, Koushik and Stoica, Ion},
    journal={arXiv preprint arXiv:2403.07974},
    year={2024}
}

@misc{yang2025qwen3technicalreport,
      title={Qwen3 Technical Report}, 
      author={An Yang and Anfeng Li and Baosong Yang and Beichen Zhang and Binyuan Hui and Bo Zheng and Bowen Yu and Chang Gao and Chengen Huang and Chenxu Lv and Chujie Zheng and Dayiheng Liu and Fan Zhou and Fei Huang and Feng Hu and Hao Ge and Haoran Wei and Huan Lin and Jialong Tang and Jian Yang and Jianhong Tu and Jianwei Zhang and Jianxin Yang and Jiaxi Yang and Jing Zhou and Jingren Zhou and Junyang Lin and Kai Dang and Keqin Bao and Kexin Yang and Le Yu and Lianghao Deng and Mei Li and Mingfeng Xue and Mingze Li and Pei Zhang and Peng Wang and Qin Zhu and Rui Men and Ruize Gao and Shixuan Liu and Shuang Luo and Tianhao Li and Tianyi Tang and Wenbiao Yin and Xingzhang Ren and Xinyu Wang and Xinyu Zhang and Xuancheng Ren and Yang Fan and Yang Su and Yichang Zhang and Yinger Zhang and Yu Wan and Yuqiong Liu and Zekun Wang and Zeyu Cui and Zhenru Zhang and Zhipeng Zhou and Zihan Qiu},
      year={2025},
      eprint={2505.09388},
      archivePrefix={arXiv},
      primaryClass={cs.CL},
      url={https://arxiv.org/abs/2505.09388}, 
}

@misc{deepcoder2025,
  title={DeepCoder: A Fully Open-Source 14B Coder at O3-mini Level},
  author={Michael Luo and Sijun Tan and Roy Huang and Ameen Patel and Alpay Ariyak and Qingyang Wu and Xiaoxiang Shi and Rachel Xin and Colin Cai and Maurice Weber and Ce Zhang and Li Erran Li and Raluca Ada Popa and Ion Stoica},
  howpublished={\url{https://www.together.ai/blog/deepcoder}},
  note={Technical Blog},
  year={2025}
}

@misc{li2023tacotopicsalgorithmiccode,
      title={TACO: Topics in Algorithmic COde generation dataset}, 
      author={Rongao Li and Jie Fu and Bo-Wen Zhang and Tao Huang and Zhihong Sun and Chen Lyu and Guang Liu and Zhi Jin and Ge Li},
      year={2023},
      eprint={2312.14852},
      archivePrefix={arXiv},
      primaryClass={cs.AI},
      url={https://arxiv.org/abs/2312.14852}, 
}

@misc{wang2025codecontestshighqualitytestcase,
      title={CodeContests+: High-Quality Test Case Generation for Competitive Programming}, 
      author={Zihan Wang and Siyao Liu and Yang Sun and Hongyan Li and Kai Shen},
      year={2025},
      eprint={2506.05817},
      archivePrefix={arXiv},
      primaryClass={cs.SE},
      url={https://arxiv.org/abs/2506.05817}, 
}

@misc{yao2025offpolicy,
  title = {Your Efficient RL Framework Secretly Brings You Off-Policy RL Training},
  url = {https://fengyao.notion.site/off-policy-rl},
  author = {Yao, Feng and Liu, Liyuan and Zhang, Dinghuai and Dong, Chengyu and Shang, Jingbo and Gao, Jianfeng},
  journal = {Feng Yao's Notion},
  year = {2025},
  month = aug,
}

@misc{peng2023yarnefficientcontextwindow,
      title={YaRN: Efficient Context Window Extension of Large Language Models}, 
      author={Bowen Peng and Jeffrey Quesnelle and Honglu Fan and Enrico Shippole},
      year={2023},
      eprint={2309.00071},
      archivePrefix={arXiv},
      primaryClass={cs.CL},
      url={https://arxiv.org/abs/2309.00071}, 
}

@article{jumper_highly_2021,
	title = {Highly accurate protein structure prediction with {AlphaFold}},
	volume = {596},
	issn = {0028-0836, 1476-4687},
	url = {https://www.nature.com/articles/s41586-021-03819-2},
	doi = {10.1038/s41586-021-03819-2},
	language = {en},
	number = {7873},
	urldate = {2026-02-04},
	journal = {Nature},
	author = {Jumper, John and Evans, Richard and Pritzel, Alexander and Green, Tim and Figurnov, Michael and Ronneberger, Olaf and Tunyasuvunakool, Kathryn and Bates, Russ and Žídek, Augustin and Potapenko, Anna and Bridgland, Alex and Meyer, Clemens and Kohl, Simon A. A. and Ballard, Andrew J. and Cowie, Andrew and Romera-Paredes, Bernardino and Nikolov, Stanislav and Jain, Rishub and Adler, Jonas and Back, Trevor and Petersen, Stig and Reiman, David and Clancy, Ellen and Zielinski, Michal and Steinegger, Martin and Pacholska, Michalina and Berghammer, Tamas and Bodenstein, Sebastian and Silver, David and Vinyals, Oriol and Senior, Andrew W. and Kavukcuoglu, Koray and Kohli, Pushmeet and Hassabis, Demis},
	month = aug,
	year = {2021},
	pages = {583--589},
}

@article{zheng2025group,
  title={Group sequence policy optimization},
  author={Zheng, Chujie and Liu, Shixuan and Li, Mingze and Chen, Xiong-Hui and Yu, Bowen and Gao, Chang and Dang, Kai and Liu, Yuqiong and Men, Rui and Yang, An and others},
  journal={arXiv preprint arXiv:2507.18071},
  year={2025}
}

@article{ma2025stabilizing,
  title={Stabilizing moe reinforcement learning by aligning training and inference routers},
  author={Ma, Wenhan and Zhang, Hailin and Zhao, Liang and Song, Yifan and Wang, Yudong and Sui, Zhifang and Luo, Fuli},
  journal={arXiv preprint arXiv:2510.11370},
  year={2025}
}

@misc{hu2026pacorelearningscaletesttime,
      title={PaCoRe: Learning to Scale Test-Time Compute with Parallel Coordinated Reasoning}, 
      author={Jingcheng Hu and Yinmin Zhang and Shijie Shang and Xiaobo Yang and Yue Peng and Zhewei Huang and Hebin Zhou and Xin Wu and Jie Cheng and Fanqi Wan and Xiangwen Kong and Chengyuan Yao and Kaiwen Yan and Ailin Huang and Hongyu Zhou and Qi Han and Zheng Ge and Daxin Jiang and Xiangyu Zhang and Heung-Yeung Shum},
      year={2026},
      eprint={2601.05593},
      archivePrefix={arXiv},
      primaryClass={cs.LG},
      url={https://arxiv.org/abs/2601.05593}, 
}

@misc{pardo2022timelimitsreinforcementlearning,
      title={Time Limits in Reinforcement Learning}, 
      author={Fabio Pardo and Arash Tavakoli and Vitaly Levdik and Petar Kormushev},
      year={2022},
      eprint={1712.00378},
      archivePrefix={arXiv},
      primaryClass={cs.LG},
      url={https://arxiv.org/abs/1712.00378}, 
}

@inproceedings{
lin2025forgetting,
title={Forgetting Transformer: Softmax Attention with a Forget Gate},
author={Zhixuan Lin and Evgenii Nikishin and Xu He and Aaron Courville},
booktitle={The Thirteenth International Conference on Learning Representations},
year={2025},
url={https://openreview.net/forum?id=q2Lnyegkr8}
}

@article{zhao2024wildchat,
  title={Wildchat: 1m chatgpt interaction logs in the wild},
  author={Zhao, Wenting and Ren, Xiang and Hessel, Jack and Cardie, Claire and Choi, Yejin and Deng, Yuntian},
  journal={arXiv preprint arXiv:2405.01470},
  year={2024}
}

@misc{deepscaler2025,
  title={DeepScaleR: Surpassing O1-Preview with a 1.5B Model by Scaling RL},
  author={Michael Luo and Sijun Tan and Justin Wong and Xiaoxiang Shi and William Y. Tang and Manan Roongta and Colin Cai and Jeffrey Luo and Li Erran Li and Raluca Ada Popa and Ion Stoica},
  year={2025},
  howpublished={\url{https://pretty-radio-b75.notion.site/DeepScaleR-Surpassing-O1-Preview-with-a-1-5B-Model-by-Scaling-RL-19681902c1468005bed8ca303013a4e2}},
  note={Notion Blog}
}

@article{yu2025dapo,
  title={Dapo: An open-source llm reinforcement learning system at scale},
  author={Yu, Qiying and Zhang, Zheng and Zhu, Ruofei and Yuan, Yufeng and Zuo, Xiaochen and Yue, Yu and Dai, Weinan and Fan, Tiantian and Liu, Gaohong and Liu, Lingjun and others},
  journal={arXiv preprint arXiv:2503.14476},
  year={2025}
}

@misc{bercovich2025llamanemotronefficientreasoningmodels,
      title={Llama-Nemotron: Efficient Reasoning Models}, 
      author={Bercovich, Akhiad and Levy, Itay and Golan, Izik and Dabbah, Mohammad and El-Yaniv, Ran and Puny, Omri and Galil, Ido and Moshe, Zach and Ronen, Tomer and Nabwani, Najeeb and others},
      year={2025},
      eprint={2505.00949},
      archivePrefix={arXiv},
      primaryClass={cs.CL},
      url={https://arxiv.org/abs/2505.00949}, 
}

@inproceedings{
snell2025scaling,
title={Scaling {LLM} Test-Time Compute Optimally Can be More Effective than Scaling Parameters for Reasoning},
author={Charlie Victor Snell and Jaehoon Lee and Kelvin Xu and Aviral Kumar},
booktitle={The Thirteenth International Conference on Learning Representations},
year={2025},
url={https://openreview.net/forum?id=4FWAwZtd2n}
}

@article{liu2025deepseek,
  title={Deepseek-v3. 2: Pushing the frontier of open large language models},
  author={Liu, Aixin and Mei, Aoxue and Lin, Bangcai and Xue, Bing and Wang, Bingxuan and Xu, Bingzheng and Wu, Bochao and Zhang, Bowei and Lin, Chaofan and Dong, Chen and others},
  journal={arXiv preprint arXiv:2512.02556},
  year={2025}
}

@misc{hu2025stepdeepresearchtechnicalreport,
      title={Step-DeepResearch Technical Report}, 
      author={Chen Hu and Haikuo Du and Heng Wang and Lin Lin and Mingrui Chen and Peng Liu and Ruihang Miao and Tianchi Yue and Wang You and Wei Ji and Wei Yuan and Wenjin Deng and Xiaojian Yuan and Xiaoyun Zhang and Xiangyu Liu and Xikai Liu and Yanming Xu and Yicheng Cao and Yifei Zhang and Yongyao Wang and Yubo Shu and Yurong Zhang and Yuxiang Zhang and Zheng Gong and Zhichao Chang and Binyan Li and Dan Ma and Furong Jia and Hongyuan Wang and Jiayu Liu and Jing Bai and Junlan Liu and Manjiao Liu and Na Wang and Qiuping Wu and Qinxin Du and Shiwei Li and Wen Sun and Yifeng Gong and Yonglin Chen and Yuling Zhao and Yuxuan Lin and Ziqi Ren and Zixuan Wang and Aihu Zhang and Brian Li and Buyun Ma and Kang An and Li Xie and Mingliang Li and Pan Li and Shidong Yang and Xi Chen and Xiaojia Liu and Yuchu Luo and Yuan Song and YuanHao Ding and Yuanwei Liang and Zexi Li and Zhaoning Zhang and Zixin Zhang and Binxing Jiao and Daxin Jiang and Jiansheng Chen and Jing Li and Xiangyu Zhang and Yibo Zhu},
      year={2025},
      eprint={2512.20491},
      archivePrefix={arXiv},
      primaryClass={cs.CL},
      url={https://arxiv.org/abs/2512.20491}, 
}

@article{Guo_2025,
   title={DeepSeek-R1 incentivizes reasoning in LLMs through reinforcement learning},
   volume={645},
   ISSN={1476-4687},
   url={http://dx.doi.org/10.1038/s41586-025-09422-z},
   DOI={10.1038/s41586-025-09422-z},
   number={8081},
   journal={Nature},
   publisher={Springer Science and Business Media LLC},
   author={Guo, Daya and Yang, Dejian and Zhang, Haowei and Song, Junxiao and Zhang, Ruoyu and Xu, Runxin and Zhu, Qihao and Ma, Shirong and Wang, Peiyi and Bi, Xiao and others},
   year={2025},
   month=sep, pages={633–638}
}

@article{wang2024openhands,
  title={Openhands: An open platform for ai software developers as generalist agents},
  author={Wang, Xingyao and Li, Boxuan and Song, Yufan and Xu, Frank F and Tang, Xiangru and Zhuge, Mingchen and Pan, Jiayi and Song, Yueqi and Li, Bowen and Singh, Jaskirat and others},
  journal={arXiv preprint arXiv:2407.16741},
  year={2024}
}

@article{metropolis1953equation,
  title={Equation of state calculations by fast computing machines},
  author={Metropolis, Nicholas and Rosenbluth, Arianna W and Rosenbluth, Marshall N and Teller, Augusta H and Teller, Edward},
  journal={The journal of chemical physics},
  volume={21},
  number={6},
  pages={1087--1092},
  year={1953},
  publisher={American Institute of Physics}
}

@article{hastings1970monte,
  title={Monte Carlo sampling methods using Markov chains and their applications},
  author={Hastings, W Keith},
  year={1970},
  publisher={Oxford University Press}
}

@article{chollet2019measure,
  title={On the measure of intelligence},
  author={Chollet, Fran{\c{c}}ois},
  journal={arXiv preprint arXiv:1911.01547},
  year={2019}
}

@article{yang2024swe,
  title={Swe-agent: Agent-computer interfaces enable automated software engineering},
  author={Yang, John and Jimenez, Carlos E and Wettig, Alexander and Lieret, Kilian and Yao, Shunyu and Narasimhan, Karthik and Press, Ofir},
  journal={Advances in Neural Information Processing Systems},
  volume={37},
  pages={50528--50652},
  year={2024}
}

@inproceedings{luong2025towards,
  title={Towards robust mathematical reasoning},
  author={Luong, Minh-Thang and Hwang, Dawsen and Nguyen, Hoang H and Ghiasi, Golnaz and Chervonyi, Yuri and Seo, Insuk and Kim, Junsu and Bingham, Garrett and Lee, Jonathan and Mishra, Swaroop and others},
  booktitle={Proceedings of the 2025 Conference on Empirical Methods in Natural Language Processing},
  pages={35406--35430},
  year={2025}
}

@article{wang2021kepler,
  title={KEPLER: A unified model for knowledge embedding and pre-trained language representation},
  author={Wang, Xiaozhi and Gao, Tianyu and Zhu, Zhaocheng and Zhang, Zhengyan and Liu, Zhiyuan and Li, Juanzi and Tang, Jian},
  journal={Transactions of the Association for Computational Linguistics},
  volume={9},
  pages={176--194},
  year={2021},
  publisher={MIT Press One Rogers Street, Cambridge, MA 02142-1209, USA journals-info~…}
}

@article{chen2025xbench,
  title={xbench: Tracking Agents Productivity Scaling with Profession-Aligned Real-World Evaluations},
  author={Chen, Kaiyuan and Ren, Yixin and Liu, Yang and Hu, Xiaobo and Tian, Haotong and Xie, Tianbao and Liu, Fangfu and Zhang, Haoye and Liu, Hongzhang and Gong, Yuan and others},
  journal={arXiv preprint arXiv:2506.13651},
  year={2025}
}

@article{pyatkin2025generalizing,
  title={Generalizing Verifiable Instruction Following},
  author={Pyatkin, Valentina and Malik, Saumya and Graf, Victoria and Ivison, Hamish and Huang, Shengyi and Dasigi, Pradeep and Lambert, Nathan and Hajishirzi, Hannaneh},
  journal={arXiv preprint arXiv:2507.02833},
  year={2025}
}

@article{vodrahalli2024michelangelo,
  title={Michelangelo: Long context evaluations beyond haystacks via latent structure queries},
  author={Vodrahalli, Kiran and Ontanon, Santiago and Tripuraneni, Nilesh and Xu, Kelvin and Jain, Sanil and Shivanna, Rakesh and Hui, Jeffrey and Dikkala, Nishanth and Kazemi, Mehran and Fatemi, Bahare and others},
  journal={arXiv preprint arXiv:2409.12640},
  year={2024}
}

@inproceedings{
anonymous2026autocode,
title={AutoCode: {LLM}s as Problem Setters for Competitive Programming},
author={Anonymous},
booktitle={The Fourteenth International Conference on Learning Representations},
year={2026},
url={https://openreview.net/forum?id=F96nsbbhXC}
}

@article{zheng2025livecodebench,
  title={LiveCodeBench Pro: How Do Olympiad Medalists Judge LLMs in Competitive Programming?},
  author={Zheng, Zihan and Cheng, Zerui and Shen, Zeyu and Zhou, Shang and Liu, Kaiyuan and He, Hansen and Li, Dongruixuan and Wei, Stanley and Hao, Hangyi and Yao, Jianzhu and others},
  journal={arXiv preprint arXiv:2506.11928},
  year={2025}
}

@article{pan2024training,
  title={Training software engineering agents and verifiers with swe-gym},
  author={Pan, Jiayi and Wang, Xingyao and Neubig, Graham and Jaitly, Navdeep and Ji, Heng and Suhr, Alane and Zhang, Yizhe},
  journal={arXiv preprint arXiv:2412.21139},
  year={2024}
}

@misc{seta,
  author = {Qijia Shen and Jay Rainton and Aznaur Aliev and Ahmed Awelkair and Boyuan Ma and Zhiqi Huang and Yuzhen Mao and Wendong Fan and Philip Torr and Bernard Ghanem and Changran Hu and Urmish Thakker and Guohao Li},
  month = Jan,
  title = {{SETA: Scaling Environments for Terminal Agents}},
  year = {2026}
}

@article{badertdinov2025swe,
  title={SWE-rebench: An Automated Pipeline for Task Collection and Decontaminated Evaluation of Software Engineering Agents},
  author={Badertdinov, Ibragim and Golubev, Alexander and Nekrashevich, Maksim and Shevtsov, Anton and Karasik, Simon and Andriushchenko, Andrei and Trofimova, Maria and Litvintseva, Daria and Yangel, Boris},
  journal={arXiv preprint arXiv:2505.20411},
  year={2025}
}

@inproceedings{krishna2025fact,
  title={Fact, fetch, and reason: A unified evaluation of retrieval-augmented generation},
  author={Krishna, Satyapriya and Krishna, Kalpesh and Mohananey, Anhad and Schwarcz, Steven and Stambler, Adam and Upadhyay, Shyam and Faruqui, Manaal},
  booktitle={Proceedings of the 2025 Conference of the Nations of the Americas Chapter of the Association for Computational Linguistics: Human Language Technologies (Volume 1: Long Papers)},
  pages={4745--4759},
  year={2025}
}

@article{jain2025r2e,
  title={R2e-gym: Procedural environments and hybrid verifiers for scaling open-weights swe agents},
  author={Jain, Naman and Singh, Jaskirat and Shetty, Manish and Zheng, Liang and Sen, Koushik and Stoica, Ion},
  journal={arXiv preprint arXiv:2504.07164},
  year={2025}
}

@article{yang2025swe,
  title={Swe-smith: Scaling data for software engineering agents},
  author={Yang, John and Lieret, Kilian and Jimenez, Carlos E and Wettig, Alexander and Khandpur, Kabir and Zhang, Yanzhe and Hui, Binyuan and Press, Ofir and Schmidt, Ludwig and Yang, Diyi},
  journal={arXiv preprint arXiv:2504.21798},
  year={2025}
}

@article{zeng2025glm,
  title={Glm-4.5: Agentic, reasoning, and coding (arc) foundation models},
  author={Zeng, Aohan and Lv, Xin and Zheng, Qinkai and Hou, Zhenyu and Chen, Bin and Xie, Chengxing and Wang, Cunxiang and Yin, Da and Zeng, Hao and Zhang, Jiajie and others},
  journal={arXiv preprint arXiv:2508.06471},
  year={2025}
}

@article{yan2025step,
  title={Step-gui technical report},
  author={Yan, Haolong and Wang, Jia and Huang, Xin and Shen, Yeqing and Meng, Ziyang and Fan, Zhimin and Tan, Kaijun and Gao, Jin and Shi, Lieyu and Yang, Mi and others},
  journal={arXiv preprint arXiv:2512.15431},
  year={2025}
}

@misc{quan2025codeelobenchmarkingcompetitionlevelcode,
      title={CodeElo: Benchmarking Competition-level Code Generation of LLMs with Human-comparable Elo Ratings}, 
      author={Shanghaoran Quan and Jiaxi Yang and Bowen Yu and Bo Zheng and Dayiheng Liu and An Yang and Xuancheng Ren and Bofei Gao and Yibo Miao and Yunlong Feng and Zekun Wang and Jian Yang and Zeyu Cui and Yang Fan and Yichang Zhang and Binyuan Hui and Junyang Lin},
      year={2025},
      eprint={2501.01257},
      archivePrefix={arXiv},
      primaryClass={cs.CL},
      url={https://arxiv.org/abs/2501.01257}, 
}

@misc{mialon2023gaiabenchmarkgeneralai,
      title={GAIA: a benchmark for General AI Assistants}, 
      author={Grégoire Mialon and Clémentine Fourrier and Craig Swift and Thomas Wolf and Yann LeCun and Thomas Scialom},
      year={2023},
      eprint={2311.12983},
      archivePrefix={arXiv},
      primaryClass={cs.CL},
      url={https://arxiv.org/abs/2311.12983}, 
}

@misc{zhou2025browsecompzhbenchmarkingwebbrowsing,
      title={BrowseComp-ZH: Benchmarking Web Browsing Ability of Large Language Models in Chinese}, 
      author={Peilin Zhou and Bruce Leon and Xiang Ying and Can Zhang and Yifan Shao and Qichen Ye and Dading Chong and Zhiling Jin and Chenxuan Xie and Meng Cao and Yuxin Gu and Sixin Hong and Jing Ren and Jian Chen and Chao Liu and Yining Hua},
      year={2025},
      eprint={2504.19314},
      archivePrefix={arXiv},
      primaryClass={cs.CL},
      url={https://arxiv.org/abs/2504.19314}, 
}

@article{wei2024measuring,
  title={Measuring short-form factuality in large language models},
  author={Wei, Jason and Karina, Nguyen and Chung, Hyung Won and Jiao, Yunxin Joy and Papay, Spencer and Glaese, Amelia and Schulman, John and Fedus, William},
  journal={arXiv preprint arXiv:2411.04368},
  year={2024}
}

@misc{openai2024gpt4technicalreport,
      title={GPT-4 Technical Report}, 
      author={OpenAI and Achiam, Josh and Adler, Steven and Agarwal, Sandhini and Ahmad, Lama and Akkaya, Ilge and Aleman, Florencia Leoni and Almeida, Diogo and Altenschmidt, Janko and Altman, Sam and Anadkat, Shyamal and others},
      year={2024},
      eprint={2303.08774},
      archivePrefix={arXiv},
      primaryClass={cs.CL},
      url={https://arxiv.org/abs/2303.08774}, 
}

@article{team2025longcat,
  title={Longcat-flash technical report},
  author={Team, Meituan LongCat and Li, Bei and Lei, Bingye and Wang, Bo and Rong, Bolin and Wang, Chao and Zhang, Chao and Gao, Chen and Zhang, Chen and Sun, Cheng and others},
  journal={arXiv preprint arXiv:2509.01322},
  year={2025}
}

@misc{gpt_5p2,
  title={GPT-5.2},
  author={OpenAI},
  year={2025},
  url = {https://openai.com/index/introducing-gpt-5-2/}
}

@misc{gemini_3,
  title={Gemini 3 ProModel Card},
  author={Google DeepMind},
  year={2025},
  url = {https://storage.googleapis.com/deepmind-media/Model-Cards/Gemini-3-Pro-Model-Card.pdf}
}

@misc{opus,
  title={System card: Claude opus 4.5},
  author={Anthropic},
  year={2025},
  url = {https://assets.anthropic.com/m/64823ba7485345a7/Claude-Opus-4-5-System-Card.pdf}
}

@article{gloeckle2024better,
  title={Better \& faster large language models via multi-token prediction},
  author={Gloeckle, Fabian and Idrissi, Badr Youbi and Rozi{\`e}re, Baptiste and Lopez-Paz, David and Synnaeve, Gabriel},
  journal={arXiv preprint arXiv:2404.19737},
  year={2024}
}

@article{xiaomi2025mimo,
  title={MiMo: Unlocking the Reasoning Potential of Language Model--From Pretraining to Posttraining},
  author={Xiaomi, LLM and Xia, Bingquan and Shen, Bowen and Zhu, Dawei and Zhang, Di and Wang, Gang and Zhang, Hailin and Liu, Huaqiu and Xiao, Jiebao and Dong, Jinhao and others},
  journal={arXiv preprint arXiv:2505.07608},
  year={2025}
}

@misc{phan2025humanitysexam,
      title={Humanity's Last Exam}, 
      author={Phan, Long and Gatti, Alice and Han, Ziwen and Li, Nathaniel and Hu, Josephina and Zhang, Hugh and Zhang, Chen Bo Calvin and Shaaban, Mohamed and Ling, John and Shi, Sean and others},
      year={2025},
      eprint={2501.14249},
      archivePrefix={arXiv},
      primaryClass={cs.LG},
      url={https://arxiv.org/abs/2501.14249}, 
}

@misc{kilocode,
    note = {Kilo Code webpage},
    howpublished = {\url{https://kilo.ai/}},
    author = {Kilo Code, Inc.},
    year = {2026},
    title = {Move at Kilo Speed}
}

@misc{roocode,
    note = {Roo Code webpage},
    howpublished = {\url{https://roocode.com/}},
    author = {Roo Code},
    year = {2026},
    title = {Your AI Software Engineering Team is here.}
}

@misc{claudecode,
    note = {Claude Code webpage},
    howpublished = {\url{https://claude.com/product/claude-code}},
    author = {ANTHROPIC PBC},
    year = {2026},
    title = {Autocomplete finishes lines. Claude Code finishes features.}
}

@misc{white2025livebenchchallengingcontaminationlimitedllm,
      title={LiveBench: A Challenging, Contamination-Limited LLM Benchmark}, 
      author={Colin White and Samuel Dooley and Manley Roberts and Arka Pal and Ben Feuer and Siddhartha Jain and Ravid Shwartz-Ziv and Neel Jain and Khalid Saifullah and Sreemanti Dey and Shubh-Agrawal and Sandeep Singh Sandha and Siddartha Naidu and Chinmay Hegde and Yann LeCun and Tom Goldstein and Willie Neiswanger and Micah Goldblum},
      year={2025},
      eprint={2406.19314},
      archivePrefix={arXiv},
      primaryClass={cs.CL},
      url={https://arxiv.org/abs/2406.19314}, 
}

@misc{lin2024wildbenchbenchmarkingllmschallenging,
      title={WildBench: Benchmarking LLMs with Challenging Tasks from Real Users in the Wild}, 
      author={Bill Yuchen Lin and Yuntian Deng and Khyathi Chandu and Faeze Brahman and Abhilasha Ravichander and Valentina Pyatkin and Nouha Dziri and Ronan Le Bras and Yejin Choi},
      year={2024},
      eprint={2406.04770},
      archivePrefix={arXiv},
      primaryClass={cs.CL},
      url={https://arxiv.org/abs/2406.04770}, 
}

@misc{openairlhf,
      title={Training language models to follow instructions with human feedback}, 
      author={Long Ouyang and Jeff Wu and Xu Jiang and Diogo Almeida and Carroll L. Wainwright and Pamela Mishkin and Chong Zhang and Sandhini Agarwal and Katarina Slama and Alex Ray and John Schulman and Jacob Hilton and Fraser Kelton and Luke Miller and Maddie Simens and Amanda Askell and Peter Welinder and Paul Christiano and Jan Leike and Ryan Lowe},
      year={2022},
      eprint={2203.02155},
      archivePrefix={arXiv},
      primaryClass={cs.CL},
      url={https://arxiv.org/abs/2203.02155}, 
}

@misc{grporlvr,
      title={DeepSeekMath: Pushing the Limits of Mathematical Reasoning in Open Language Models}, 
      author={Zhihong Shao and Peiyi Wang and Qihao Zhu and Runxin Xu and Junxiao Song and Xiao Bi and Haowei Zhang and Mingchuan Zhang and Y. K. Li and Y. Wu and Daya Guo},
      year={2024},
      eprint={2402.03300},
      archivePrefix={arXiv},
      primaryClass={cs.CL},
      url={https://arxiv.org/abs/2402.03300}, 
}

@misc{genrm,
      title={Generative Verifiers: Reward Modeling as Next-Token Prediction}, 
      author={Lunjun Zhang and Arian Hosseini and Hritik Bansal and Mehran Kazemi and Aviral Kumar and Rishabh Agarwal},
      year={2025},
      eprint={2408.15240},
      archivePrefix={arXiv},
      primaryClass={cs.LG},
      url={https://arxiv.org/abs/2408.15240}, 
}

@article{bradley1952rank,
  title={Rank analysis of incomplete block designs: I. the method of paired comparisons},
  author={Bradley, Ralph Allan and Terry, Milton E},
  journal={Biometrika},
  volume={39},
  number={3/4},
  pages={324--345},
  year={1952},
  publisher={JSTOR}
}
